# A Vision-Language Foundation Model for Precise and Comprehensive Brain Tumor Diagnosis from Preoperative Multimodal Data

Project page link: https://hku-healthai.github.io/brainvlm_project.github.io/

Yinong Wang [†a], Jianwen Chen [†a], Zhou Chen, MD [†b], Shuwen Kuang, MD [†c], Haoning Jiang [a], Yanzhao Shi [v], Huichun Yuan, MD [d], Yan-ran (Joyce) Wang, PhD [e], Bing Wang, MD [g], Lei Wu, MD [h], Bin Tang, MD [i], Li Meng, MD [j], Baihua Luo, MD [k], Bin Zhou, MD [f,l], Wei Ding, MD [m], Weiming Zhong, MD [n], Wei Hou, MD [o], Yuanbing Chen, MD [p], Zhiping Wan, MD [q], Wei Wang, MD [r], Zhenkun Xiao, MD [g], Wenwu Wan, MD [s], Allen He[t], Yuyin Zhou, PhD [u], Prof Longbo Zhang, MD [b,f,x], Feifei Wang, PhD [v], Prof Zhixiong Liu, MD [b,f,x], Prof Michael Iv, MD [w], Xuan Gong, MD PhD [b,f,x,*], Liangqiong Qu, PhD [a,*]

[a] *School of Computing and Data Science, The University of Hong Kong, China*
[b] *Department of Neurosurgery, Xiangya Hospital, Central South University, China*
[c] *Department of Oncology, Xiangya Hospital, Central South University, China*
[d] *Changde Hospital, Xiangya School of Medicine, Central South University (The First People's Hospital of Changde City), China*
[e] *Department of Biomedical Data Science, School of Medicine, Stanford University, Stanford, USA*
[f] *Department of Neurosurgery, Xiangya Hospital, Central South University, Jiangxi (National Regional Center for Neurological Diseases), China*
[g] *Department of Neurosurgery, The Second Affiliated Hospital, Hengyang Medical School, University of South China, China*
[h] *Department of Neurosurgery, The Second Affiliated Hospital, Jiangxi Medical College, Nanchang University, China*
[i] *Department of Neurosurgery, The First Affiliated Hospital, Jiangxi Medical College, Nanchang University, China*
[j] *Department of Radiology, Xiangya Hospital, Central South University, China*
[k] *Department of Pathology, Xiangya Hospital, Central South University, China*
[l] *Department of Neurosurgery, Jiangxi Provincial People's Hospital, The First Affiliated Hospital of Nanchang Medical College, China*
[m] *The Affiliated Children's Hospital of Xiangya School of Medicine, Hunan Children's Hospital, China*
[n] *Department of Neurosurgery, Shenzhen Second People's Hospital, China*
[o] *Department of Neurosurgery, First Hospital of Lanzhou University, China*
[p] *Department of Neurosurgery, The Third Xiangya Hospital, Central South University, China*
[q] *Department of Neurosurgery, Tongji Hospital, School of Medicine, Tongji University, China*
[r] *Department of Radiology, Tongji Hospital, School of Medicine, Tongji University, China*
[s] *Department of Neurosurgery, Chongqing Traditional Chinese Medicine Hospital, China*
[t] *Basis International School Park Lane Harbour, China*
[u] *Department of Computer Science and Engineering, University of California, Santa Cruz, USA*
[v] *Department of Electrical and Electronic Engineering, The University of Hong Kong, China*
[w] *Department of Radiology, School of Medicine, Stanford University, Stanford, USA*
[x] *National Clinical Research Center of Geriatric Disorders, Xiangya Hospital, Central South University, China*

* Corresponding authors
Email addresses: gong.xuan@csu.edu.cn (Xuan Gong), liangqqu@hku.hk (Liangqiong Qu)
† These authors contributed equally to this work.

## Summary

**Background** Non-invasive presurgical diagnosis of brain tumor types from Magnetic Resonance Imaging (MRI) is essential but challenging due to overlapping imaging features across tumor types, inter-observer variability, and the extensive training required for expertise. We aimed to develop an MRI-based Artificial Intelligence (AI) model for automatic and reliable brain tumor classification with diagnostic uncertainty quantification and radiology reports generation.

**Methods** We developed BrainVLM to classify all 12 World Health Organization (WHO) 2021 brain tumor types. BrainVLM integrates an uncertainty quantification strategy to indicate prediction reliability and a module for generating radiology reports to elucidate the clinical rationale. BrainVLM was trained on multi-modal data (MRI scans, demographics, and radiology reports) from 40,043 individuals. It was validated on 5,211 patients with pathologically confirmed brain tumors, including 3,877 held-out patients from the primary hospital and 1,334 patients from 11 independent hospitals. We further conducted two proof-of-concept studies to validate its clinical utility in AI-clinician workflows: 1) a blinded multi-reader study where 12 neuroradiologists across varying experience levels interpreted 248 retrospective cases with or without AI assistance, and 2) a real-world prospective study in which 1,009 patients were independently and blindly assessed by BrainVLM and radiologists before surgery. Additionally, we demonstrated BrainVLM's utility in preoperative molecular subgroup prediction for adult-type diffuse gliomas, using a multi-center cohort of 632 patients.

**Findings** In primary evaluation, BrainVLM achieved an area under the curve (macro-AUC) of 0.85 (95% CI: 0.84–0.86), and an F1 score of 0.82 (95% CI: 0.81–0.83), surpassing neuroradiologists (F1 = 0.80 (95% CI: 0.79–0.81)). In external validation across 11 centers, BrainVLM achieved an AUC = 0.80 (95% CI: 0.79–0.82) and F1 = 0.75 (95% CI: 0.73–0.78), compared with F1=0.71 (95% CI: 0.69-0.73) for neuroradiologists. In prospective real-world evaluation, BrainVLM maintained performance comparable to neuroradiologists. In the blinded multi-reader study, the AI-assisted neuroradiologists demonstrated a statistically 27.6% (absolute 0.16, 95% CI: 20%-35%) improvement in F1 score across all experience levels compared to unaided assessments ($p<0.0001$), while reducing diagnostic time by 34.7% (absolute 55.8 seconds, 95% CI: 29.9%-38.9%). Across 5,211 cases, 72% (3,008/4,156) of correct diagnoses had high-confidence scores (>90%), whereas 70% (736/1,055) of incorrect diagnoses exhibited low-confidence scores (50–85%). For molecular subgroup prediction in adult-type diffuse gliomas, BrainVLM achieved AUCs of 0.95 (95% CI: 0.93-0.96) and 0.88 (95% CI: 0.84-0.91) in primary and external validation, respectively.

**Interpretations** By integrating uncertainty quantification and report generation, BrainVLM improves radiologists' diagnostic performance, reduces MRI interpretation time, and allows lower-confidence cases to be effectively flagged for human review. These findings show that AI-assisted diagnosis based on presurgical MRI can support clinical decision-making, with the potential to enhance early brain tumor diagnosis, alleviate clinicians' workloads, and ultimately improve patient care.

## Research in context

### Evidence before this study

We searched PubMed and Google Scholar for English-language studies published up to December 6, 2025, using the terms “MRI” AND (“brain tumor” OR “CNS tumor” OR “brain cancer”) AND (“machine learning” OR “artificial intelligence” OR “deep learning” OR “foundation model”) in titles or abstracts. We identified no reports describing the deployment of an MRI-based AI model capable of diagnosing the full spectrum of the 12 major brain tumor types defined by WHO CNS5 (the 2021 fifth edition of the WHO Classification of Tumors of the Central Nervous System). Existing literature on AI-assisted brain tumor diagnosis focused almost exclusively on gliomas, meningiomas, and brain metastases, with little attention to other WHO CNS tumor types. Critically, to our knowledge, no previous work has rigorously quantified model reliability or provided clinically interpretable diagnostic rationales (such as radiology style reports) alongside brain tumor classifications. As a result, existing AI outputs for brain tumor classification are limited to isolated predictions, lacking the clinical narrative components that are imperative for clinician trust and real-world adoption.

### Added value of this study

To our knowledge, this is the first study to develop and validate an AI model capable of comprehensive classification across the full spectrum of all 12 major brain tumor types from preoperative multimodal data. By leveraging the largest and most diverse multi-modal neuro-oncology dataset to date, our model achieved robust diagnostic performance with high generalizability, superior or comparable to board-certified neuroradiologists in sensitivity, precision, and F1 scores on both internal and multi-center external validation cohorts. By integrating diagnostic precision with automatic radiology reports generation and reliable uncertainty quantification, our BrainVLM could substantially improve neuroradiologists’ diagnostic accuracy and efficiency across a wide spectrum of tumor types, supporting neuroradiologists at all levels of experience. We publicly release our source code and our training dataset to promote transparency, reproducibility, and future research in the field of neuro-oncology AI.

### Implications of all the available evidence

Use of AI foundation models, when trained on sufficiently large and diverse datasets, has the potential to substantially improve the accuracy and efficiency of preoperative tumor diagnosis across a broad spectrum of tumor types, benefiting radiologists at all levels of experience. Reliable uncertainty quantification and automated report generation enhance the interpretability and clinical utility of AI predictions, helping to build clinician trust and streamline decision-making. Prospective clinical studies are warranted to confirm the real-world benefits and safety of integrating AI-assisted diagnostics into standard neuro-oncological management.

# Introduction

Brain neoplasms can have long-lasting and life-altering physical, cognitive, and psychological impacts on patients’ lives, and can occur in any anatomical region of the encephalon and adjacent structures.[1–3] WHO CNS5 identifies 12 major brain tumor types and over 100 distinct tumor subtypes, ranging from benign and indolent neoplasms such as grade 1 meningioma to malignant and aggressive tumors such as grade 4 glioblastoma.[4] Common treatment options for brain tumors include neurosurgical resection, radiotherapy, and chemotherapy, with surgery serving as the primary initial management in most cases. However, the choice of treatment modality, surgical approach, and the extent of resection depends largely on the initial presurgical diagnosis.[5] For example, maximal safe gross total resection is generally recommended for patients with diffuse gliomas. In contrast, lymphomas and some germ cell tumors may be managed effectively with chemotherapy and/or radiotherapy without surgical resection.[6,7] Hence, a prompt and accurate preoperative diagnosis of brain tumors is an important first step in guiding appropriate treatment.

MRI is the primary diagnostic modality for presurgical diagnosis and longitudinal monitoring of patients with brain tumors.[8] In routine practice, neuroradiologists must review multi-parametric MRI scans to identify and diagnose potential lesions and then convey their findings and impression in a radiology report. However, this manual interpretation is time-consuming and error-prone due to overlapping radiologic characteristics of different brain tumors, inter-observer variability, and potential perceptual errors. This problem is further compounded by the global shortage of specialist neuroradiologists due to the extensive training efforts required to gain expertise. Together, these unmet needs call for advanced techniques to develop an automatic brain tumor diagnostic tool for rapid and precise diagnoses and report generation, thereby accelerating the time-sensitive diagnosis process and reducing clinician workload.

AI, particularly the recently emerged vision language models (VLMs),[9,10] has the potential to advance automatic disease diagnosis. Yet, compared to the flourishing development of AI diagnostic frameworks for other tumors, comprehensive AI models for presurgical brain tumor diagnosis remain significantly underexplored. At least two critical challenges hinder the widespread clinical adoption of AI-based brain tumor diagnosis. First, current publicly available datasets for brain tumor analysis are limited in scope, predominantly featuring gliomas[11], meningiomas, and metastases, while underrepresenting other tumors in the WHO CNS5.[12] Second, existing AI models predict brain tumor types without assessing the confidence and reliability of the prediction or explaining the underlying clinical rationale. This issue is particularly pronounced in large language models (LLMs) or VLMs, which often exhibit unwarranted high confidence in erroneous predictions.[13] This overconfidence in AI predictions, coupled with the opaque decision-making process, undermines clinicians' trust and poses a substantial barrier to their clinical translation and adoption.

In this study, we developed BrainVLM, an MRI-based AI model for precise and automated classification of all 12 WHO CNS5-defined brain tumor categories. For each case, BrainVLM outputs a diagnosis, a confidence score to quantify diagnostic reliability, and a radiology report to explain the clinical rationale. We compared BrainVLM's performance against board-certified neuroradiologists and state-of-the-art AI models, using pathological diagnoses and radiology reports as reference standards. Furthermore, we validated its clinical utility in two proof-of-concept AI–clinician workflow studies: 1) a blinded multi-reader study in which 12 neuroradiologists of varying experience levels interpreted 248 retrospective cases with and without AI assistance, and 2) a real-world prospective study in which 1009 patients were assessed simultaneously by BrainVLM and radiologists prior to surgery. Finally, we demonstrate BrainVLM's utility in preoperative molecular subgroup prediction for adult-type diffuse gliomas using multi-center cohorts.

# Methods

### Datasets and study design

For the development of BrainVLM, we curated BrainTumor48K, a comprehensive multi-modal brain tumor dataset. It comprises 47,947 individuals aggregated from 39 public repositories and 12 collaborating medical centers (Figure 1a, Table 1, and Supplementary Tables 1-4, appendix pp 46-50). The BrainTumor48K comprises 33,149 patients with pathologically confirmed brain tumors (21,993 from public repositories and 11,156 from medical centers) and 14,798 healthy controls (sourced from public repositories) (Figure 1a). The detailed public repositories and their pre-processing pipeline for the public repositories are summarized in the Supplementary Section S.1.1.1 (appendix pp 2-3). Data from 12 independent collaborating medical centers included demographics, 3D multi-parametric MRI scans (T1-weighted (T1), T1 contrast-enhanced (T1c), T2-weighted (T2), and T2-Flair sequences (T2f)), expert-curated radiology reports, and pathological diagnoses.

Overall, the BrainTumor48K dataset spans all 12 major brain tumor types and most of the subtypes defined by the 2021 WHO CNS5.[4] The detailed number of patients across the 12 major brain tumor types and their subtypes is shown in the right panel of Figure 1a.

Data from participating medical institutions were obtained with ethics approval from their respective Institutional Review Boards (IRBs), and the details of these IRBs are presented in the Supplementary Section S1.3 (appendix pp 6-7). The requirement for informed consent was waived by the IRBs. All data were de-identified in compliance with institutional policies and ethical standards. This study complies with the TRIPOD+AI guidelines for the reporting artificial intelligence-based prediction models.[14]

### Development of AI foundation model and retrospective validation

BrainVLM (Figure 1b) was trained on a dataset aggregating 35,227 publicly available cases (including patients with brain tumors and healthy controls) and 4,816 patients from the primary institutional cohort (Xiangya Hospital). A validation set, consisting of 500 public cases and 120 primary cohort cases, was reserved for hyperparameter tuning and model selection. Retrospective diagnostic performance was evaluated on 5,211 patients with pathologically confirmed brain tumors, including 3,877 held-out patients from the primary hospital, and 1,334 patients from 11 independent hospitals (Figure 1a). More details regarding the BrainVLM network architecture, uncertainty quantification, training strategies and validation procedures can be found in the Supplementary Sections S2 (appendix pp 8-25).

### Prospective validation procedures

For the prospective study, we consecutively enrolled 1162 patients with an initial diagnosis of brain lesions at Xiangya Hospital and additional 349 consecutive patients from two external institutional cohorts (Supplementary Tables 3, 4, appendix pp 49-50). Following brain MRI acquisition, clinical information and MRI images were collected and preprocessed according to standardized protocols. Patients were excluded if they had a history of prior brain surgery, previous brain radiotherapy or stereotactic radiosurgery, missing MRI images, or incomplete MRI protocols. Following the application of these criteria (Figure 2c), eligible patients were divided into two groups: surgical group (n = 886; 639 Xiangya, 247 external) and non-operative group (n = 123; 74 Xiangya, 49 external). Eligible patients' data were prospectively predicted using BrainVLM prior to definitive clinical diagnosis. Following patient discharge, BrainVLM outputs were validated against gold standard references: postoperative pathological findings for patients who underwent surgery, or consensus discharge diagnoses established by a multidisciplinary panel of senior clinicians through comprehensive review of radiological and clinical data for non-surgical cases. Finally, patients with non-tumoral pathological findings were also excluded.

**Multi-reader study for AI-augmented clinical assessments**

Multi-reader studies are widely adopted to assess the clinical validity of medical AI models.[15] To achieve a more rigorous and realistic evaluation of BrainVLM's clinical utility in augmenting neuroradiological diagnosis, we developed a gold-standard test dataset comprising 248 patients covering 12 major WHO CNS5 brain tumor types. For each case, readers received only the multi-parametric MRI scans (T1, T1c, T2, T2-Flair) and patient metadata (age, sex); no clinical history was provided. Diagnostic performance was systematically compared across three paradigms: (a) neuroradiologists-only as a reader, (b) BrainVLM as an autonomous diagnostic agent, and (c) neuroradiologists augmented by BrainVLM as a reader. In paradigm (c), BrainVLM assisted neuroradiologists by providing diagnoses with associated confidence scores and a radiology report. Twelve neuroradiologists, stratified by expertise (5 juniors: 3-5 years; 4 seniors: 5-10 years; 3 experts: 10+ years), were recruited in the blinded crossover study. For each case interpretation, neuroradiologists recorded a diagnosis and a self-rated diagnostic confidence score. To minimize memorization effects in scenarios (a) and (c), we reshuffled the order of the cases and modified their anonymized identifiers for the second read, and allowed a one-month wash-out period between the two reads.

**Statistical analysis**

BrainVLM was compared against board-certified neuroradiologists (using clinical radiology reports) and four AI models, with postoperative pathology as the gold standard. RadFM,[16] Merlin,[17] and VST[18] were fine-tuned on BrainTumor48K, while ChatGPT-4o[19] served as a zero-shot generalist baseline. Tumor classification performance was quantified using sensitivity, specificity, precision, Cohen's Kappa, F1, and the area under the curve (AUC). For multi-label classification, we computed the class-frequency-weighted F1 and both macro- and micro-averaged AUC. Report generation was assessed with BLEU-4[20], RaTEScore[21], RadGraph-XL[22], and an LLM-as-Judge framework (Qwen 2.5-72B). We applied two-tailed Wilcoxon signed-rank tests to assess statistical significance and reported 95% confidence intervals. Following previous work, we applied non-parametric bootstrap resampling (1,000 replicates) to estimate 95% CIs.[23] P values were not adjusted for multiple comparisons. The performance metrics in the benchmark analyses (Figures 1d and 3d) and tumor categories in the multi-reader study (Figures 4a-d) were interpreted as distinct analytical objectives, rather than repeated tests of a single hypothesis. Therefore, no Bonferroni or other multiplicity correction was applied. We performed Shapley value analysis on the primary and external test datasets to evaluate

BrainVLM's robustness to missing MRI sequences (Supplementary Section S3.2, appendix pp 25-26).

**Role of funding source**
The funders of this study had no role in data collection, analysis, interpretation, writing of the manuscript, and the decision to submit the manuscript for publication.

## Results

### Diagnostic performance of BrainVLM in primary and external test datasets

We first evaluated BrainVLM's diagnostic ability using the retrospective primary test dataset (n = 3,877). BrainVLM achieved a macro-AUC of 0.85 (95% CI: 0.84–0.86, Figure 1c; 0.87 for both intra-axial and extra-axial tumors, Supplementary Figure 10b, appendix p 65) and a weighted-average F1 score of 0.82 (95% CI: 0.81–0.83), with precision of 0.84 (95% CI: 0.83–0.85) and sensitivity of 0.82 (95% CI: 0.81–0.83; Figure 1d). This exceeded expert neuroradiologists' assessments (F1 = 0.80, 95% CI: 0.79–0.81; precision = 0.71, 95% CI: 0.70–0.73; sensitivity = 0.80, 95% CI: 0.79–0.81), with enhanced inter-rater agreement (Cohen's κ = 0.75, 95% CI: 0.74–0.76 vs 0.69, 95% CI: 0.68–0.70). In benchmarking against four comparator AI models (RadFM, Merlin, VST, and ChatGPT-4o), BrainVLM improved key performance metrics by at least 25%, including in several low-prevalence tumor types (Figure 1d; Supplementary Table 11, appendix p 56). DeLong tests showed that BrainVLM had a higher AUC than each comparison model (Supplementary Table 5 and Supplementary Figure 10, appendix pp 51, 65).

In the external multicentric dataset, BrainVLM maintained robust performance. It achieved a macro-AUC of 0.80 (95% CI: 0.79–0.82, Figure 1c; 0.76 for intra-axial and 0.84 for extra-axial tumors, Supplementary Figure 10b, appendix p 65) and an F1 score of 0.75 (95% CI: 0.73–0.78; precision = 0.77, 95% CI: 0.75–0.78; sensitivity = 0.75, 95% CI: 0.74–0.77; Figure 1d). These results also surpassed both neuroradiologist consensus (F1 = 0.71, precision = 0.72, sensitivity = 0.71) and state-of-the-art models (RadFM: F1 = 0.43, Merlin: F1 = 0.52, VST: F1 = 0.41, ChatGPT-4o: F1 = 0.30).

BrainVLM exhibited slight performance variation across different tumor types (Figure 2a). In the primary test dataset, BrainVLM matched or surpassed expert-level neuroradiologists' performance for common brain tumors: MEN (F1 = 0.91 vs 0.91), GGN (F1 = 0.82 vs 0.81), and CPN (F1 = 0.78 vs 0.74). For low-prevalence or imaging-ambiguous brain tumors, BrainVLM showed superiority: HEM (F1 = 0.64 vs 0.45), EMB (F1 = 0.68 vs 0.58), and CPT (F1 = 0.55 vs 0.47). Confusion matrix analysis demonstrated substantial concordance between BrainVLM and human experts in diagnostic errors, with over 90% overlap in the top two misclassified tumor categories across both the primary test and external datasets (Figure 2b).

Notably, BrainVLM achieved robust performance using only standard MRI sequences (T1, T1c, T2, T2-Flair) and demographic data, whereas neuroradiologists relied on supplemental advanced imaging (e.g., diffusion-weighted imaging, MR angiography, MR spectroscopy, and MR Perfusion) in 56.1% of cases (Supplementary Table 7, appendix p 53). Together, these results establish BrainVLM as an innovative AI system that achieves neuroradiologist-level performance on most brain tumor types across heterogeneous clinical settings without relying on advanced imaging or protocol harmonization.

## Real-world prospective study

To further evaluate the utility of BrainVLM in real-world clinical settings, we conducted a multi-center prospective study aimed at assessing its performance in authentic diagnostic workflows. In the surgical group, BrainVLM achieved F1 scores of 0.78 (primary) and 0.80 (external), surpassing the radiologists' performance of 0.75 and 0.77, respectively (Figure 3a). In the non-operative group, the model maintained robust diagnostic performance, yielding F1 scores of 0.85 in the primary cohort and 0.75 in the external cohort (Figure 3a), both exceeding the neuroradiologists' benchmarks (0.79 and 0.72, respectively). These results demonstrate that BrainVLM provides reliable diagnostic support across diverse clinical management pathways and independent patient populations.

## Uncertainty-aware clinical decision support

Conventional VLMs can generate incorrect diagnoses with inappropriately high confidence or fail to provide uncertainty quantification, potentially undermining clinicians' trust.[24] To address this issue, we developed a consensus-driven strategy to augment BrainVLM with reliable uncertainty quantification (Supplementary Section S2.2, appendix pp 16-18). Across 5,211 cases in the primary and external test datasets, 72% of correct diagnoses (n = 3,008/4,156) were associated with high-confidence (>90%), whereas 70% of incorrect diagnoses (n = 736/1,055) exhibited confidence scores of 50-85%. By contrast, the corresponding model without the consensus-driven strategy assigned low confidence to only 15% of incorrect predictions (n = 160/1,055; Figure 3c). Formal calibration analysis showed good agreement between predicted confidence and observed outcomes, with an expected calibration error (ECE)[25] of 0.036 and a Brier score[26] of 0.1448 (Figure 3b). This was further supported by the reliability diagram, in which observed diagnostic accuracy increased with predicted confidence (Figure 3b and Figure 3c).

We next examined whether these uncertainty estimates could inform decision support in diagnostically ambiguous cases. Cases assigned lower confidence by BrainVLM were associated with lower inter-reader agreement among neuroradiologists (Supplementary Section S3.8.2, appendix p 31), supporting the clinical relevance of the model's uncertainty estimates. We therefore implemented a confidence-triggered supplementary diagnosis mechanism in BrainVLM (Supplementary Section S2.3.1, appendix p 21): when the confidence of the top-ranked diagnosis (Top 1) was below 75%, BrainVLM additionally presented the second-ranked diagnosis (Top 2, Supplementary Figure 9a and d, appendix p 64) for radiologist review. Under this strategy, overall F1 improved from 0.82 (95% CI: 0.81–0.83) to 0.86 (95% CI: 0.85–0.87) in the primary dataset and from 0.75 (95% CI: 0.73–0.78) to 0.78 (95% CI: 0.75–0.80) in the external dataset (Supplementary Section 3.1, appendix p 25).

## Radiology report generation

We next evaluated BrainVLM's ability to generate radiology reports to accompany its diagnostic predictions. We compared BrainVLM-generated reports with those produced by other VLMs. BrainVLM consistently outperformed all baselines across both primary and external test datasets (Figure 3d). In the primary dataset, BrainVLM achieved superior BLEU-4 (0.43 vs 0.02–0.38), F1RadGraph-XL (0.57 vs 0.15–0.46), and RaTEScore (0.75 vs 0.47–0.63). Similarly, in the external dataset, BrainVLM maintained performance, with BLEU-4 (0.35 vs 0.02–0.29), F1RadGraph-XL (0.52 vs 0.22–0.42), and RaTEScore (0.69 vs 0.42–0.58). We further compared BrainVLM with other VLMs on two key components of MRI reporting: signal intensity and tumor location. BrainVLM achieved report accuracies of 0.86 for contrast enhancement, 0.80 for T1 signal intensity, and 0.81 for T2 signal intensity, with a lesion localization accuracy of 0.72; these values were higher than those of comparator VLMs (Figure 3f, Supplementary Section S3.4, Supplementary Figure 7d, appendix pp 27, 63).

**Preoperative molecular subgroup prediction in adult-type diffuse gliomas**

Adult-type diffuse gliomas constitute the predominant malignant tumors affecting the central nervous system,[27] comprising three main subtypes: IDH-mutant astrocytoma, IDH-mutant oligodendroglioma, and IDH-wildtype glioblastoma.[4] Accurate distinction among these subtypes is essential for optimizing treatment strategies and predicting patient outcomes.[28] We finetuned BrainVLM on available adult-type diffuse glioma cases from BrainTumor48K, with threefold cross-validation on the primary dataset (n = 494) and external validation on an independent multi-center cohort (n = 138, subtype distributions in Supplementary Section 2.4, appendix p 21). In primary dataset, BrainVLM achieved a macro-averaged AUC of 0.95 (95% CI: 0.93-0.96; Figure 3e), surpassing Merlin (AUC 0.85, 95% CI: 0.82-0.86) and RadFM (AUC 0.76, 95% CI: 0.74-0.78). In the external cohort, BrainVLM maintained robust performance (AUC 0.88, 95% CI: 0.84-0.91), substantially exceeding Merlin (AUC 0.73, 95% CI: 0.69-0.77) and RadFM (AUC 0.63, 95% CI: 0.58-0.67). These results underscore the exceptional performance of BrainVLM for molecular subtyping, highlighting its potential as a versatile foundation model for brain tumor classification.

**AI-augmented clinical assessments**

In the 248-case multi-reader study, BrainVLM, when evaluated independently, outperformed junior and senior neuroradiologists across brain tumor types and performed comparably with expert neuroradiologists (Figure 4a-d and Table 2). With BrainVLM assistance, neuroradiologists' performance improved across all seniority groups, with a 27.6% increase in mean F1 score and a 34.7% reduction in diagnostic time compared with unaided assessment (both $p<0.0001$). F1 scores increased from 0.52 to 0.67 in junior neuroradiologists, from 0.55 to 0.74 in senior neuroradiologists, and from 0.79 to 0.85 in expert neuroradiologists (Table 2), bringing junior and senior readers closer to unaided expert-level performance.

To characterize clinician–AI interaction in diagnostically challenging settings, we analyzed cases in the lowest quartile of mean reader-reported confidence during unaided assessment (Supplementary Section S3.10, appendix pp 34-35). In this 25% low-confidence subset, the most common diagnostic trajectory across all reader seniority groups was trajectory F (initial clinician and BrainVLM both correct, final diagnosis remains correct; Figure 4f). Trajectory B (initially incorrect diagnosis corrected by a correct AI suggestion) was also frequent, 15.2% / 27.5% / 29.2% of junior / senior / expert assessments. By contrast, trajectory C (initially correct diagnosis reversed by an incorrect AI suggestion) was the least frequent outcome across groups (1.3%, 1.0%, and 0%, respectively; Figure 4f).

To further assess reader responses to incorrect AI outputs, we restricted analysis to cases in which BrainVLM generated an incorrect diagnosis (n = 52 of 248; Supplementary Section

S3.10, appendix pp 34-35). In this subset, overall accuracy was similar before and after AI assistance (50.0% vs 50.9%), with performance maintained or slightly improved in senior and expert readers and a modest decrease in juniors (Figure 4g). This slightly improved performance maybe because many incorrect BrainVLM predictions were associated with low reliability scores (Figure 4e), which may have prompted readers to apply more caution when reviewing these outputs. Overall, these findings do not suggest systematic over-reliance on incorrect AI outputs, although junior neuroradiologists appeared more susceptible to adverse AI influence.

## Discussion

In this study, we developed and clinically evaluated BrainVLM, a multimodal vision-language foundation model for presurgical classification of brain tumors across all 12 WHO CNS5-defined tumor types. Three findings are particularly noteworthy. First, BrainVLM showed strong and generalizable diagnostic performance in retrospective evaluation, achieving AUCs of 0.85 in the primary validation set and 0.80 in the external validation set, with sensitivity, precision, and F1 scores broadly comparable to those of board-certified neuroradiologists across most tumor types. In subgroup analysis (appendix pp 31-33), performance was broadly consistent across sex, MRI vendor, ethnicity, and most age subgroups. Second, in prospective real-world evaluation, BrainVLM achieved a statistically significant 4% higher F1 score than neuroradiologists, suggesting potential clinical utility beyond retrospective benchmarking. Third, the model extended image-based tumor classification towards molecular stratification, achieving a superior AUC on external validation (0.88) relative to state-of-the-art VLMs (0.63–0.73). Taken together, these findings suggest that multimodal foundation models could support diagnostically demanding tasks in neuro-oncology, where overlapping MRI phenotypes, inter-observer variability, and limited familiarity with rare entities continue to constrain consistent presurgical diagnosis.[29]

The robust diagnostic performance of BrainVLM is likely attributable to the scale and diversity of its training dataset, and to its multimodal design. BrainTumor48K represents, to our knowledge, one of the largest and most comprehensive multi-modal presurgical datasets utilized for model development in neuro-oncology to date. Pretraining on extensive corpora comprising public repositories and real-world clinical data might have enabled BrainVLM to accommodate variability in imaging quality, acquisition parameters, patient populations, imaging equipment, and institutional protocols. Another key factor contributing to BrainVLM's performance is its efficient integration of multi-modal data, including MRI scans, patients' demographics, and radiological reports. This framework has enabled us to leverage critical clinical information such as age and radiology reports, leading to more accurate and comprehensive diagnoses.

AI-clinician collaboration holds significant promise in medical image interpretation.[30–33] However, even when AI models exhibit robust diagnostic performance and generalizability, translating these capabilities into clinical practice necessitates overcoming a critical barrier: clinician trust in AI-driven decision-making. The effective integration of AI models into clinical workflows requires clinicians to (1) understand the AI's decision rationale to inform their judgments, and (2) develop sufficient trust in the system's recommendations. To address these challenges, BrainVLM introduces two clinician-trust-focused innovations: a consensus-driven uncertainty quantification strategy and an automated radiology report generation module. In our evaluation, BrainVLM demonstrated clinically meaningful self-assessment capabilities, assigning high-confidence probabilities (>90%) to 72% of its correct diagnoses while appropriately tagging 70% of its incorrect diagnoses with lower confidence scores 50-85%. This self-awareness capability is particularly impactful in diagnostically challenging scenarios. This mechanism ensures only high-confidence predictions are clinically actionable, while automatically triaging ambiguous cases to human experts for secondary review. This approach also prevents clinicians from over-reliance on model outputs, thereby aligning with responsible clinical workflows in healthcare settings. Complementing this reliability mechanism, BrainVLM exhibits an exceptional capability to generate preliminary medical reports—a critical function for communicating diagnostic findings and guiding care across several specialties.[30,34] These reports offer clinicians actionable insights beyond isolated predictions, enabling clinicians to process cases more efficiently, thereby improving turnaround times and expanding access to specialty-level reporting[35]. The clinical validation through a multi-reader study with 12 neuroradiologists across expertise levels further confirmed BrainVLM's trust-building potential.

In clinical practice, incomplete MRI acquisition (e.g., missing T1c) frequently occurs due to insufficient clinician awareness or contraindications. While conventional multi-sequence-trained models exhibit considerable performance degradation under such suboptimal imaging conditions, BrainVLM maintains diagnostic performance comparable to the full-sequence baseline in scenarios where T2, T2-Flair, or T1 sequences are missing (appendix pp 25-26). A notable exception is the absence of T1c sequences, where we observe a 13.4% decline in F1 score (appendix pp 25-26), highlighting the critical role of post-contrast imaging in brain tumor assessment. Another feature relevant to clinical deployment is that BrainVLM was trained on raw, minimally processed MR images, without skull stripping or segmentation (appendix p 11). Unlike many conventional neuroimaging studies, which typically require preprocessing steps such as skull stripping, BrainVLM directly analyzes raw institutional data, with only slice-level resampling applied (appendix p 11). This minimal preprocessing preserves the complete informational content of the original MRI data, making BrainVLM particularly well-suited for our comprehensive brain tumor characterization study. Collectively, BrainVLM's ability to deliver robust diagnostic performance without preprocessing workflows (e.g., skull stripping) or segmentation requirements, combined with its resilience to missing MRI sequences, establishes its clinical efficacy as an adaptable and practical solution for real-world neuroimaging challenges.

Several limitations exist in our study. First, although online datasets were collected globally, real-world diagnostic applications in this study were confined to East Asian patients. Second, despite improved performance in low-prevalence tumors, data on rare tumors remain relatively limited. Further validation studies with larger sample sizes and more diverse ethnic populations, especially for rare tumor types, are warranted to better evaluate the model's generalizability and robustness. Third, while patient demographics is incorporated into BrainVLM, the model does not currently integrate multimodal clinical data such as medical history, lab findings and neurological examinations. Including these data types could further improve diagnostic accuracy and better align the model with real-world clinical decision-making. Correspondingly, in the multi-reader study, readers were only given MRI, age, and sex, matching BrainVLM's inputs, to ensure a fair comparison; this may underestimate clinicians' real-world performance, as full clinical context is routinely accessible in routine care. Fourth, although we used patient-level partitioning, independent institutional cohorts, exclusion of recurrent or postoperative follow-up cases, and external test centers that did not contribute to training or validation, residual risks of dataset overlap or data leakage cannot be completely excluded. This is particularly relevant for large public datasets aggregated from multiple sources and because direct cross-center patient matching was not possible after de-identification. Finally, we assessed diagnostic performance, report-generation quality and reader performance with or without AI assistance, but did not examine subsequent treatment decisions and clinical outcomes, such as extent of resection, surgical complications or long-term prognosis".

In conclusion, BrainVLM demonstrates the potential of multimodal foundation models to support complex diagnostic tasks in neuro-oncology. Trained on well-curated multimodal neuroimaging data and validated across internal, multi-center external, and prospective clinical cohorts, the model achieves generalizable performance for classifying a broad spectrum of brain tumors. BrainVLM may serve as a reliable assistant tool to improve the accuracy, efficiency, and consistency of preoperative brain tumor diagnosis and automated report generation. Our framework, which integrates systematic data curation, advanced vision-language modeling, uncertainty quantification, automated report generation, and rigorous clinical evaluation, offers a structured approach for developing specialized foundation models in other medical fields.

**Contributors**

L.Q. and X.G. conceptualized the study. J.C., Y.W., H.J., L.Q., and A.H. conducted data collection and preprocessing from public repositories. X.G., Z.C., S.K., B.Z., B.W., Z.X., H.Y., Z.W., W.W., B.T., W.H., W.D., W.Z., L.M., A.H., B.L., L.W., Y.C., and W-w.W. collected data from medical centers. X.G., Z.C., S.K., J.C., Y.W., and H.J. performed data preprocessing for institutional data. Y.W., J.C., S.Z., H.J., and L.Q. developed and trained the deep learning algorithms and conducted model analysis. X.G., L.Q., M.I., Y-R.W., F.W., Y.Z., L.Z., and Z.L. provided domain knowledge and interpreted the findings. L.Q., X.G., Y.W., and J.C. drafted the primary manuscript and prepared the figures, with input from all authors. All authors had access to all the data in the study, reviewed and approved the final version of the study. All authors contributed to the general discussion, manuscript revision, and approved the final version. L.Q. and X.G. co-supervised the study. L.Q. and X.G. were responsible for the decision to submit the manuscript.

**Data sharing**

The online training data are available on their corresponding website. The real-world data are not publicly accessible owing to patient confidentiality requirements. Reasonable requests for access may be considered by the corresponding author, subject to approval from the institutional review boards at all participating centers. The BrainVLM project page is available at https://hku-healthai.github.io/brainvlm_project.github.io/

**Declaration of interests**

We declare no competing interests.

**Acknowledgments**

This study was supported by the National Natural Science Foundation of China (62306253), the Early Career Fund (27207025, 27204623), the Clinical Research Fund of the National Clinical Research Center for Geriatric Disorders (2023LNJJ19), the Natural Science Foundation of Hunan Province (2023JJ30927), and the Guangdong Natural Science Fund-General Program (2024A1515010233).

**Table 1:** Characteristics of our retrospective online dataset, primary dataset, and external test dataset for each brain tumor type.

| | **Online dataset** | | | **Primary dataset** | | | | **External test dataset** | | | | **Entire dataset** |
|---|---|---|---|---|---|---|---|---|---|---|---|---|
| | No. of subjects | Dimension | | No. of subjects | Sex | | Age in years Mean ± SD (range) | No. of subjects | Sex | | Age in years Mean ± SD (range) | |
| | | 2D | 3D | | Male | Female | | | Male | Female | | |
| Total | 21,993 | 13,096 | 8,897 | 8,813 | 4,256 | 4,557 | 45 ± 18 (1-86) | 1,334 | 665 | 669 | 52 ± 15 (1-86) | 32,140 |
| MET | 1,654 | 879 | 775 | 735 | 435 | 300 | 56 ± 11 (7-79) | 116 | 69 | 47 | 60 ± 11 (34-86) | 2,505 |
| GCT | 81 | 62 | 19 | 378 | 286 | 92 | 19 ± 13 (1-67) | 18 | 11 | 7 | 20 ± 13 (1-40) | 477 |
| GGN | 9,935 | 4,743 | 5,192 | 3,028 | 1,681 | 1,347 | 41 ± 19 (1-86) | 414 | 244 | 170 | 50 ± 15 (5-79) | 13,377 |
| MEN | 5,363 | 3,804 | 1,559 | 2,395 | 723 | 1,672 | 53 ± 12 (1-81) | 355 | 115 | 240 | 56 ± 13 (8-79) | 8,113 |
| TSR | 674 | 614 | 60 | 796 | 402 | 394 | 46 ± 16 (2-81) | 186 | 98 | 88 | 51 ± 15 (5-75) | 1,656 |
| MNM | 912 | 457 | 455 | 447 | 234 | 213 | 39 ± 18 (1-78) | 76 | 40 | 36 | 43 ± 16 (8-76) | 1,435 |
| CPN | 1,865 | 1,392 | 473 | 416 | 138 | 278 | 46 ± 15 (1-75) | 90 | 43 | 47 | 54 ± 10 (34-75) | 2,371 |
| CPT | 259 | 216 | 43 | 129 | 63 | 66 | 28 ± 20 (1-66) | 6 | 1 | 5 | 45 ± 10 (39-62) | 394 |
| HEM | 535 | 396 | 139 | 164 | 100 | 64 | 47 ± 22 (2-77) | 44 | 27 | 17 | 62 ± 11 (30-82) | 743 |
| EMB | 288 | 165 | 123 | 205 | 131 | 74 | 18 ± 15 (1-70) | 18 | 12 | 6 | 18 ± 15 (3-40) | 511 |
| PIN | 346 | 312 | 34 | 70 | 36 | 34 | 36 ± 20 (1-73) | 6 | 2 | 4 | 39 ± 13 (21-57) | 422 |
| MEL | 81 | 56 | 25 | 50 | 27 | 23 | 50 ± 14 (14-73) | 5 | 3 | 2 | 36 ± 25 (21-69) | 137 |

Abbreviations: MET = brain metastases, GCT = germ cell tumors, GGN = gliomas, glioneuronal tumors, and neuronal tumors, MEN = meningioma, TSR = tumors of the sellar region, MNM = mesenchymal, non-meningothelial tumors, CPN = cranial and paraspinal nerve tumors, CPT = choroid plexus tumors, HEM = hematolymphoid tumors, EMB = embryonal tumors, PIN = pineal region tumors and MEL = melanocytic tumors.

**Table 2:** Comparative diagnostic performance across neuroradiologists-only, BrainVLM, and AI-augmented neuroradiologists paradigms. Twelve neuroradiologists with varying years of experience (junior [3-5 years], senior [5-10 years], expert [10+ years]) participated in this blinded crossover study. Boldface indicates the highest performance metric within each tumor type across AI, neuroradiologists, and AI-augmented neuroradiologists cohorts.

| | $F_1$ score | | | | | | |
|---|---|---|---|---|---|---|---|
| | AI model | Junior | Junior-AI | Senior | Senior-AI | Expert | Expert-AI |
| MET | 0.74 | 0.51 | 0.62 | 0.44 | 0.66 | 0.80 | **0.87** |
| GCT | 0.75 | 0.28 | 0.54 | 0.40 | 0.59 | 0.70 | **0.79** |
| GGN | 0.83 | 0.64 | 0.72 | 0.66 | 0.80 | 0.84 | **0.91** |
| MEN | 0.85 | 0.74 | 0.81 | 0.77 | 0.85 | **0.90** | **0.90** |
| TSR | 0.86 | 0.74 | 0.77 | 0.78 | 0.81 | 0.85 | **0.88** |
| MNM | 0.56 | 0.10 | 0.33 | 0.07 | 0.42 | 0.50 | **0.58** |
| CPN | 0.77 | 0.57 | 0.74 | 0.69 | 0.79 | 0.73 | **0.84** |
| CPT | 0.65 | 0.30 | 0.53 | 0.36 | 0.68 | 0.77 | **0.88** |
| HEM | 0.75 | 0.22 | 0.56 | 0.21 | 0.70 | 0.70 | **0.86** |
| EMB | **0.82** | 0.14 | 0.62 | 0.31 | 0.75 | 0.54 | 0.68 |
| PIN | 0.77 | 0.43 | 0.62 | 0.48 | 0.68 | 0.70 | **0.75** |
| MEL | 0.25 | 0.00 | 0.40 | 0.00 | 0.69 | 0.63 | **0.80** |
| Frequency -weighted $F_1$ | 0.78 [0.73, 0.83] | 0.52 [0.48, 0.55] | 0.67 [0.63, 0.71] | 0.55 [0.51, 0.58] | 0.74 [0.70, 0.77] | 0.78 [0.74, 0.82] | **0.85 [0.81, 0.89]** |
| Accuracy | 0.79 [0.74, 0.83] | 0.55 [0.52, 0.59] | 0.69 [0.65, 0.72] | 0.59 [0.56, 0.62] | 0.75 [0.76, 0.87] | 0.80 [0.75, 0.83] | **0.86 [0.82, 0.89]** |
| Total Time (Mean ± SD) | **51 ± 3** | 189 ± 41 | 112 ± 13 | 161 ± 43 | 104 ± 21 | 100 ± 5 | 75 ± 15 |

Abbreviations: MET = brain metastases, GCT = germ cell tumors, GGN = gliomas, glioneuronal tumors, and neuronal tumors, MEN = meningioma, TSR = tumors of the sellar region, MNM = mesenchymal, non-meningothelial tumors, CPN = cranial and paraspinal nerve tumors, CPT = choroid plexus tumors, HEM = hematolymphoid tumors, EMB = embryonal tumors, PIN = pineal region tumors and MEL = melanocytic tumors. SD: Standard Deviation.

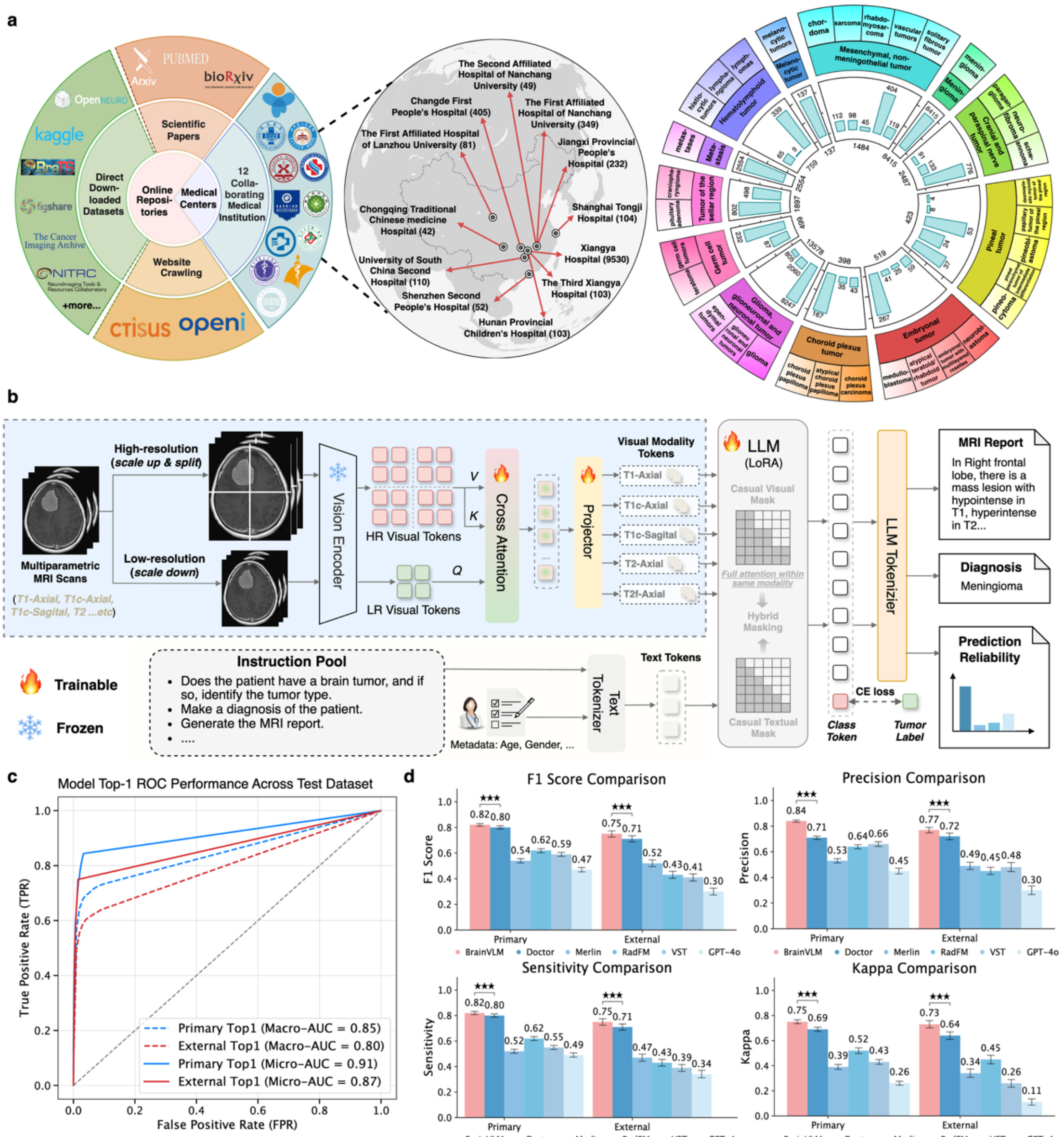

Figure 1: **Overview of this study. a,** A multi-modal brain tumor dataset was constructed from 39 online sources and 12 collaborating medical institutions, comprising 47,947 patients. Each case includes either 2D MRI slices or multi-sequence 3D MRI scans (T1-weighted[T1], T1 contrast-enhanced [T1c], T2-weighted [T2], T2-Flair [T2f]), paired with pathologically confirmed diagnoses, optional radiology reports (or text descriptions). The detailed number of patients across 12 brain tumor types and their subtypes is shown on the right panel. **b,** we developed BrainVLM, a vision-language foundation model designed for brain tumor analysis. The model processes multi-parametric MRI scans at different resolutions through a shared vision encoder and integrates text data via an LLM-based architecture. BrainVLM was trained to classify tumor types with quantified diagnostic uncertainty and a radiology report. **"Frozen"** components (Vision Encoder and LLM backbone) retain their pre-trained weights to ensure structural stability and leverage broad medical knowledge. **"Trainable"** components (MLP projector and LoRA-based adapters) are optimized during fine-tuning via backpropagation. **c,** ROC curves for top-1 predictions on the retrospective primary and external datasets, with macro-averaged AUCs of 0.85/0.80 and micro-averaged 0.91/0.87, respectively. **d,** BrainVLM outperformed neuroradiologists and baseline models in sensitivity, precision, F1 score, and Cohen's $\kappa$ across both datasets. Two-tailed Wilcoxon signed-rank tests were performed to compare BrainVLM with neuroradiologists for each metric. P values were not adjusted for multiple comparisons because each metric was prespecified and interpreted as a distinct measure of diagnostic performance. Statistical significance is indicated as ns (not significant, p>0.05), ⋆ (p<0.05), ⋆⋆ (p<0.01), and ⋆⋆⋆ (p<0.0001). Bootstrap resampling (1,000 replicates) was used to estimate 95% confidence intervals, shown as error bars around the mean.

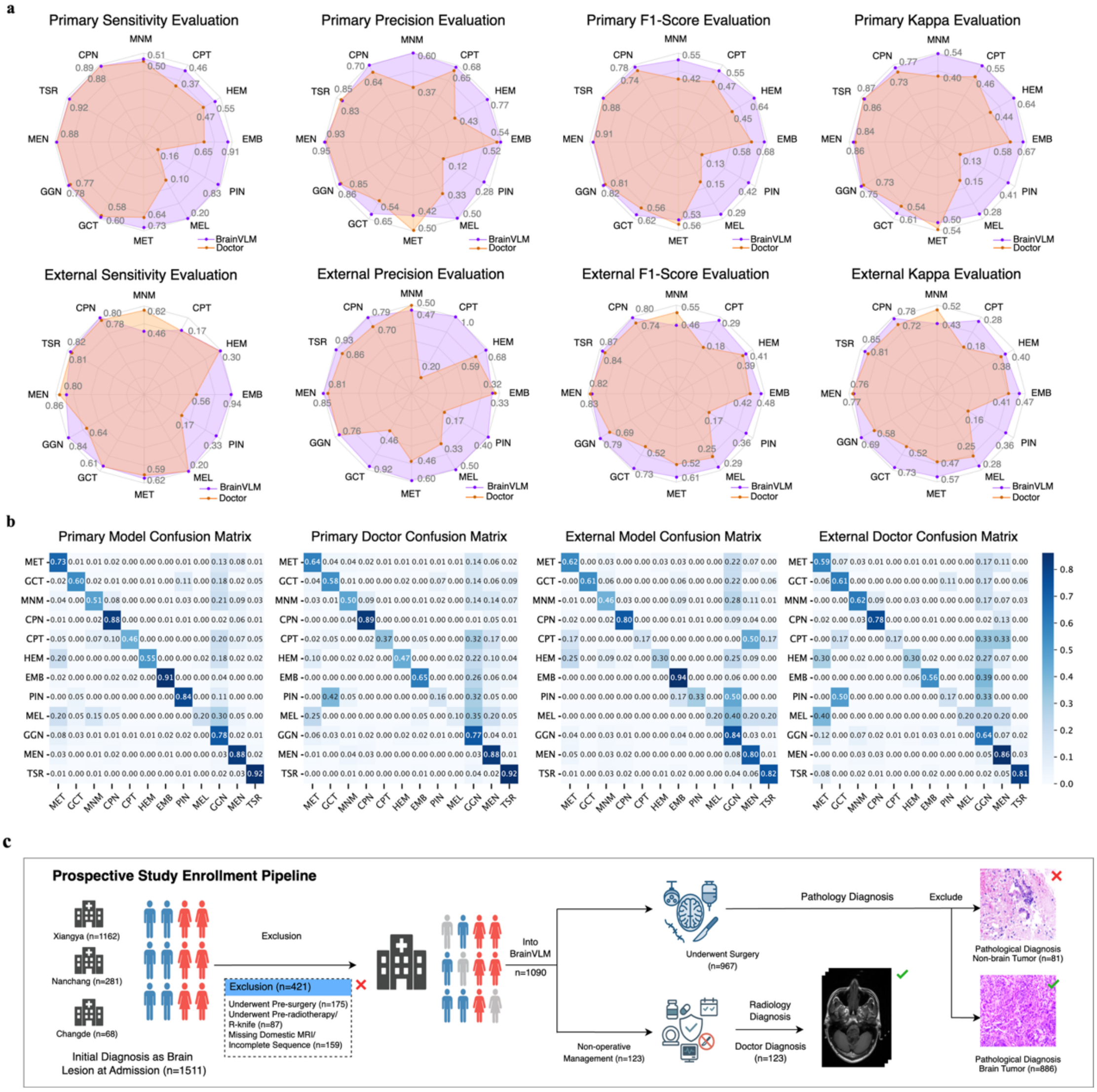

**Figure 2 a,** Performance comparison of BrainVLM and board-certified neuroradiologists on 12 tumor types across primary and external datasets. Model and human expert results are visualized as a dodecagon, with each vertex corresponding to a specific tumor type. For both datasets, we report sensitivity, precision, F1 score, and Cohen's $\kappa$ for each class. BrainVLM demonstrates either superior or comparable performance relative to neuroradiologists on both primary and external cohorts, including for rare tumor types (e.g., EMB, GCT, HEM). **b,** Confusion matrices comparing BrainVLM's predictions to those of neuroradiologists across 12 tumor types in both primary and external datasets. The foundation model showed stronger diagonal concentration than neuroradiologists, indicating improved classification accuracy. Each matrix was normalized by true labels to highlight per-class performance. Notably, the model's diagnostic error patterns showed substantial concordance with those of human experts. **c,** Flowchart of prospective patient enrollment. A total of 1,511 patients with suspected brain lesions were prospectively enrolled. After excluding those who had received prior treatment or lacked diagnostic sequences, 1,090 patients were eligible for evaluation. Among them, 967 underwent surgery; after excluding 123 cases with non-brain tumor pathology, 886 patients with confirmed brain tumors were included for model validation. The remaining 81 patients who received non-operative management were also included for evaluation.

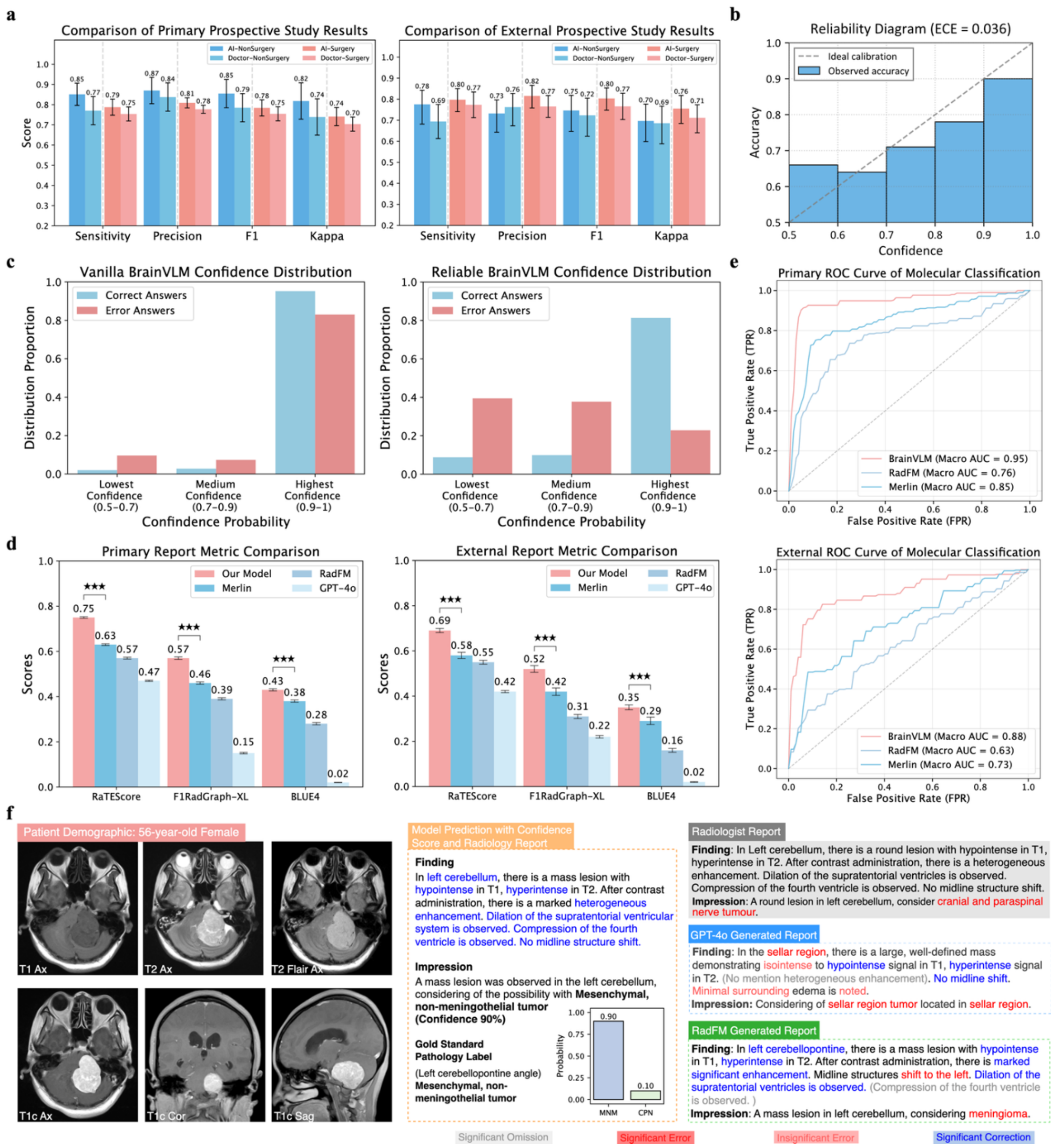

**Figure 3: Performance comparison in the prospective cohort, as well as uncertainty quantification, comparative evaluation of radiology report generation, and molecular subgroup prediction. a,** Comparison of primary and external prospective study results. The bar charts evaluate BrainVLM (AI) versus human doctors across surgical and non-surgery groups using Sensitivity, Precision, F1, and Kappa metrics. **b,** Reliability diagram (ECE = 0.036) illustrating model calibration and the alignment between observed accuracy and confidence probability. **c,** the comparison of confidence score distributions between Vanilla BrainVLM, which does not incorporate our consensus-driven strategy for uncertainty quantification, and Reliable BrainVLM, which benefits from the integration of this strategy. **d,** Report metric comparison of four models using RaTEScore, F1RadGraph-XL, and BLEU4 on both primary and external datasets. Two-tailed Wilcoxon signed-rank tests were performed to compare BrainVLM with comparator models; p values were not adjusted for multiple comparisons, as each metric was interpreted individually rather than as repeated tests of a single hypothesis. **e,** ROC curves comparison for preoperative molecular subgroup prediction in adult-type diffuse gliomas among RadFM, Merlin, and our model on both primary and external datasets. **f,** A representative case study comparing generated reports from our model, ChatGPT-4o, and RadFM against the human radiologist's reference report, highlighting significant corrections, errors, and omissions. For BrainVLM's diagnosis, the associated probability and confidence score are also presented.

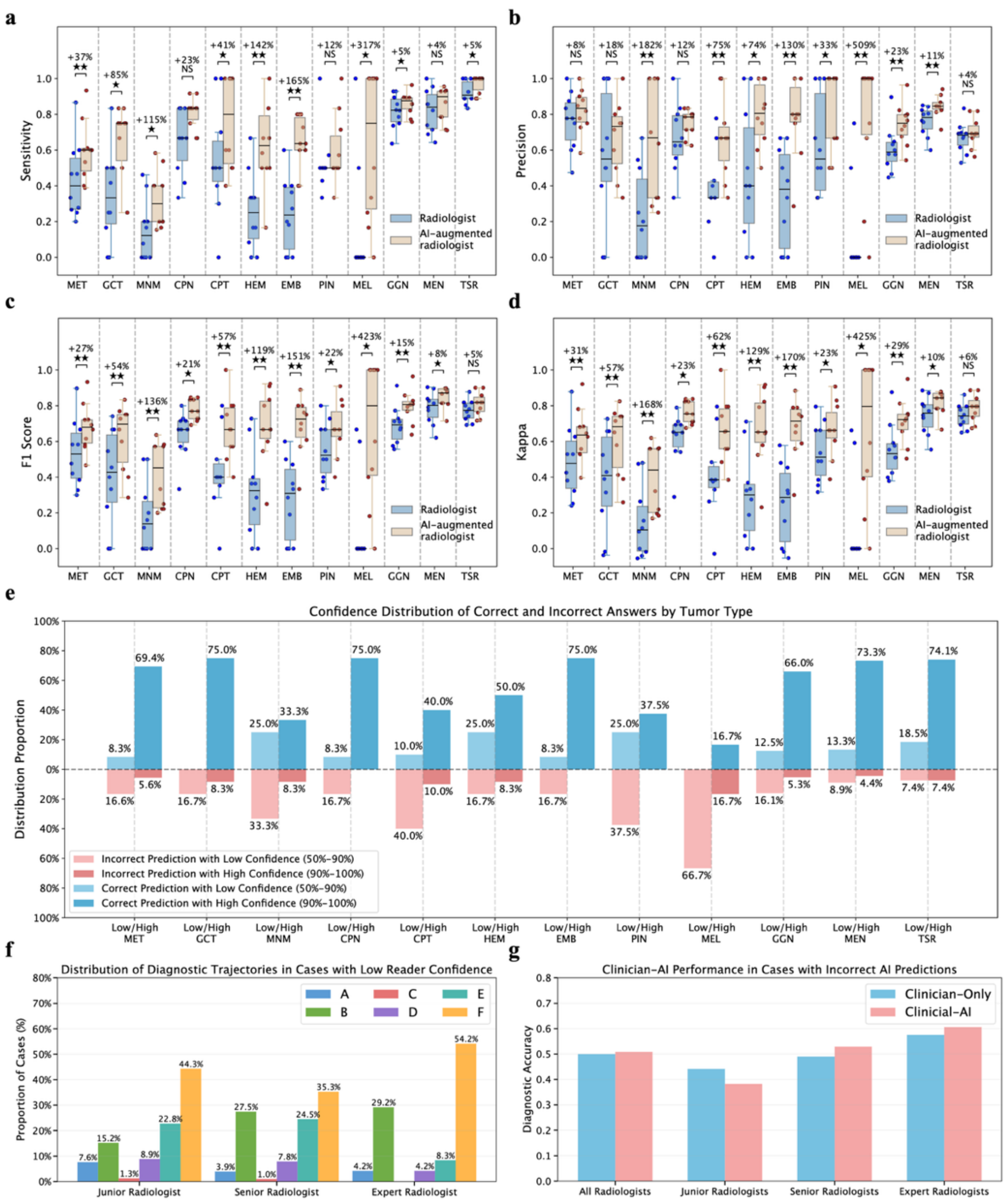

Figure 4: **Blinded multi-reader study on 248 patients with interpretations by 12 neuroradiologists (5 juniors [3-5 years], 4 seniors [5-10 years], 3 experts [10+ years]) with and without AI assistance. a-d,** Comparison between the diagnostic performance of neuroradiologists and AI-augmented neuroradiologists. Box plots summarize sensitivity, precision, F1 score, and Cohen's $\kappa$ for individual neuroradiologists and their AI-augmented counterparts. Pairwise comparisons between each neuroradiologist and their corresponding AI-augmented counterpart were performed using two-tailed Wilcoxon signed-rank tests without correction for multiple comparisons. Statistical significance is indicated as ns (not significant, $p>0.05$), $\star$ ($p<0.05$), $\star\star$ ($p<0.01$), and $\star\star\star$ ($p<0.001$). P values were not adjusted for multiple comparisons because per-tumor-type analyses were prespecified and interpreted separately for each tumor category. **e,** Confidence score distribution for BrainVLM's predictions on these 248 cases across 12 major WHO CNS5 brain tumor types. The figure is vertically divided: upper blue bars show confidence distributions for correct predictions, lower pink bars show incorrect ones. **f,** Distribution of Diagnostic Trajectories in Cases with Low Reader Confidence. The bar chart shows the distribution of four diagnostic trajectories after exposure to BrainVLM among the subset with the lowest 25% of initial clinician confidence, stratified by reader seniority. Trajectories are defined as: (A) maintaining a correct diagnosis despite incorrect AI; (B) successfully correcting an initial error; (C) erroneously changing a correct diagnosis due to incorrect AI; (D) both clinician and AI remaining incorrect; (E) failing to adopt a correct AI suggestion; and (F) both initial clinician diagnosis and AI suggestion being correct, with the final diagnosis remaining correct. **g,** Clinician-AI performance in cases with incorrect AI predictions. The bar chart compares clinician diagnostic accuracy without AI assistance and after review of BrainVLM output, restricted to cases in which BrainVLM generated an incorrect diagnosis, stratified by reader seniority (junior, senior, and expert).

**Contents of Supplementary**

# S1 Dataset

## S1.1 Construction of BrainTumor48K dataset

We assembled the BrainTumor48K dataset through systematic curation from 39 online sources and 12 collaborating medical centers. To guide the search and data collection processes, we first established a comprehensive set of brain tumor keywords, including 12 major brain tumor types and all subtypes, along with their synonyms and related terms. Next, we leveraged retrieval-augmented generation (RAG) and prompt engineering techniques to design an automated preprocessing pipeline (see Supplementary Figure 1). This pipeline transformed the curated raw datasets into structured formats of {*Patient_ID*: multi-parametric 3D MRI volumes (or 2D MRI slice), pathological diagnosis label, optional radiology report (or text description)} for model training. We also expanded our datasets by incorporating data from healthy individuals. For the multi-parametric MRI scans, we considered T1-weighted (T1), T1 contrast-enhanced (T1c), T2-weighted (T2), and T2-Flair (FLAIR), which are commonly used in brain tumor diagnosis.

### S1.1.1 Dataset construction from online repositories

We compiled our dataset from diverse online repositories, categorized based on their inherent structure and the required preprocessing complexity. The first category consists of well-structured sources of medical imaging data, such as established repositories (e.g., BraTS2023,[1–6] TCIA,[7–20] Kaggle,[21–24], Ctisus[25], and other public repositories[26–40]) as well as specialized medical platforms like Radiopaedia.[41] This category encompasses sources that provide readily accessible, structured medical imaging data (including 3D multi-parametric MRI volumes or 2D slices), which are typically paired with pathological labels. Depending on the source, additional information, such as patient metadata or radiology reports, may also be available. The second category involves non-dedicated imaging sources requiring advanced figure/text extraction, such as PubMed Central scientific archives.[42] This platform, not originally designed for medical imaging storage, necessitated specialized pipelines for isolating figures/text from heterogeneous formats (e.g., publication figures, video stills) and reconciling fragmented clinical annotations. In the following, we detail the distinct data pre-processing pipelines developed for each category.

**Pre-processing pipeline for structured repositories and pre-curated datasets.** Datasets in this category are highly heterogeneous, encompassing various sources with differing image formats, annotations, and metadata. These datasets range from simplified collections (where each patient is represented by a single 2D MRI slice with a slice-level pathological label) to comprehensive datasets containing multi-parametric 3D MRI sequences, tumor segmentation masks, patient metadata, and pathological labels. For datasets providing only tumor classification labels, we processed the data into a standardized structure of the form: {*Patient_ID*: 3D MRI volumes (or 2D MRI slice), pathological label}. The resulting standardized data were subsequently used for tumor classification tasks. For datasets containing 3D multi-parametric MRI scans (or 2D MRI slice), segmentation masks, and metadata (e.g., BraTS,[1–6] TCIA[7–20] and Kaggle datasets[21,22]), we first applied the Brainnetome Atlas[43] to localize anatomical brain regions based on the tumor masks. We then leveraged useful metadata (including imaging modality, sex, diagnostic label, and tumor description) to generate comprehensive medical reports via GPT-4o prompted with RAG techniques (see Supplementary Figure 1b). For certain web-sourced datasets, such as Radiopaedia,[41] which include multi-parametric MRI scans and accompanying radiology reports, we removed irrelevant information and reformatted the data into a consistent structure: {*Patient_ID*: 3D MRI volumes, pathological label, radiology report}.

This comprehensive pre-processing pipeline enabled the assembly of a diverse cohort comprising (see Supplementary Table 2): 1) 6,957 brain tumor patients with 3D multi-parametric MRI scans, radiology reports, and associated pathological labels; 2) 1940 brain tumor patients with 3D multi-parametric MRI scans and pathological labels only; 3) 233 patients with 2D MRI slice-diagnosis; and 4) 3,178 healthy control patients with multi-parametric 3D MRI scans. Apart from the datasets detailed above, this preprocessing pipeline also incorporated subjects from additional repositories, such as Br35H,[21] Brain Tumor Classification Dataset,[24] and Brain Tumor MRI Images 44 Classes.[23] However, as these repositories only contain 2D MRI slices with associated pathological labels and lack patient identifiers, they were not included in our dataset summary here (see Supplementary Table 2).

**Pre-processing pipeline for unstructured web-based resources.** Web-crawled brain tumor data from platforms such as PubMed Central,[42] ImageCLEF[44] often contain noisy 2D MRI scans and misaligned or inconsistent text descriptions. To address these challenges, we developed a comprehensive data preprocessing pipeline that transforms raw, heterogeneous web data into a structured format: {*Patient_ID*: 2D MRI slice, text description, pathological label}, see Supplementary Figure 1a2.

For scientific articles from PubMed Central, we systematically retrieved 2D brain MRI slices by querying predefined brain tumor keywords, associating each image with both “golden captions” and inline textual mentions as candidate descriptions. However, these figures frequently contain composite images with multiple subfigures and inconsistent captioning. To resolve this, we developed a three-stage refinement protocol: Firstly, we applied a Faster R-CNN object detector,[45] trained on the MedICaT dataset,[46] to perform subfigure separation and irrelevant images filtering. Secondly, we employed GPT-4o with specialized prompts for accurate subcaption separation. Finally, we established a precise one-to-one subfigure-caption correspondence through bounding box coordinate alignment.[47] ImageCLEF data required minimal processing, with only irrelevant images filtered, as slice-text pairs were directly utilized.

In total, our curated dataset includes 24,333 unique subjects: 23,849 from PubMed Central (116,296 2D MRI slice-text pairs), and 484 from ImageCLEF (4,631 2D MRI slice-text pairs).

**S1.1.2 Dataset construction from medical institutions**

To address the scarcity of rare brain tumor types/subtypes, **the need for a prospective validation study** and the limited availability of expert-curated radiology reports in online repositories, we curated data from twelve collaborating hospitals (Supplementary Table 1). For each patient with a pathologically confirmed diagnosis, we collected de-identified multi-parametric MRI scans (T1, T1c, T2, T2-Flair) that are commonly used for CNS tumor diagnosis, along with de-identified radiology reports and associated pathological diagnoses. In total, we gathered multi-modal data from 8,813 patients at the primary Xiangya Hospital, 1,334 patients from 11 other independent institutions, and 1009 patients in the prospective validation cohort (713 patients from Xiangya Hospital (May 2025 to June 2025), 247 patients (Jan 2025 to Feb 2026) from First Affiliated Hospital of Nanchang University and 49 patients from Changde First People’s Hospital). Demographic and statistical summaries of these medical centers are provided in Figure 1 and Supplementary Tables 1, 3, 4.

The original institutional radiology reports were in Chinese and often included descriptions of other modalities (e.g., Magnetic Resonance Angiography (MRA), Magnetic Resonance Spectroscopy (MRS), Blood Oxygen Level Dependent Imaging (BOLD) signals; Supplementary Table 7). Thus, we implemented a four-stage standardization protocol to process these reports (Supplementary Figure 1c). Firstly, we used regular-expression filters to eliminate non-T1/T1c/T2/T2-Flair content. Secondly, we developed a clinician-validated bilingual lexicon (Chinese-English) to accurately map medical terms, translating terms such as “T2 长信号” to “T2 Hyperintense” and “脑实质” to “brain parenchyma”. Thirdly, leveraging this lexicon as a knowledge base, we utilized RAG techniques and prompt engineering strategies through the advanced capabilities of ChatGPT-4o,[48] to facilitate the translation of the reports from Chinese to English. To ensure terminological precision and minimize hallucinations, we implemented a structured, human-in-the-loop workflow. Following the initial generation, we employed an iterative multi-agent refinement loop where an ensemble of independent LLMs (DeepSeek[49], ChatGPT, and Gemini[50]) cross-checked the outputs, automatically flagging errors that were fed back for targeted correction.[51] This self-correction cycle repeated until consensus was reached. Finally, the translated reports were thoroughly reviewed and refined by board-certified neuroradiologists to ensure their accuracy. To formally evaluate the overall translation quality and inter-reviewer agreement, we randomly sampled 200 generated reports. Five neuroradiologists independently assessed the holistic quality of each complete report using a standardized 5-level scoring rubric (ranging from Level 1: Unacceptable to Level 5: Clinically Accurate):

- Level 5: Clinically Accurate / Excellent – The translation is entirely accurate, precise, and requires no edits. It fully conveys the original clinical meaning.
- Level 4: Minor Inaccuracies / Good – Contains minor omissions or negligible inaccuracies that do not impact clinical interpretation or patient management. For example, if "鞍上区" (suprasellar region) was translated as "in sellar region" instead of "suprasellar region," this would receive a Level 4 rating, indicating a minor anatomical inaccuracy without significant clinical impact.
- Level 3: Moderate Errors / Fair – Contains errors that do not lead to fundamental misdiagnosis but significantly compromise report reliability, clarity, or completeness.
- Level 2: Significant Errors / Poor – Contains serious factual clinical errors that could potentially mislead clinical decision-making or patient care. For example, mistranslating "无强化" (no enhancement) as "enhancement present" would fall into this category.
- Level 1: Unacceptable / Harmful – Report content is completely detached from or contradictory to the actual imaging findings, potentially leading to patient harm or incorrect treatment.

Under the five-reader evaluation, 97.5% (195/200) of the translated reports were rated Level 5 by all readers (clinically accurate, no edits required). The remaining 2.5% (5/200) contained at least one Level 4 rating (minor inaccuracies without clinical impact). No report received any Level 1-3 rating. To quantify expert agreement while accounting for the highly skewed rating distribution, we computed a Gwet's AC1 statistic of 0.977.[52,53] According to standard benchmarks, this represents almost perfect inter-rater agreement, confirming the high clinical reliability of the translated institutional dataset.

## S1.2 Dataset composition and curation framework of BrainTumor48K

To further enhance clarity and provide a comprehensive understanding of our dataset composition and processing, we have developed a detailed dataset organization pipeline. As illustrated in Supplementary Figure 2, BrainTumor48K comprises 24,716 2D MRI cases and 23,231 3D MRI cases. The 2D MRI data includes image slices paired with captions or labels from online repositories; these are used primarily for representation learning and not for clinical supervision.

Of the 23,231 3D MRI cases in our dataset, 18,113 are paired with both radiology reports and pathologically confirmed diagnostic labels, while the remaining 5,118 consist of 3D MRI scans and labels exclusively for classification tasks. The 18,113 cases are paired with radiology reports and pathologically confirmed diagnosis labels. Of these, 11,156 cases originate from our institutional archives, and 2,606 cases are sourced from online *Radiopaedia*, all accompanied by expert-curated radiology reports. Collectively, these 13,762 cases are characterized by high-quality imaging, professional radiology reports, and verified diagnostic labels, forming the backbone of our model's clinical validity.

For the remaining 4,351 3D MRI-report cases, their reports were generated with the assistance of large language models (LLMs). Importantly, this AI-assisted generation was not free-form writing: it was strictly confined to converting existing patient metadata (e.g., tumor type, tumor location, and, where available, signal intensity) into a structured radiology-report format. No new clinical information was introduced at any stage. The process only transformed or standardized information already present in the original dataset (detailed in Supplementary Section S1.1.1 and illustrated in Supplementary Figure 1a).

In summary, 13,762 3D MRI-report pairs (11,156 institutional + 2,606 expert public) are derived from purely professional sources, forming a robust, high-quality foundation. The remaining 4,351 pairs are structured, metadata-derived reports that contain no fabricated information.

## S1.3 Ethics approval for datasets from medical institutions

All data obtained from participating private medical institutions were obtained with ethics approval, including the Medical Ethics Review Committee of Xiangya Hospital Central South University (2025030464), the Medical Ethics Committee of Changde Hospital, Xiangya School of Medicine, Central South University (The First People's Hospital of Changde City) (2025-290-01), the ShanghaiTongji Hospital (Tongji Hospital of Tongji University) Ethics Committee (K-W-2025-008), the University of South China Second Hospital Clinical Research Ethics Review Committee (2025057), the Shenzhen Second People's Hospital Clinical Research Ethics Committee (2025-748-01PJ), the Third Xiangya Hospital Central South University Ethics Committee (25485), the Jiangxi Provincial People's Hospital Medical Ethics Committee (2025(100)), the Chongqing Traditional Chinese Medicine Hospital Medical Ethics Committee (2025-IIT-KS-19), the The First Affiliated Hospital of Nanchang University Medical Ethics Committee (IIT[2025]-787), the Hunan Provincial Children's Hospital Ethics Review Committee (HCHLL-2025-249), the First Affiliated Hospital of Lanzhou University Clinical Research (Pharmaceuticals, Devices) Ethics Committee (LDYYLL2025-2038), and the Second Affiliated Hospital of Nanchang University Biomedical Research Ethics Committee (O-[2025]-233). The clinical trial protocol is registered under the identifier NCT07126821. All data were deidentified in compliance with institutional policies and ethical standards

## S1.4 Patient and public involvement

Patients and the public were not involved in the design, conduct, reporting, or dissemination of this research. This observational study used only anonymized data from brain tumor patients, with no direct patient or public involvement in either the retrospective or prospective components. Patient and public involvement will be considered in future clinical trials and during further clinical development of the diagnostic model.

## S1.5 Estimation of total subject count from online resources

Accurately estimating subject-level statistics from the aggregated 39 online resources is challenging due to the inconsistent presence or absence of unique patient identifiers. For resources that do provide explicit patient identifiers, such as TCIA and BraTS, patient counts were directly extracted. In contrast, for web-crawled resources like PubMed Central[42] and ImageCLEF,[54] we conservatively assigned one unique patient per composite figure. This methodology likely results in an underestimate, as subfigures within composites may pertain to distinct cases, and images from the same video may originate from different patients. For certain directly downloaded datasets, including the Br35H,[21] Brain Tumor Classification Dataset,[24] and Brain Tumor MRI Images 44 Classes, which lack patient identifiers, we excluded these from our subject count estimation. The detailed number of patients and their associated multi-modal data in each online resource are illustrated in Supplementary Table 2.

## S2 Model Development

### S2.1 Development of BrainVLM for tumor classification and report generation

Developing an AI model for early brain tumor diagnosis necessitates the effective integration of presurgical multimodal data, including multi-parametric MRI scans, patient metadata, and radiology reports. Current AI-based brain diagnostic approaches largely rely on the subjective interpretation of pure MRI data, sometimes supplemented with patient metadata.[55,56] This neglects the rich, complementary information available in multi-modal datasets, potentially limiting the model's performance. Besides, the current vision language foundation models (VLMs[57]) face challenges as the input length increases with the quantity and resolution of images. This is particularly alarming in brain tumor studies, which require the analysis of high-resolution (HR), multi-parametric MRI scans (T1, T1c, T2, T2-Flair) from various views (axial, coronal, sagittal).

To address this, we developed BrainVLM, a first-of-its-kind VLM that effectively synergizes multi-modal medical visual and textual information. It tackles the challenges through optimization across three aspects: (1) Innovative multi-modal model architecture: Incorporating anatomical detail-aware vision encoder and hybrid causal-full attention strategies into a multi-modal network to handle complex multi-modal data efficiently. (2) Clinical-motivated data standardization: Developing a standardized dataset processing protocol that emulates clinical practices and incorporates special markers to differentiate different visual data. (3) Progressive three-stage training strategy: Using a progressive training approach to incrementally enhance the model's multi-modal processing capabilities.

#### S2.1.1 Multi-modal model architecture

As shown in Fig. 1b, BrainVLM consists of three core components: (1) an anatomical detail-aware vision encoder for efficient, fine-grained image feature extraction, (2) a 2-layer Multilayer Perceptron (MLP) projector that bridges the image encoder with the LLM, and (3) a hybrid causal-full attention-based LLM for reasoning and understanding the multi-modal information.

**Anatomical detail-aware vision encoder.** We employed the vision backbone of BioMedCLIP[58] as the basic visual encoder E. This model constitutes a Vision Transformer (ViT) pre-trained on a large-scale 2D medical imaging dataset using vision-language contrastive learning.[59] The standard ViT of original BioMedCLIP accepts only $224 \times 224$ input dimensions and cannot directly process high-resolution (HR) MRI slices (typically 400–500 pixels in width/height). Direct downsampling risks losing critical diagnostic details such as tumor margins and signal heterogeneity, while splitting the image into several sub-images for individual encoding results in excessive visual tokens for LLMs. This is computationally prohibitive due to the transformer's quadratic complexity. We therefore designed an anatomical detail-aware vision encoder to capture finer anatomical details while maintaining computational efficiency.

Our designed visual encoder incorporated two key components: mixed-resolution visual encoding and attention-based HR injection. Let I denote an input MRI scan with dimensions $N \times 3 \times w_o \times h_o$, where $N$ represents the number of slices (e.g., $N = 1$ for single-slice MRI; $N = 32$ for a 3D MRI volume with 32 MRI slices), and $w_o \times h_o$ specifies the original spatial resolution. We first generated two resized versions: 1) a low-resolution (LR) input $I^l$ scaled down to $N \times 3 \times 224 \times 224$, and 2) an HR input $I^h$ scaled up to size $N \times 3 \times 448 \times 448$. The LR input was encoded through the visual encoder $E$ to generate global LR tokens $z^l = E(I^l) \in \mathbb{R}^{N\times 768}$. Here $z^l$ is extracted from the final layer's [CLS] token of the ViT, capturing global patterns.

Next, $I^h$ was partitioned into four non-overlapping HR sub-inputs $\{I_k^h\}_{k=1}^4$ using a grid method,[60] where $k$ indexes the HR sub-inputs. Each HR sub-input was resized to $N \times 3 \times 224 \times 224$ and independently encoded by $E$ to generate HR visual tokens as $z^{h_k} = E(I_k^h)$. Unlike the global [CLS] token extraction used for the LR input, we retained all spatial patch tokens (excluding [CLS]) from the ViT's final layer when processing HR sub-inputs. This yields features $z^{h_k} \in \mathbb{R}^{N\times 196\times 768}$ that encode fine-grained anatomical details. The final combined HR representation is denoted as $z^h = \{z^{h_k}\}_{k=1}^4 \in \mathbb{R}^{4\times N\times 196\times 768}$.

Finally, we injected $z^h$ into $z^l$ via an LR-HR cross-attention module, to obtain HR-informed LR tokens $z^{hl}$. Here, $z^l$ served as query $Q$, $\{z^{h_k}\}_{k=1}^4$ served as the context (key $K$ and value $V$), resulting the final $z^{hl} \in \mathbb{R}^{N\times 768}$. This approach merged global context $z^l$ with anatomical details from $z^h$, while maintaining computational efficiency.

For multi-parametric MRI based brain tumor diagnosis (e.g., T1, T1c, T2, T2-Flair), each 3D MRI sequence was processed independently by our encoder to extract modality-specific features, denoted as $z_m^{hl} \in \mathbb{R}^{N\times 768}$. These features were then concatenated along the slice dimension: $z^{\text{comb}} = \|_{m=1}^{M} z_m^{hl} \in \mathbb{R}^{(M\cdot N)\times 768}$,. where $M$ denotes the number of MRI sequences. For example, a 5-sequence protocol (e.g., T1 axial, T1c axial, T1c sagittal, T2 axial, T2-Flair axial) with $N = 32$ slices per sequence yielded a combined representation $z^{\text{comb}} \in \mathbb{R}^{160\times 768}$, integrating complementary diagnostic information across sequences.

**MLP projector.** The combined feature representation $z^{\mathrm{comb}}$ was mapped into the LLM language space via a 2-layer MLP projector, converting visual information into a format digestible for LLM. This transformation bridged the modality gap by converting anatomical visual patterns into a sequential token format compatible with linguistic processing.

**Hybrid causal-full attention based LLM.** The MLP-projected visual embeddings were combined with patient metadata (e.g., age and sex) and textual instructions to form a multimodal prompt. This prompt was then processed by our adapted LLM for tumor classification, uncertainty quantification, and report generation. We utilized the pre-trained Llama-3.1-8B-Instruct[61] model as our base LLM, which provides a mainstream language backbone with grouped query attention and open-source adaptability. Traditional causal attention in Llama-3.1-8B-Instruct restricts tokens to preceding positions only, which is suboptimal for MRI interpretation.[62] Therefore, we implemented a hybrid attention strategy following previous works.[63,64] Specifically, full attention was applied to image tokens for capturing complex visual interdependencies within the same 3D MRI sequence or 2D MRI slice, while causal attention was retained for sequential tokens to preserve data structure. This hybrid architecture enabled the model to simultaneously capture cross-modal visual relationships and maintain linguistic coherence.

### S2.1.2 Clinical-motivated data standardization protocol

Language data in LLM is typically uniform and consistent, while our task involves more complex visual data. Randomly feeding unprocessed neuroimaging inputs from different modalities and views into BrainVLM complicates the learning process and impedes model convergence. Thus, we proposed a standardized dataset processing protocol that emulates clinical practices through the following three key strategies: 1) volumetric slice-count harmonization to ensure uniform slice dimensions, 2) clinical MRI sequence integration mirroring radiological workflows, and 3) special markers to differentiate multi-modal data and task-driven prompt design to differentiate different tasks. This protocol reduced learning ambiguity while preserving diagnostically critical features essential for brain tumor analysis.

**MRI preprocessing via volumetric slice-count harmonization.** The 3D MRI sequences used in our study typically exhibit substantial variability in the number of acquired slices. This variation occurs both across scanners/institutions (ranging from dozens to hundreds of slices per sequence) and between sequences for the same subject (e.g., 20 slices in native T1 vs. 40 in T1c sequences). This substantial slice variation significantly complicates model training. We therefore standardized each 3D MRI sequence to 32 slices using cubic spline interpolation (implemented via SimpleITK[65]), ensuring smooth volumetric resampling along the slice direction. Critically, unlike conventional neuroimaging pipelines that typically involve extensive preprocessing—such as resampling to isotropic voxels and skull stripping — BrainVLM performs only this volumetric slice-count harmonization. This minimal preprocessing preserves the complete information content of the original MRI data, making it particularly well-suited for our comprehensive brain tumor characterization study. While conventional skull-stripping may aid certain tumor analyses, it is not universally appropriate. Crucially, lesions located in regions like the pia mater or invading the cranial vault can be inadvertently removed during skull stripping, leading to the loss of critical anatomical and pathological information – a risk our approach deliberately avoids (Supplementary Figure 3).

**Clinical MRI sequence integration.** Emulating the clinical workflow for MRI-based tumor identification is critical for aligning machine learning models with diagnostic practice, thereby improving lesion identification and characterization.[66,67] In clinical practice, comprehensive brain tumor assessment often initiates with acquiring axial T1 and T2 sequences, *optionally* complemented by axial T2-Flair imaging. Following intravenous administration of gadolinium-based contrast agents, multiple-plane T1c (axial, sagittal, and coronal views) are obtained to evaluate enhancement patterns and blood-brain barrier disruption. Based on initial findings or clinical suspicion, additional sequences or planes (e.g., coronal or sagittal T1/T2 for sellar region lesions) may be acquired. During interpretation, neuroradiologists typically begin by examining T2 or T2-Flair images to identify potential lesions. They then use T1c images to detect small lesions and assess tumor enhancement, comparing these with T1 images of the same view for confirmation.

To emulate this clinical workflow and streamline model learning, we organized the MRI data to reflect both the clinical acquisition and interpretation workflows. Specifically, our integration strategy prioritizes clinically meaningful comparisons: the primary input consisted of paired native T1 and T1c scans acquired in the same plane, facilitating direct assessment of lesion enhancement. This core pairing was supplemented by representative T2 and T2-Flair scans, as well as an additional T1c scan obtained from a different imaging plane than the primary T1/T1c pair. For example, if the primary T1/T1c pair was axial, we included an additional T1c scan from the coronal or sagittal plane to provide multiplanar context. When specific sequences or planes were unavailable, alternative available sequences were included to maximize diagnostic coverage. In summary, BrainVLM utilized five core 3D MRI sequences as visual input—T1 (axial or another view), T1c from the same view as T1, T2 (axial or another view), T2-Flair (axial or another view), and an additional T1c from a different view than T1. This structured yet flexible approach maintained comprehensive diagnostic information while faithfully reflecting real-world clinical reasoning. Further details regarding the processing of cases with more than five MRI sequences are provided in the following Model Inference Section S2.3.

**Special tokens for data type differentiation and task-driven prompt design.** BrainVLM processed heterogeneous inputs spanning 2D MRI slices, 3D MRI volumes, and free-text patient metadata (e.g., age, sex), while concurrently addressing distinct task types including classification, report generation, and reliability assessment. To enable the model to distinguish between these data types and tasks, we implemented the following specialized tokens and task-driven prompts. Specifically, to help the model distinguish different data types, we designed special tokens: 2D single image slices were enclosed with '<image>' and '</image>', while 3D MRI volumes used '<vid>' and '</vid>' tags. A special token '<s>' was inserted between volumes to indicate the different sequence's input. We also introduced specialized tumor tokens (e.g., '<class_0>' for MET) to encode the original tumor names for classification. Concurrently, to enhance the model's task-specific learning capabilities, we designed customized prompts that explicitly inform the model of the input structure and expected objectives (see Supplementary Table 9). Specifically, the finalized prompt that was input into BrainVLM integrates three components: (1) predefined special visual tokens (e.g., <vid>, <image>) to structure the imaging inputs; (2) patient metadata; and (3) task-specific instructions that explicitly define the model's objective for each task (e.g., generating a clinical description for patients in the radiology report generation task or making a self-assessment confidence score for the reliability task). These special tokens and unified prompt design ensure robust alignment between input interpretation and clinical task requirements, enhancing the model's interpretative accuracy and adaptability to diverse diagnostic objectives.

**Unified task-aware tokenization and autoregressive decoding mechanism.** BrainVLM is designed to perform simultaneous multi-task generation within a single autoregressive pass. The following section details the step-by-step decoding mechanism enabled by our LoRA fine-tuned architecture. The complete data flow, encompassing input processing, tokenization, and sequential output generation, is visualized in Supplementary Figure 4. BrainVLM receives multi-modal input, including multi-parametric MRI scans, patient metadata, and an instruction prompt. MRI scans are processed by a shared vision encoder, and the extracted visual features are then projected into a sequence of visual tokens, aligning them with the LLM's token embedding space. Concurrently, patient metadata and instructional prompts (e.g., "Generate report, diagnosis, and uncertainty for this MRI.") are tokenized into discrete tokens using a Byte-Pair-Encoding (BPE) tokenizer. These visual and textual tokens are then concatenated and fed into the LLM decoder (fine-tuned with LoRA) for autoregressive generation. The LoRA fine-tuned LLM generates output in an autoregressive manner, guided by special delimiter and class tokens that structure the generation process. Specifically, to enable the simultaneous output of three distinct components (i.e., report, diagnosis, and uncertainty), we introduce three special start tokens into the vocabulary: <Report>, <Diagnosis>, and <Uncertainty>. During autoregressive generation, the model generates tokens sequentially. It first produces the radiology report tokens after <\Report>. These report tokens are standard BPE sub-word units that, once the report section is complete, are later decoded into a human-readable report via the BPE tokenizer.

After the report is complete, the model encounters <\Report> and transitions to generating the diagnosis. For diagnosis, we added 14 specialized classification tokens (e.g., <class_0> through <class_13>) to the vocabulary, each corresponding to one of the 11 WHO CNS5 tumor grades or to one of the three gliomas, glioneuronal tumors, and neuronal tumors (GGN) subtypes. When the model outputs a token like <class_5>, it is then directly mapped to the corresponding tumor label (e.g., "MEN" for meningioma) via a simple lookup table.

Following the diagnosis token, the model reaches <Uncertainty> and generates the confidence score as a sequence of numerical BPE tokens (e.g., "0", ".", "9"). Specifically, during the training, the confidence score was discretized into six levels (e.g., 50%, 60%, 70%, 80%, 90%, 95%) to simplify learning; at inference, the model outputs the appropriate token(s) that correspond to one of these levels. These tokens are concatenated and decoded into a human-readable confidence score (e.g., 0.9). Thus, all three components are generated in a single autoregressive pass. The BPE tokens for the report are decoded into free-text, the specialized classification token is mapped directly to a diagnosis label, and the numerical tokens are converted to a confidence score.

**S2.1.3 Three-stage progressive training strategy of BrainVLM**

Our curated BrainTumor48K dataset comprises three distinct data types: 53.14% 2D MRI slices paired with text descriptions, 38.14% multi-parametric 3D MRI volumes with radiology reports, and 8.72% 3D MRI sequences annotated solely with diagnostic labels. Direct end-to-end training on this heterogeneous data composition may overwhelm models with conflicting objectives. The 2D isolated MRI slices lack volumetric context essential for tumor characterization, while complete volumetric inputs demand foundational slice-level recognition capabilities that 2D training establishes. To address these challenges, we implemented a curriculum learning paradigm with three progressive training stages. This strategy enables the model to progressively interpret complex visual patterns, starting from simple 2D slices analysis and advancing to more intricate 3D volumetric data interpretation (see Supplementary Figure 5a).

**Training stage 1: 2D MRI slice-text representation learning.** Volumetric tumor analysis fundamentally requires slice-level recognition capabilities, a skill radiologists cultivate through detailed examination of individual-MRI slices. To establish this foundation while maximizing BrainTumor48K's heterogeneous resources, we initiated BrainVLM's training with a focus on 2D MRI slice-level representation learning. During this phase, we optimized only the MLP projector and the LR-HR cross-attention module within our anatomical detail-aware encoder, keeping the vision encoder and LLM module frozen to maintain stability. This stage focused on training 2D MRI slice-text alignment and utilized two main sources: (i) all available 2D MRI slice-text pairs in BrainTumor48K, and (ii) strategically extracted tumor slices from 3D MRI volumetric sequences, paired with AI-generated slice-level text descriptions. For public datasets with tumor masks, such as BraTS, we extracted slices with more than 5% tumor occupancy and generated clinically validated text descriptions using RAG, refined by ChatGPT-4.0. For institutional scans, neuroradiologists manually selected pathologically significant slices from each 3D MRI sequence, and the corresponding slice-level reports were crafted using RAG refined by ChatGPT-4.0. Finally, this meticulous process resulted in 10,996,702 curated 2D MRI slice-text pairs. This 2D MRI slice-level representation learning approach enabled BrainVLM to establish fundamental tumor recognition capabilities, laying a solid foundation for subsequent advanced 3D volumetric analysis.

**Training stage 2: Volumetric context integration via hybrid 2D-3D training.** In this phase, we aimed to transfer the learned 2D representation capabilities into the tomographic 3D MRI data space, facilitating BrainVLM's 3D tumor characterization capability. We utilized all available 2D and 3D training datasets from BrainTumor48K and designed a dynamic learning strategy interleaving 2D slice-level and 3D volumetric tasks during training. Crucially, at each iteration, we stochastically sampled training instances with 30% probability allocated to 2D tasks (leveraging isolated slices for tumor description and slice-level diagnosis) and 70% probability to 3D tasks (utilizing full sequences for volumetric report generation, classification, and diagnosis). During this stage, the MLP projector, LR-HR cross attention module, and Low-Rank Adaptation (LoRA)[68] parameters in the LLM model were trained, while the vision encoder and other parameters in the LLM model remained frozen. This transitional stage ensures that the model develops a cohesive understanding of tumor characteristics across different dimensions.

**Training stage 3: Comprehensive 3D volumetric instruction tuning.** In the final stage, we focused on leveraging the full high-quality 3D data available in the BrainTumor48K dataset to enhance BrainVLM's advanced volumetric analysis capabilities. The projector, LR-HR cross-attention module and LoRA parameters in the LLM model were trained, while all other parameters remained frozen.

### S2.1.4 Loss function

We introduced dual-objective losses to unleash BrainVLM's generative and discriminative capabilities across clinical tasks, including free-text generation and dedicated tumor classification. Specifically, we first employed a standard causal language modeling loss $L_{\mathrm{LM}}$, which enables the model to follow natural language instructions and generate coherent diagnostic texts. Given a token sequence ( $w_i^{(1)}$, $w_i^{(2)},\ldots,w_i^{(T)}$) in a training sample $x_i$, this loss is formulated as:

$$L_{\mathrm{LM}} = -\sum_{t=1}^{T} \log P(w_i^{(t)} \mid w_i^{(1)}, w_i^{(2)}, \ldots, w_i^{(t-1)}; \theta),$$

where $P(w_i^{(t)} \mid w_i^{(1)}, w_i^{(2)}, \ldots, w_i^{(t-1)}; \theta)$ denotes the predicted probability of token $w_t$ given the preceding tokens. $\theta$ denotes the model parameters. $t = 1,\ldots,T$ denotes the tokens in the sequence. This objective encourages contextual understanding and fluent text generation by minimizing the negative log-likelihood of ground-truth tokens.

Furthermore, to enhance the tumor-specific discrimination, we introduced an auxiliary classification loss $L_{\mathrm{CL}}$ to perceive fine-grained medical patterns. We added special tokens to encode the original tumor label (such as <tumor_3> token for Glioma), and employed a classification loss based on these tokens. This loss is computed as:

$$L_{\mathrm{CL}} = -\sum_{c=1}^{14} y_c^{(i)} \log \hat{y}_c^{(i)},$$

where $y_c^{(i)}$ denotes the ground-truth label for the c-th tumor class, and $\hat{y}_c^{(i)}$ is the predicted probability. This loss enables fine-grained recognition across 14 tumor categories, including 11 major brain tumor types in WHO-CNS5 and 3 GNN subcategories.

The overall objective combines both losses:

$$L_{\text{total}} = L_{\text{LM}} + \alpha L_{\text{CL}},$$

where α is a tunable coefficient balancing language generation and tumor classification. This joint optimization framework enables BrainVLM to effectively capture both general linguistic semantics and task-specific diagnostic cues.

#### S2.1.5 Data augmentation

Brain MRI data exhibited domain shifts driven by patient-specific factors (e.g., age, sex, physique) and MRI hardware variability,[69] resulting in significant intra-modality variation (Supplementary Figure 6). We applied tailored augmentation strategies for both 2D and 3D MRI data to improve BrainVLM's generalization across diverse clinical settings. For 2D MRI slices, we adopted standard augmentation techniques,[70] including random flipping, cropping, and intensity transformations. For 3D MRI sequences, we first identified key challenges such as noise artifacts, intensity inhomogeneity, and motion-induced blurring (examples shown in Supplementary Figure 6). Then, to improve robustness against these effects, each 3D training sample was augmented with: (1) random rotation within ±60 degrees, and (2) intensity variation, motion blur, or random noise, each applied with 30% probability.

### S2.2 Model reliability implement

Brain tumor diagnosis is a high-stakes task, where a misdiagnosis can lead to inappropriate treatments, prolonged patient suffering, and potentially fatal outcomes. However, conventional VLMs may generate erroneous predictions with high confidence,[71] undermining clinical trust and limiting practical utility. As illustrated in the left panel of Fig. 2c in the main manuscript, our vanilla BrainVLM model (without reliability mechanisms) also exhibited this concerning behavior, frequently assigning high confidence scores to erroneous predictions. To address this critical limitation, we presented an uncertainty quantification strategy that calibrates prediction confidence in BrainVLM: accurate diagnoses receive low uncertainty scores (high confidence score), while erroneous predictions are assigned high uncertainty (low confidence score). This ensures only high-confidence, reliable predictions are acted upon.

Existing uncertainty quantification approaches for VLMs and LLMs rely on multiple sampling strategies (e.g., Best-of-N) that incur substantial computational overhead regardless of question complexity.[72] While adaptive sampling techniques attempt to reduce costs through early-stopping heuristics, they remain constrained by manually designed rules that lack generalizability across models and tasks.[73] We introduce a consensus-driven confidence calibration strategy that directly transforms model disagreement into calibrated confidence signals, eliminating the need for heuristic rules or fixed sampling costs. Our approach features two key innovations: (1) Consensus-driven reliability dataset construction, where a reliability dataset is automatically constructed using agreement statistics from multiple model instances via consensus-driven clustering; (2) Confidence-aware reliable BrainVLM finetuning: Distillation of consensus-derived reliability knowledge into our BrainVLM framework via parameter-efficient fine-tuning. We refer to the baseline BrainVLM without uncertainty quantification as the vanilla BrainVLM, while our enhanced version incorporating the proposed uncertainty quantification strategy is termed reliable BrainVLM.

**S2.2.1 Consensus-driven reliability dataset construction**

To create a high-quality reliability dataset for confidence calibration, we leveraged the natural training dynamics of BrainVLM. Specifically, when our best-performing model was near convergence, we sampled $N$ model checkpoints at different training iterations to create an ensemble of $N$ vanilla BrainVLM variants. For each training sample $x_i$ with pathological label $y_{\text{true}}^{(i)}$, we independently processed $x_i$ through all $N$ models to obtain $N$ diagnostic predictions $\{\hat{y}_i^{(1)}, \ldots, \hat{y}_i^{(N)}\}$ per sample (see Supplementary Figure 5b). These responses were then clustered by predicted diagnoses, with the largest cluster defining the consensus label $y_{\text{cons}}^{(i)}$. Formally, the consensus label is computed as:

$$y_{\text{cons}}^{(i)} = \underset{c \in C}{argmax} \sum_{k=1}^{N} \mathbb{I}(\hat{y}_i^{(k)} = c),$$

where $\boldsymbol{C}$ denotes the set of all possible diagnostic classes and $\mathbb{I}$ indicates the indicator function. This consensus label represents the prediction with the strongest inter-model agreement. We then computed a confidence score $s_{\text{con}}^{(i)}$ for each sample $x_i$:

$$s_{\text{con}}^{(i)} = \frac{N - N_w^{(i)}}{N},$$

where $N_w^{(i)} = \sum_{k=1}^{N} \mathbb{I}(\hat{y}_i^{(k)} \neq y_{\text{true}}^{(i)})$, which indicates the number of incorrect predictions among the $N$ models. A high $N_w$ implies greater disagreement, resulting in a lower confidence score, while lower $N_w$ yields higher confidence scores.

For each sample $(x_i, y_{\text{true}}^{(i)})$, the above process yields per-sample supervision tuples

$(x_i, y_{\text{cons}}^{(i)}, s_{\text{con}}^{(i)})$, where $y_{\text{cons}}^{(i)}$ represents the ensemble-aggregated prediction, and $s_{\text{con}}^{(i)}$ quantifies reliability based on inter-model consensus strength $y_{\text{cons}}^{(i)}$ and clinical validity alignment with $y_{\text{true}}^{(i)}$. Collectively, these tuples constitute the reliability dataset, which provides reliable supervision for training BrainVLM to generate well-calibrated, confidence-aware predictions.

**S2.2.2 Confidence-aware reliable BrainVLM finetuning**

After the construction of the reliability dataset, we then distilled the consensus-derived reliability knowledge into our reliable BrainVLM through confidence-aware reliable fine-tuning. Specifically, as illustrated in Supplementary Figure 5b, we transformed the vanilla BrainVLM into reliable BrainVLM by parameter-efficient fine-tuning on our constructed reliability dataset. In this process, we also reformulated the output scheme to generate an additional confidence score, yielding responses such as: "Report:..., diagnosis: <tumor_i>. My confidence: 80%". Crucially, we employed our curated tuples $(x_i, y_{\text{cons}}^{(i)}, s_{\text{con}}^{(i)})$ rather than ground-truth pairs $(x_i, y_{\text{true}}^{(i)})$ as supervision signals, enabling simultaneous learning of diagnostic predictions and their associated confidence levels. During this fine-tuning process, we updated only the projector, LR-HR cross-attention module, and LoRA parameters within the LLaMA model, keeping both the vision encoder and LLaMA backbone frozen. The model was trained for two epochs on the reliability-annotated dataset.

## S2.3 Model inference

Let $S = \{\text{T1}, \text{T1c}, \text{T2}, \text{FLAIR}\}$ denote the set of MRI sequences. For each sequence $s \in S$, let $V_s \subseteq \{\text{axial}, \text{sagittal}, \text{coronal}\}$ represent the available views for $s$ ; $V_s$ may be empty if the corresponding sequence is missing. Given a patient, the complete input MRI space can be formulated as:

$$X = \{(x_s^v, s, v) \mid s \in \mathcal{S}, v \in \mathcal{V}_s\}, \quad (1)$$

where $x_s^v$ denotes the MRI image corresponding to sequence $S$ and view $V$ . As described in Clinical MRI sequence integration Section S2.1.2), during the inference time, BrainVLMtakes five core MRI sequences—T1 (axial or another view), T1c from the same view as T1, T2, T2-Flair, and an additional T1c from a different view than T1—along with patient metadata and task-specific instructions as input. Formally, the five core MRI sequences $C_m$ satisfy:

$$C_m = \{(x_{\mathrm{T1}}^{v_0}, \mathrm{T1}, v_0), (x_{\mathrm{T1c}}^{v_0}, \mathrm{T1c}, v_0), (x_{s_3}^{v_3}, s_3, v_3), (x_{s_4}^{v_4}, s_4, v_4)\ (x_{s_5}^{v_5}, s_5, v_5)\}, \quad (2)$$

where $v_0, v_3, v_4, v_5 \in V_s$ are chosen from available views, and the selection follows these criteria:

- $v_0 \in \mathcal{V}_{\mathrm{T1}}$ is any available view of T1 (e.g., axial, sagittal, or coronal).
- The first two items are T1 and T1c from the same view $v_0$.
- For $s_4$ and $s_5$,select from {T2, T2-Flair} according to availability:
  - If both T2 and T2-Flair are available, then $s_3 = \mathrm{T2}$, $s_4$ =T2-Flair, with $v_3 \in \mathcal{V}_{\mathrm{T2}}$, $v_4 \in \mathcal{V}_{\mathrm{T2-Flair}}$.
  - If only one of T2 or T2-Flair is available, assign it to $s_3$ and select any other available MRI sequence for $s_4$.
  - If neither T2 nor T2-Flair is available, select any other available MRI sequences for $s_3$ and $s_4$.
- For $s_5$, prioritize selecting T1c from any view $v_5 \in \mathcal{V}_{\mathrm{T1c}}$ such that $v_5 \neq v_0$. If not available, select any unused view from the remaining available sequences.

This selection strategy dynamically adapts to the available sequences and views for each patient, ensuring robust performance even in the presence of missing sequences. During inference, BrainVLM takes the selected five core MRI sequences, along with patient metadata and task-specific instructions, as input. It then generates a diagnostic prediction, a radiology report, and an associated confidence score.

**Adaptive sequence sampling and aggregation.** In real-world clinical practice, MRI studies frequently employ thick slice acquisitions spanning multiple anatomical planes (e.g., axial, sagittal, and coronal views for both T1 and T2 sequences), yielding studies exceeding the standard five-sequence input. To address this variability, we define an adaptive sequence sampling strategy as follows.

Let $C$ be the set of all valid five-sequence combinations constructed from $X$ according to the above selection criteria:

$$C = \{C_m \subset X \ || C_m| = 5 \text{ and } C_m \text{ satisfies the selection constraints above}\}. \quad (3)$$

For each $C_m \in C$, we define the model inference function:

$$(y_m, r_m, s_{con}^m) = f_\theta(C_m, d, t), \quad (4)$$

where $f_\theta$ defines the reliable BrainVLM parameters, $d$ denotes patient metadata, $t$ denotes the task-specific prompt, $y_m$ is the predicted clinical diagnosis, $r_m$ is the generated radiology report, and $s_{con}^m$ is the confidence score. To aggregate results across all sampled combinations, we define the final clinical diagnosis $\hat{y}^{(1)}$ as the most frequent prediction ("majority voting") among all $\{y_m\}_{C_m} \in c$:

$$\hat{y}^{(1)} = \mathrm{mode}(\{\mathrm{y_m}\}_{m=1}^{M}), \quad (5)$$

where $\mathrm{mode}(\cdot)$ indicates the statistical mode (most frequent element) in the ensemble. Here $\hat{y}^{(1)}$ is also known as the TOP 1 prediction. The final confidence $\hat{s}_{con}^1$ that associated with the predicted diagnosis $\hat{y}^{(1)}$ can be aggregated as:

$$\hat{s}_{con}^{(1)} = \frac{1}{N_1}\sum_{m=1}^{M} s_{con}^m \cdot \delta(y_m, \hat{y}_1), \quad N_1 = \sum_{m=1}^{M} \delta(y_m, \hat{y}^1), \quad (6)$$

where $\delta(a,b) = 1$ if $a = b$ else 0. The final radiology report is generated by aggregating reports associated with the TOP 1 diagnosis. Specifically, since our report are structured into predefined sections (e.g., T1 signal characteristics), we apply majority voting within each section to select the most frequent content among all reports associated with the TOP 1 predicted diagnosis $r_m \mid y_m = \hat{y}^{(1)}$.

This adaptive ensemble strategy ensures that the model can robustly process heterogeneous and incomplete clinical MRI acquisitions by (1) systematically sampling all valid five-sequence input sets from the available MRI space $X$, (2) independently inferring on each combination, and (3) performing robust aggregation via majority voting to determine the final (Top 1) clinical prediction.

### S2.3.1 Confidence-triggered TOP 2 supplementary diagnosis

In clinical workflows, neuroradiologists may consider multiple diagnostic possibilities when interpreting clinically ambiguous cases, such as "considering meningioma or glioma". To emulate this process, we implement a confidence-triggered supplementary diagnosis mechanism. If the confidence score $\hat{s}_{con}^{(1)}$ for the Top 1 prediction $\hat{y}^{(1)}$ falls below a predefined threshold (set at 75% in this study), a secondary (top-2) diagnosis is also reported. The Top-2 diagnosis $\hat{y}^{(2)}$ is defined as: $\hat{y}^{(2)} = \text{mode}(\{y_m \mid y_m \neq \hat{y}_1\})$. Our experimental results in Supplementary Section S3.1 demonstrate the effectiveness of this approach in enhancing diagnostic reliability in ambiguous cases.

## S2.4 Finetuning of molecular subgroup classification task

In this phase, we fine-tuned BrainVLM using the stage-3 checkpoint, leveraging the adult-type diffuse glioma data from BrainTumor48K. We performed threefold validation on the Xiangya primary Dataset (n = 494; 333 IHD-w GBMs, 86 IDH-m Astros, 75 IDH-m ODGs), each fold consisted of 328 patients (222 IDH-w GBMs, 57 IDH-m Astros, 50 IDH-m ODGs) for training and 166 patients (111 IDH-w GBMs, 29 IDH-m Astros, 25 IDH-m ODGs) for validation. We followed the MRI sequences integration strategy defined in the Section S2.1.2 and S2.3 for molecular classification training and prediction. In this training stage, we updated only the projector, LR-HR cross-attention module and LoRA parameters within the LLaMA model, keeping other parameters frozen.

The results demonstrated BrainVLM's robust capability to differentiate molecular subtypes of adult diffuse gliomas. BrainVLM achieved a frequency-weighted *F1* score of 0.92 (95% CI: 0.91–0.93, Supplementary Figure 7c) in the threefold validation on the Xiangya primary dataset, outperforming Merlin (*F1* = 0.76, 95% CI: 0.75–0.77) and RadFM (*F1* = 0.74, 95% CI: 0.73–0.75). Across molecular subgroups, BrainVLM demonstrated superior performance: IDH-w GBMs (*F1* = 0.95 vs. 0.88–0.91), IDH-m Astros (*F1* = 0.89 vs. 0.53–0.62), and IDH-m ODGs (*F1* = 0.84 vs. 0.58–0.63). On external validation, BrainVLM achieved a frequency-weighted *F1* score of 0.88 (Supplementary Figure 7c), surpassing Merlin (*F1* = 0.73) and RadFM (*F1* = 0.69). Consistent with primary validation, BrainVLM showed higher performance across subgroups: IDH-w GBMs (*F1* = 0.94 vs. 0.82–0.85), IDH-m Astros (*F1* = 0.79 vs. 0.32–0.44), and IDH-m ODGs (*F1* = 0.69 vs. 0.27–0.34).

## S2.5 Competing methods

We compared BrainVLM with three state-of-the-art AI models—including two VLMs (RadFM[74] Merlin[75]) and the large language model ChatGPT-4o,[48] as well as a representative traditional deep learning model, Video Swin Transformer (VST).[76] RadFM utilizes a 12-layer 3D ViT as a vision encoder to process both 2D and 3D data, and uses MedLLaMA-13B as the large language model for reasoning and a tradtional deep learning method video-swin transformer. For our experiments, we fine-tuned the pretrained RadFM on our BrainTumor48K dataset, updating the LoRA parameters of MedLLaMA-13B and the projection module. Merlin is a 3D vision-language foundation model developed for abdominal CT interpretation. It was pretrained on both radiology reports and structured electronic health records (EHRs), using a 3D ResNet152 as the vision encoder and RadLlama-7B as the language model. The vision encoder was pretrained on a large-scale paired dataset of 3D CT scans and text reports using the InfoNCE loss. For our experiments, we fine-tuned Merlin's vision encoder based on its pretrained 3D ResNet152 weights, and applied LoRA-based fine-tuning to the RadLlama-7B language model. Both RadFM and Merlin are designed for single-modality radiology inputs. For a fair comparison, we adopted the same input strategy in BrainVLM as in these baseline models. Specifically, we provided five MRI sequences as input, processed each sequence through the respective vision encoders, and concatenated the extracted features before passing them to the language model. To ensure a rigorous and fair comparison, we extensively fine-tuned all baseline models on the BrainTumor48K dataset with sufficient hyperparameter optimization, training each model until convergence. For VST, we employed five dedicated modality-specific encoders for each modality required by BrainVLM (Five sequences, two T1c sequences in different planes, a non-contrast T1 matching one T1c plane, and T2 and T2f sequences from any available views). The high-dimensional features extracted from each encoder were concatenated and processed through a fully connected layer to produce the final 12-category diagnostic output.

ChatGPT-4o[48] does not natively support 3D medical imaging input. Therefore, for each test case, experienced clinicians manually selected the most representative tumor-containing slice from each MRI sequence (four to five sequences per case). These clinician-validated slices were combined and provided as a set of 2D images to ChatGPT-4o for report generation and diagnostic prediction. It is important to note that this approach simplifies the diagnostic task, as tumor-containing frames are pre-identified by experts, whereas BrainVLM, RadFM, and Merlin operate directly on the full 3D raw MRI volumes without manual selection.

Video Swin Transformer is a traditional deep learning model that has demonstrated strong performance in video understanding tasks and is widely used in medical imaging analysis.[77] We included VST as a non-VLM baseline to benchmark the performance of pure vision-based deep learning approaches. We implemented a VST baseline for brain tumor diagnosis by fine-tuning a pretrained video-swin transformer on the BrainTumor48K dataset. For VST, we employed five dedicated modality-specific encoders for each modality required by BrainVLM (i.e., two T1c sequences in different planes, a non-contrast T1, and T2 and T2f sequences). These high-dimensional features extracted from each encoder were then concatenated and processed through a fully connected layer to generate the final 14 category diagnostic output.

### S2.6 Quantitative assessment and statistical analysis

**Diagnosis statistical metrics.** We evaluated BrainVLM's tumor prediction performance using sensitivity, specificity, precision, Cohen's *Kappa*, *F1*, and Area Under the Curve (AUC). We used both the common language metrics (Bilingual Evaluation Understudy (BLEU)[78]) and clinical relevance metrics (RaTEScore[79] and RadGraph-XL[80]) to measure the BrainVLM's report generation performance. Given true-positive (TP), false-positive (FP), true-negative (TN) and false-negative (FN) rates, the precision, sensitivity (recall), specificity, and *F1* score of each class are calculated as follows:

$$Precision = \frac{TP}{TP + FP},$$

$$Sensitivity = \frac{TP}{TP + FN},$$

$$Specificity = \frac{TN}{TN + FP},$$

$$F_1\text{-score} = \frac{2 \cdot Precision \cdot Sensitivity}{Precision + Sensitivity},$$

$$Cohen's\ Kappa = \frac{P_o - P_e}{1 - P_e},$$

where

$$P_o = \frac{TP + TN}{TP + FP + TN + FN} \text{ and } P_e = \frac{(TP+FP)\times(TP+FN)}{(TP+FP+TN+FN)^2} + \frac{(TN+FN)\times(TN+FP)}{(TP+FP+TN+FN)^2}.$$

**Uncertainty metric.** To evaluate the reliability of BrainVLM's uncertainty predictions and quantify the alignment between predicted confidence and diagnostic accuracy, we employed Expected Calibration Error (ECE) and Brier Score.The ECE measures the weighted average difference between accuracy and confidence across $M$ bins:

$$ECE = \sum_{m=1}^{M} \frac{|B_m|}{N} | acc(B_m - conf(B_m) |$$

where $N$ is the total number of samples, while $acc(B_m)$ and $conf(B_m)$ represent the accuracy and average confidence within bin $B_m$. The Brier score further assesses the mean squared error between the predicted probability distribution and the ground truth:

$$BS = \frac{1}{N} \sum_{i=1}^{N} \sum_{k=1}^{K} (p_{i,k} - y_{i,k})^2 ,$$

where $K$ denotes the number of classes, $p_{i,k}$ the predicted probability, and $y_{i,k}$ the binary indicator for class $k$ . Lower values in both metrics indicate a model that more accurately reflects its own diagnostic certainty.

**Report metric.** BLEU-4 is a widely used metric for evaluating lexical similarity between machine-generated and human reference texts.[78] Unlike traditional natural language metrics, both RadGraph-XL and RaTEScore prioritize domain-specific content critical to radiology, making them particularly suitable for evaluating automated report generation systems. RaTEScore evaluates clinical reports by identifying medical entities across imaging modalities and regions and then calculates a clinical relevance-weighted similarity score. Similarly, RadGraph-XL is a metric designed to evaluate the quality of radiology reports by assessing their semantic and clinical accuracy. It constructs a graph-based representation of entities and relationships within a report, capturing anatomical findings, observations, and their interconnections, such as whether a finding is present, absent, or uncertain. By comparing these graphs between generated and reference reports, RadGraph-XL quantifies similarity using an F1 score, thereby providing a robust measure of how well a generated report conveys clinically relevant information.

### S2.7 Implementation details

We employed AdamW[81] with weight decay regularization ($\lambda = 0.05$) across all training stages, coupled with a cosine decay scheduler and 300-step linear warmup. During the training stage 1 (2D MRI-text representation learning), the learning rate was linearly scaled from $1\times10^{-6}$ to $1 \times 10^{-4}$, and the batch size was set to 256. For training stage 2 (hybrid 2D-3D training) and stage 3 (3D volumetric learning), the learning rate followed the same warmup initialization ($1\times10^{-6}$). However, the peak learning rate was set to $3 \times 10^{-5}$ for training stage 2 and $1 \times 10^{-5}$ for training stage 3, respectively. For the molecular classification finetuning, the peak learning rate was set to $1 \times 10^{-4}$. We set the batch size to 1 on each GPU device for both training stages 2 and 3, resulting in a total effective batch size of 4. All experiments and the implementation of the BrainVLM were conducted using Python (version 3.9). We also utilized torch (2.0.0), transformers (4.45.2), huggingface-hub (0.29.2), tokenizers (0.20.3) and peft (0.2.0) for model training. For image processing and data augmentation, scikit-image (0.24.0), SimpleITK (2.4.1), monai (1.4.0), numpy (1.26.4), opencv-python (4.7.0) and nibabel (5.3.2) were applied. For training stage 1, we used 8 NVIDIA 4090 GPUs for multi-GPU training. For the training stages 2 and 3, as well as the reliable training, we used 4 NVIDIA L40 GPUs.

# S3 More analytical experiments on BrainVLM

## S3.1 Top-2 prediction result based on uncertainty score

To mirror clinical workflows in clinically ambiguous cases, where neuroradiologists consider multiple diagnostic possibilities, we implemented a confidence-triggered supplementary diagnosis mechanism in BrainVLM (Supplementary Section S2.3.1, appendix p 20). When the confidence score for the primary diagnosis (Top 1) falls below a threshold (set as 75% in this study), the BrainVLM automatically presents the second-most probable diagnosis (Top 2) for radiologist review. On the primary dataset, this mechanism increased the Top-2 F1 score from 0.83 to 0.86 (Macro-AUC: 0.85 → 0.88), while performance on the external dataset improved from 0.76 to 0.78 (Macro-AUC: 0.80 → 0.81) (Supplementary Figures 9a and 10a). More specifically, BrainVLM surpassed neuroradiologists' Top 2 recommendations across 11 out of 12 tumor types in the primary dataset and 10 out of 12 tumor types in the external dataset, achieving comparable results in CPT but underperforming marginally in MNM (Supplementary Figure 9d).

## S3.2 Robustness of BrainVLM to missing MRI sequences

BrainVLM performs inference using four standard MRI sequences (T1, T1c, T2, and T2-Flair). In clinical practice, incomplete acquisition of MRI sequences (e.g., missing an entire sequence) frequently occurs due to insufficient clinician awareness or contraindications. To evaluate the model's robustness to such scenarios, we conducted an ablation study by systematically excluding one of the four standard sequences. For each excluded sequence, we maintained the required five-image input structure by randomly selecting an additional view from the remaining sequences. For example, when removing T1, we substituted an extra view from T1c, T2, or T2-Flair (e.g., supplementary sagittal T1c). For fair comparison, our analysis included only patients meeting the five-image requirement after sequence removal. This resulted in 2,701 patients in the primary test dataset and 870 patients in the external test dataset. As shown in Supplementary Table 10 and Supplementary Figure 7b, BrainVLM achieves diagnostic performance comparable to the full-sequence baseline in scenarios missing T2, T2-Flair, or T1 sequences, with less than a 5% overall reduction in F1 score. However, the unavailability of T1c sequences caused a 13.4% F1-score decline, underscoring its critical role in brain tumor assessment.

Furthermore, we employed Shapley value analysis[82] to evaluate the contribution of each input sequence (T1, T1c, T2, and T2-Flair) to the model's decisions. As shown in Supplementary Figure 7a, the analysis revealed consistent patterns with our ablation results: omission of the T1c sequence led to the most substantial decline in diagnostic performance across all tumor types, highlighting its critical importance. Additionally, EMB classification exhibited uniform sensitivity to the removal of any single sequence (approximately 35% mean performance drop), while GCN and MEN predictions remained stable when T2 or T2-Flair was missing. In contrast, CPT predictions were more dependent on T1, T1c and T2-Flair, but showed resilience to the exclusion of T2. These findings align well with clinical diagnostic experience, where specific sequences are known to be more informative for certain tumor types.

### S3.3 Subcategories classification performance in GGN

GGN collectively represent a vital and diverse category of brain tumors with significant heterogeneity in therapeutic approaches and clinical prognoses. These tumors could be stratified into three subcategories according to the WHO CNS5 classification, including gliomas, glioneuronal tumors, and ependymal tumors. To extend the clinical utility of BrainVLM, we further evaluated the diagnostic performance of the model among these three subtypes, benchmarking against both neuroradiologists and state-of-the-art deep learning models (RadFM, Merlin and VST) In this evaluation, the models were tasked with classifying brain tumors into 14 distinct categories, encompassing the 11 major WHO CNS5 tumor types and the three GGN subcategories. BrainVLM achieved F1 scores of 0.78 for gliomas, 0.43 for glioneuronal tumors, and 0.30 for ependymal tumors (Supplementary Figure 9b) throughout the entire test cohort, demonstrating substantial improvements over both neuroradiologists consensus scores (0.75, 0.07, 0.24, respectively) and baseline models (RadFM: (0.60, 0.05, 0.09); Merlin: (0.53, 0.07, 0.14) and VST: (0.52, 0.06, 0.07)). These results underscore BrainVLM's superior discriminative capability in fine-grained tumor categorization.

### S3.4 Validation of generated report descriptions

In the main text, we quantitatively evaluated BrainVLM's report generation performance using standard n-gram metrics (e.g., BLEU-4) and entity-level clinical metrics (RaTEScore, RadGraph-XL). These rigorous automated benchmarks effectively measure lexical similarity and the preservation of key clinical entities. To further augment this evaluation and assess the generated reports from a semantic reasoning perspective akin to human experts, we introduced an "LLM-as-a-judge" framework to specifically evaluate two clinically critical aspects: (1) the accuracy of signal characteristics across multi-parametric MRI sequences (T1, T1c, T2, T2-Flair), and (2) the precision of tumor localization descriptions. Following previous work,[83] we categorized both intensity and location descriptions into three levels: correct, partially correct, and incorrect, assigning scores of 1.0, 0.5, and 0.0, respectively. For intensity descriptions, partially correct refers to cases where, for example, a prediction of "hyperintense" for an actual "hyperintense-isointense" lesion. For location descriptions, partially correct refers to cases where a prediction of the "Fourth Ventricle" for a lesion located in the "Cerebellum". To facilitate robust and unbiased matching and judgment for this detailed analysis, we leveraged the strong reasoning capabilities of the open-source large language model, Qwen-2.5-72B.[84] As shown in Supplementary Figure 7d, BrainVLM consistently outperformed baseline models across all evaluated dimensions based on the LLM-judge assessments. Specifically, in characterizing Contrast Enhancement (CE), a critical feature for tumor grading, BrainVLM achieved a high correctness rate of 0.80 and an average score of 0.86, surpassing both Merlin (Correct = 0.72, Avg. Score=0.78) and RadFM (Correct = 0.70, Avg. Score = 0.76). Similarly, for intrinsic signal intensity patterns on T1, T2, and T2-Flair sequences, BrainVLM maintained superior accuracy (Correct rates: T1 = 0.71, T2 = 0.72, T2-Flair = 0.74) with consistently higher average scores compared to Merlin and RadFM. Furthermore, for the challenging task of tumor localization (lesion), BrainVLM achieved a correct identification rate of 0.60 with an average score of 0.72, significantly outperforming the best baseline model Merlin (Correct = 0.51, Avg. Score = 0.62) and RadFM (Correct = 0.47, Avg. Score = 0.57). These results confirm that BrainVLM generates radiology reports with higher clinical fidelity and significantly lower hallucination rates than current state-of-the-art comparators.

## S3.5 Expert-based clinical assessment of generated reports

While the LLM-as-a-judge framework (Section S3.4) and standard automated metrics (e.g., BLEU-4, RaTEScore, RadGraph-XL) provide rigorous quantitative evaluation of generated reports, they cannot fully capture the holistic clinical judgment of how faithfully a generated report reflects what an experienced neuroradiologist would consider an accurate description of a patient's imaging findings. To address this gap and assess the generated reports from a clinical-judgment perspective, we conducted an additional expert-based assessment to directly evaluate the clinical fidelity of BrainVLM-generated reports relative to existing comparators (Merlin, RadFM, and GPT-4o).

Specifically, we randomly selected 100 patient cases from the primary test cohort, covering a broad range of WHO CNS5 tumor types. For each case, four reports were generated by BrainVLM, Merlin, RadFM, and GPT-4o. The reports were anonymized and randomly shuffled to remove any identifying information about the source model, after which three expert neuroradiologists independently scored each report on a 0–5 Likert scale based on consistency with the ground-truth clinical report. The scoring rubric defined 0 as completely inconsistent (no clinically relevant overlap) and 5 as fully consistent (semantically equivalent to the ground-truth report), with intermediate levels graded in 1-point increments. The three raters were blinded to which model produced each report to minimize evaluator bias.

As shown in Supplementary Figure 8, BrainVLM consistently received the highest score from all three raters, with mean scores of 4.77, 4.73, and 4.83 (Raters 1–3, respectively). Averaged across the three raters, BrainVLM achieved an overall mean score of 4.78, substantially outperforming RadFM (3.92), Merlin (3.78), and GPT-4o (3.70). These results confirm that, beyond favorable performance on automated and LLM-as-a-judge metrics, the reports generated by BrainVLM are also judged by independent expert clinicians to be more consistent with ground-truth radiology reports, supporting the clinical fidelity of BrainVLM's report generation capability beyond what can be inferred from automated metrics alone.

## S3.6 Diagnosis performance in intraaxial and extraaxial tumor

To address potential overestimation of diagnostic accuracy arising from lesion location cues (e.g., specific regions like the pineal or sellar areas), we performed a stratified analysis by categorizing the dataset into intra-axial[85] and extra-axial[86] tumors based on their primary anatomical origin. According to established radiological definitions, intra-axial tumors (originating within the brain parenchyma) in this study included Gliomas, glioneuronal tumors, and neuronal tumors (GGN), Choroid plexus tumors (CPT), Embryonal tumors (EMB), Pineal region tumors (PIN), and Hematolymphoid tumors (HEM), while extra-axial tumors (originating outside the brain parenchyma) comprised Meningioma (MEN), Germ cell tumors (GCT), Tumors of the sellar region (TSR), Cranial and paraspinal nerve tumors (CPN), Mesenchymal non-meningothelial tumors (MNM), Melanocytic tumors (MEL), and Brain Metastases (MET). This classification ensured that the model's performance was evaluated independently across distinct anatomical compartments as defined in clinical practice.

The stratified performance metrics, detailed in Supplementary Figure 10b and d,

demonstrate that BrainVLM maintains robust diagnostic reliability across both anatomical compartments in the primary and external retrospective dataset. For intra-axial tumors, the macro-AUC values were 0.87 (Top-1) and 0.89 (Top-2) in the primary cohort, and 0.76 (Top-1) and 0.77 (Top-2) in the external validation cohort; for extra-axial tumors, the model achieved macro-AUCs of 0.87 (Top-1) and 0.90 (Top-2) in the primary test cohort, and 0.84 (Top-1) and 0.86 (Top-2) in the external test cohort. These results closely mirror the overall analysis, confirming that BrainVLM's performance is consistent across different anatomical compartments and is not solely reliant on location-based cues for its diagnostic predictions.

### S3.7 False negative finding investigation

We have extracted and analyzed the FNR for both BrainVLM and board-certified neuroradiologists across all 12 tumor subcategories. As demonstrated in Supplementary Figure 11, BrainVLM consistently exhibits FNRs that are largely comparable to, and in many instances, demonstrably lower than, those of board-certified neuroradiologists across both the primary and external retrospective datasets. A notable strength of BrainVLM is its ability to achieve substantially lower FNRs for several critical tumor types. This includes Embryonal tumors (EMB) (0.09 vs. 0.35 in the primary dataset; 0.06 vs. 0.44 in the external dataset), Pineal region tumors (PIN) (0.16 vs. 0.84 in the primary dataset; 0.67 vs. 0.83 in the external dataset), and Gliomas, glioneuronal tumors, and neuronal tumors (GGN) in the external dataset (0.16 vs. 0.36). These instances highlight areas where BrainVLM offers a clear and impactful diagnostic advantage.

Furthermore, for a substantial number of other tumor types, including Brain Metastases (MET), Germ cell tumors (GCT), Cranial and paraspinal nerve tumor (CPN), Hematolymphoid tumors (HEM), and Tumors of the sellar region (TSR), BrainVLM's FNRs were either closely similar to or slightly lower than those of neuroradiologists across both datasets, demonstrating strong performance parity. For example, BrainVLM achieved FNRs of 0.27 vs. 0.36 (primary) and 0.38 vs. 0.41 (external) for MET, 0.40 vs. 0.42 (primary) and 0.39 vs. 0.39 (external) for GCT, 0.20 vs. 0.22 (external) for CPN, and 0.08 vs. 0.08 (primary) and 0.18 vs. 0.19 (external) for TSR.

We also acknowledge instances where neuroradiologists showed lower FNRs, offering valuable targets for future model refinement. This includes mesenchymal, non-meningothelial tumors in the external dataset (MNM, 0.54 vs. 0.38) and meningiomas in the external dataset (MEN, 0.20 vs. 0.14). It is also important to note that for several inherently challenging and rare categories, such as CPT, HEM, and MEL in the external dataset, both BrainVLM and neuroradiologists exhibited relatively high FNRs (e.g., CPT: 0.83 for both; HEM: 0.70 for both; MEL: 0.80 for both). In conclusion, these comprehensive FNR comparisons, leveraging detailed data from both primary and external retrospective datasets, further substantiate BrainVLM's robust clinical utility.

## S3.8 Uncertainty quantification of BrainVLM

To rigorously evaluate BrainVLM's uncertainty awareness and clinical reliability, we have now rigorously assessed BrainVLM's calibration performance using standard metrics (Expected Calibration Error (ECE) and the Brier score) on the primary and external retrospective datasets. Here, ECE is a metric quantifying how closely uncertainty scores (predicted probabilities) match true outcomes, and Brier score measures the mean squared difference between the uncertainty score and the actual outcome. We achieved an ECE of 0.0360 and a Brier score of 0.1448. This remarkably low ECE indicates that the model's uncertainty scores are well aligned with true outcomes, maintaining an average calibration gap of 3.6%. The Brier score of 0.1448 further reflects the model's overall predictive accuracy, indicating a high degree of closeness between the uncertainty scores and the actual clinical outcomes. We further assessed its behavior through two complementary dimensions: 1) Stratified Confidence distribution analysis comparing high-performing versus low-performing diagnostic categories to validate calibration robustness, and 2) Clinical ambiguity verification on the bottom 10% low-confidence cases via a multi-reader consensus study to confirm that low confidence reflects genuine diagnostic difficulty.

### S3.8.1 Confidence distributions: top 3 vs. bottom 5 classes

To further validate the robustness of BrainVLM's calibration, we performed a comparative analysis of confidence distributions between the highest-performing (Top 3) and lowest-performing (Bottom 5) diagnostic categories. Specifically, we have now performed a more granular analysis of our Reliable BrainVLM's confidence distributions, focusing specifically on the Bottom 5 performing classes (based on F1-score) and the Top 3 performing classes for comparison (Supplementary Figure 12). Specifically, we generated detailed confidence distribution plots for the Bottom 5 classes (MEL [N = 25], PIN [N = 25], CPT [N = 47], MNM [N = 193], and HEM [N = 93]) as well as the Top 3 classes (TSR [N = 531], MEN [N = 1954], and GGN [N = 1584]) for comparison; see Supplementary Figure 12.

We observe that in the Top 3 classes, the vast majority of correct predictions (74% for TSR, 78% for MEN, and 70% for GGN) fall within the "Highest Confidence" bracket (0.9–1.0) range, while most incorrect predictions are associated with lower confidence scores. This is expected, as these classes are relatively easier for BrainVLM to classify. In contrast, for the Bottom 5 classes, the confidence scores for correct predictions are generally more dispersed across the confidence spectrum, reflecting the higher difficulty, pathological complexity, and rarity of these cases. Nevertheless, incorrect predictions in these challenging classes still predominantly correspond to low or medium confidence levels (75% for MEL, 80% for PIN, 71% for CPT, 81% for MNM, and 89% for HEM), further supporting the reliability of BrainVLM's confidence scores in indicating prediction uncertainty.

We note that for MEL (Melanocytic tumors), 100% of correct answers fall into the high-confidence bracket. This is reasonable given the very small sample size (N = 25); the few correctly classified cases happen to be those for which the model was most confident. This limited data warrant caution in interpreting the confidence distribution, and a larger sample would be needed for more robust conclusions. Additionally, for MNM, we observe that correct predictions show a higher proportion of medium or low confidence scores. Clinically, this is reasonable, as the MNM category encompasses a highly heterogeneous group under WHO CNS5, including lymphomas, sarcomas, and vascular tumors, with diverse radiological phenotypes. This heterogeneity presents a significant challenge for the model.

In summary, across both easy and challenging classes, our Reliable BrainVLM demonstrates the ability to concentrate correct predictions in high-confidence regions while suppressing confidence for errors. This is an important characteristic for safe clinical deployment.

**S3.8.2 Multi-readers assessment on bottom 10% confidence cases**

We performed an independent multi-reader study on the bottom 10% of cases (N = 385) where BrainVLM exhibited the lowest confidence. These cases were blindly reviewed by two junior radiologists (J1 and J2) and two expert neuroradiologists (E1 and E2). Supplementary table 12 illustrates the diagnostic landscape of these low-confidence cases, reflecting their inherent clinical complexity. We first compared the diagnostic accuracy of BrainVLM against human readers on this bottom 10% subset. BrainVLM achieved an accuracy of 65.0%, significantly outperforming the junior radiologists (0.51 and 0.44) and closely approaching the performance of experts (0.72 and 0.66) on the same subset. Notably, these expert scores represent a marked decline from the 0.80 average accuracy typically observed in our broader multi-reader study (N = 248). This universal drop in performance across both AI and experts confirms that BrainVLM's uncertainty scores successfully isolate cases that are inherently more difficult to diagnose.

To evaluate whether the subset identified by BrainVLM was objectively ambiguous, we conducted a multi-level consistency analysis involving two junior radiologists (J1, J2) and two senior experts (E1, E2). The results revealed a Cohen's Kappa of 0.33 between the two junior radiologists (J1 vs. J2), indicating only minimal agreement. While professional experience improved the consensus, the consistency between the two senior experts (E1 vs. E2) reached a Cohen's Kappa of 0.60. Furthermore, we compared the top-performing junior radiologist with the top-performing expert, resulting in a Kappa of 0.44, which highlights a significant diagnostic gap even between the most proficient readers across different seniority levels. Although this represents moderate-to-substantial agreement, it remains significantly below the high consensus threshold (Kappa > 0.80)[87] typically expected in routine neuro-oncological diagnosis. This diagnostic gap—where even seasoned experts exhibit a notable margin of disagreement—confirms that the bottom 10% subset identified by our model consists of inherently challenging and borderline cases.

## S3.9 Subgroup analysis across age, sex, MRI vendor and ethnicity

To rigorously evaluate the fairness and generalizability of BrainVLM, we conducted a comprehensive subgroup analysis across age, sex, and MRI vendor using both our primary and external retrospective cohorts (Supplementary Figure 13a). Regarding ethnicity, we acknowledge that our test dataset predominantly represents Asian populations due to our collaborating institutions being based in China. To directly address this limitation and assess generalizability, we augmented our evaluation with three public external datasets from European and North American healthcare systems. It is important to note that these newly acquired datasets were not part of the training process and served as truly external test sets. Detailed results for age, sex, MRI vendor, and ethnicity are provided below.

### S3.9.1 Analysis of demographic variables: age and sex

**Age-related Performance.** BrainVLM maintains a stable weighted F1-score (0.81) across most age groups. A relative performance dip was observed in the 0–20 years subgroup (F1 = 0.71, N = 479). Importantly, this trend aligns with human reader performance, as radiologists also exhibited their lowest F1 scores in this pediatric cohort (F1 = 0.67 compared with F1 = 0.79 in the overall retrospective test dataset). We observed that the performance decline in this specific subgroup is primarily driven by lower F1 scores in GGN, MEN and TSR. This variance is largely attributable to the intrinsic epidemiological rarity and atypical radiological presentations of these tumors in pediatric and adolescent patients. According to the CBTRUS Statistical Report,[88] the incidence of meningiomas and high-grade gliomas increases exponentially with age, making them exceptionally rare in the 0-20 population. Furthermore, tumors in this age group, such as craniopharyngiomas in the sellar region—often exhibit distinct morphological features and biological behaviors compared to their adult counterparts.[89] These results indicate that the performance variance is driven by inherent clinical complexity rather than algorithmic bias.

**Sex-related Performance.** We observed that the diagnostic F1-score for males (F1 = 0.80, N = 2,350) was slightly lower than for females (F1 = 0.84, N = 2,861). This finding is highly consistent with the diagnostic patterns of radiologists (male F1 = 0.75 vs. female F1 = 0.83). The consistency between BrainVLM and expert readers underscores the model's ability to mirror clinical reality across sex strata.

**S3.9.2 Analysis of MRI vendor**

We identified 1,881 patients (Primary: 1,451; External: 430) with complete metadata for the vendor-specific robustness analysis, as vendor identifiers were unavailable for all other cases (Supplementary Figure 13a). BrainVLM demonstrated robust performance across major MRI vendors, including GE (F1 = 0.81, N = 847), Siemens (F1 = 0.80, N = 774), and Toshiba (F1 = 0.87, N = 113). While smaller cohorts (e.g., UIH, N = 24, AIITECH, N = 44) showed more variance due to limited sample sizes, the overall performance remained consistently high (F1 = 0.75 and 0.76), proving that BrainVLM is generally resilient to variations in hardware signatures and acquisition protocols.

**S3.9.3 Analysis of ethnicity**

To assess performance across diverse ethnic populations, we evaluated BrainVLM on three public datasets from Europe and North America EGD,[90] Brain-Mets-Lung[91] and Vestibular-Schwannoma-MC2[92] as our external test datasets. The detailed results were presented in Supplementary Figure 14.

Erasmus Glioma Database (EGD)[90]: This dataset comprises 774 patients from a large-scale European glioma cohort, all with the full suite of MRI modalities (T1, T1c, T2, T2-Flair). Applying BrainVLM to these full MRI modalities yielded an F1-score of 0.91, precision 0.93, and sensitivity 0.91.

Brain-Mets-Lung (Yale University)[91]: This cohort consists of 100 patients with brain metastases curated from North American healthcare systems. Nevertheless, this cohort presents a significant challenge as only T1 and T1c sequences are available. Despite the restricted input, BrainVLM achieved an F1-score of 0.66.

Vestibular-Schwannoma-MC2 (London)[92]: This is a European multicenter dataset of 190 schwannoma patients. This database includes T1, T1c, and T2 sequences, but lacks T2-Flair datasets. On this cohort, BrainVLM achieved an F1-score of 0.83

Overall, BrainVLM demonstrated stable diagnostic performance across these diverse geographical and ethnic populations. While the F1-scores in the metastases and schwannoma cohorts were slightly lower compared to our primary test set (0.65 vs. 0.66 on metastases, and F1 0.83 vs. 0.88 on schwannoma, Supplementary Figure 13b), this marginal decline is largely attributable to the missing MRI modalities in these datasets. These results underscore BrainVLM's capacity to maintain diagnostic reliability even when transitioned to previously unseen healthcare environments and ethnic groups.

#### S3.9.4 Conclusion of subgroup analysis

Our comprehensive analysis demonstrates that BrainVLM maintains high diagnostic stability across diverse demographic and technical variables. The model's performance fluctuations across age and sex groups closely mirror those of experienced radiologists, suggesting that these variations stem from inherent clinical complexity rather than algorithmic bias. Furthermore, its consistent performance across different MRI vendors and its robust validation on western cohorts confirms the model's generalizability across diverse ethnicity populations. These results collectively validate BrainVLM as a fair and resilient tool for heterogeneous clinical environments.

### S3.10 Clinician-AI interaction in incorrect AI scenarios

In the multi-reader study, we specifically examined whether incorrect AI suggestions exert a negative impact on diagnostic outcomes. We assessed the direction of diagnostic changes within AI augmented workflows and evaluated whether erroneous AI outputs lead to automation bias or adversely affect clinician decision-making.

First, to examine the possibility of automation or anchoring bias in diagnostically challenging cases, we conducted a targeted diagnostic shift analysis focusing strictly on the subset of cases where clinicians reported their lowest 25% confidence scores (extracting based on the confidence score of the unaided cases). We then assessed the direction of diagnostic change when readers assessed cases with AI assistance. Specifically, clinician–AI interactions were categorized into six trajectories: (A) the clinician remained correct despite an incorrect AI suggestion; (B) the clinician successfully corrected an initially incorrect diagnosis; (C) the clinician changed from a correct diagnosis to an incorrect one after an incorrect AI suggestion; (D) both the initial clinician diagnosis and AI were incorrect, and the final diagnosis remained incorrect; (E) the clinician failed to adopt a correct AI suggestion, remaining incorrect; and (F) both the initial clinician diagnosis and AI suggestion were correct, with the final diagnosis remaining correct. Across all reader seniority groups, Trajectory F was the most common outcome overall. Notably, successful error correction (Trajectory B) was highly prominent and increased significantly with reader seniority, whereas cases of being erroneously misled by an incorrect AI (Trajectory C) were consistently the least common across all levels (Figure 4f). This demonstrates that rather than acting as a negative anchor, the AI actively and successfully helped neuroradiologists recover from their own diagnostic uncertainties.

Second, to explicitly address scenarios in which the AI was incorrect, we performed an additional analysis within the expanded 248-case multi-reader study, restricted to cases in which BrainVLM generated an incorrect diagnosis. In this subset (57 cases), we compared readers' unassisted baseline accuracy with their final accuracy after reviewing the incorrect AI output (Figure 4g). Overall accuracy remained stable (50.0% without AI vs 50.9% with AI). Junior neuroradiologists showed a modest decrease in accuracy (~6%), whereas senior and expert neuroradiologists maintained or slightly improved their performance on these cases (seniors: 49.0% to 52.9%; experts: 57.6% to 60.6%). A possible explanation is that many incorrect AI predictions were accompanied by low reliability scores, which may have prompted readers to apply greater caution when reviewing these outputs. These findings suggest that, within the setting of this reader study, exposure to incorrect AI outputs did not lead to systematic over-reliance overall.

## S3.11 Case studies

### S3.11.1 Case studies in high-confidence error prediction cases

While BrainVLM demonstrates high calibration (ECE = 0.0360), a small fraction of cases may yield high-confidence but incorrect predictions. We present two such examples (Supplementary Figure 15) involving Brain metastases tumor (MET) and Mesenchymal, non-meningothelial tumor (MNM), to provide insights into how such challenging cases should be addressed and managed within a clinical workflow.

Case 1: An MET was misclassified as a Glioma (GGN) with 80% confidence. The high confidence likely resulted from the larger tumor volume, heterogeneous enhancement and prominent peritumoral edema, a feature more commonly associated with high-grade gliomas, which confused the model.

Case 2: An MNM (pathological confirmed with Solitary fibrous tumor) was misclassified as a cranial and paraspinal nerve tumor (CPN) with 90% confidence. The high confidence likely resulted from the tumor location usually occurring CPN. Cases misclassified by the model with high confidence also present substantial diagnostic challenges to radiologists.

Nevertheless, these mistaken classifications did not alter the subsequent clinical management, given that all cases were eligible for surgical resection. To mitigate the risk of over-reliance on erroneous high-confidence predictions, we envision a deployment strategy where confidence scores are used to stratify risk: low-confidence cases trigger human experts for secondary review, while high-confidence predictions are accompanied by interpretability images to facilitate clinician verification rather than blind acceptance. BrainVLM is an assistive diagnostic tool and the clinician retains final responsibility, especially when AI confidence conflicts with clinical intuition.

### S3.11.2 Case studies in multi-readers' time reduction

We have included 2 representative cases from our multi-reader study that demonstrate a substantial reduction in diagnostic time with BrainVLM integration (Supplementary Figure 16).

Case 1: A 62-year-old female patient with histopathologically confirmed brain metastasis. Six MRI sequences were available (axial T1, T1-contrast, T2, T2-Flair; coronal and sagittal T1-contrast) for this patient. In our study, this case was first interpreted in the expert-only setting and then, after a one-month washout period, re-interpreted by the same expert with AI support. In the expert-only setting, the expert neuroradiologist misidentified the lesion as a glioma with confidence 60%. The diagnostic process took 132 seconds. Our BrainVLM outputs "Brain metastase tumor" (MET) as the most probable category with a confidence score of 75%. Guided by this AI-generated suggestion, the expert outputs the correct diagnosis in only 40 seconds, demonstrating a 70% reduction in diagnostic time (132s to 40s).

Case 2: An 8-year-old female patient with histopathologically confirmed Glioma. Six MRI sequences were available (axial T1, T1-contrast, T2, T2-Flair; coronal and sagittal T1-contrast) for this patient. In our study, this case was first interpreted in the AI-augmented setting and then, after a one-month washout period, re-interpreted by the same expert without AI support. In the expert-only setting, the neuroradiologist correctly identified glioma but with relatively low confidence (60%), requiring 80 seconds to analyze this case. In the AI-augmented setting, BrainVLM outputs "glioma" as the most probable category with a high confidence score of 90%. Using this AI output as decision support, the expert finalized the same diagnosis in only 23 seconds. This case illustrates a 71% reduction in diagnostic time (80s to 23s). This synergistic effect significantly accelerated the decision-making process, demonstrating BrainVLM's role in reinforcing clinical certainty for complex cases.

## S4 Regulatory considerations, failure modes, and medico-legal implications

**Regulatory considerations.** This study adhered to strict data governance protocols to protect the privacy of patients from the real world. All clinical and imaging data were fully anonymized and encrypted during storage and transmission. Access controls were implemented to ensure that only approved study personnel could interact with the dataset, and data usage was contractually and ethically bound to the specific aims of this research. These measures align with the principles of patient data protection required by regulatory frameworks such as Health Insurance Portability and Accountability Act (HIPAA) or General Data Protection Regulation (GDPR), ensuring the integrity and confidentiality of the data used for model development and validation.

**Failure modes.** Several scenarios where model performance degrades or fails in a clinically significant manner and solutions. (a) Low-confidence predictions: Cases where BrainVLM's prediction confidence falls below a predefined threshold (e.g., < 85%) are flagged for secondary review by human experts. (b) High-confidence discrepancies: BrainVLM's high-confidence prediction conflicts with the clinician's observation or the patient's clinical presentation, requiring further verification. (c) Scope limitation (non-tumoral cases): BrainVLM is specifically designed for tumor type classification tasks (i.e., differentiation among known tumor types), rather than initial screening to differentiate between tumoral and non-tumoral cases. Therefore, its application requires prior identification of brain tumor cases, ideally through a dedicated screening model.

**Medico-legal implications.** BrainVLM may serve as an assistive diagnostic tool, not a substitute for the neuroradiologist or clinician. The final diagnostic responsibility must always remain with the clinician.

## S5 Data and code availability

**Data availability.** The online training data are available on their corresponding websites. We will publish the scripts for pre-processing the online dataset upon publication of this manuscript. The institutional training data that support the findings of this study are available from the corresponding author upon request. The detailed publicly available resources can be accessed at: PubMed Central (https://pmc.ncbi.nlm.nih.gov/), Ctisus (https://www.ctisus.com/), ImageCLEFmedical23 (https://www.imageclef.org/2023/medical/caption), Br35H (https://www.kaggle.com/datasets/ahmedhamada0/brain-tumor-detection), Figshare Brain Tumor Dataset (https://www.kaggle.com/datasets/ashkhagan/figshare-brain-tumor-dataset), Brain Tumor Classification (https://www.kaggle.com/datasets/sartajbhuvaji/brain-tumor-classification-mri), Radiopaedia (https://radiopaedia.org/home), Brain Tumor MRI8 images 44 classes (https://www.kaggle.com/datasets/fernando2rad/brain-tumor-mri-images-44c), LGG-1p19qDeletion (https://www.cancerimagingarchive.net/collection/lgg-1p19qdeletion/), BraTS23 (https://synapse.org/brats), ReMIND (https://www.cancerimagingarchive.net/collection/remind/), RHUH-GBM

(https://www.cancerimagingarchive.net/collection/rhuh-gbm/), UPENN-GBM (https://www.cancerimagingarchive.net/collection/upenn-gbm/), GLIS-RT (https://www.cancerimagingarchive.net/collection/glis-rt/), QIN GBM Treatment Response (https://www.cancerimagingarchive.net/collection/qin-gbm-treatment-response/), LUMIERE (https://github.com/ysuter/gbm-data-longitudinal), OpenNeuro Diffuse Gliomas (https://openneuro.org/datasets/ds004717/versions/1.0.0), Brain-Tumor- Progression (https://www.cancerimagingarchive.net/collection/brain-tumor-progression/), RIDER (https://www.cancerimagingarchive.net/collection/rider-lung-ct/), ACRIN-DSC- MR-Brain (https://www.cancerimagingarchive.net/collection/acrin-dsc-mr-brain/), ACRINFMISO-Brain (https://www.cancerimagingarchive.net/collection/acrin-fmiso-brain/), Meningioma-SEG-CLASS (https://www.cancerimagingarchive.net/collection/meningioma-seg-class/), Brain metastase MRI dataset (https://www.nature.com/articles/s41597-023-02123-0), Brain-TR- GammaKnife, AOMIC- ID1000 (https://openneuro.org/datasets/ds003097), AOMIC- PIOP1 (https://openneuro.org/datasets/ds002785), IXI (https://brain-development.org/ixi-dataset/), OpenNeuro Listening Task (https://openneuro.org/datasets/ds004285/versions/1.0.0), QTAB dataset (https://openneuro.org/datasets/ds004146/versions/1.0.4), OpenNeuro Visual Audiovisual Speech (https://openneuro.org/datasets/ds003717/versions/1.0.1), OpenNeuro Dynamic Passive Threat (https://openneuro.org/datasets/), OpenNeuro UCLA Consortium (https://openneuro.org/datasets/ds000030/versions/00016), OpenNeuro NIMH Healthy (https://openneuro.org/datasets/ds004215), DLBS dataset (https://openneuro.org/datasets/ds004856/versions/1.0.0), OpenNeuro Healthy Adults (https://openneuro.org/datasets/ds002330/versions/1.1.0/1.1.0), EGD Dataset (https://www.healthinformationportal.eu/health-information-sources/erasmus-glioma-database), Brain-Mets-Lung Dataset (https://www.cancerimagingarchive.net/collection/brain-mets-lung-mri-path-segs/) and Vestibular-schwannoma-MC2 (https://www.cancerimagingarchive.net/collection/vestibular-schwannoma-mc-rc2/). Note that some online resources (e.g., Radiopaedia) cannot be directly accessed for AI research purposes. We have obtained official authorization for non-commercial use in this study.

**Code availability.** BrainVLM will be fully available at https://github.com/HKU-HealthAI/BrainVLM upon acceptance of the paper. We will release all model weights and relevant source code for pretraining, fine-tuning, and inference to facilitate research transparency and community collaboration.

## Supplementary Tables

**Supplementary Table 1 |** Characteristics of the retrospective primary dataset (Xiangya Hospital) and external test datasets from 11 independent medical centers.

| | No. of subjects | Sex | | Age in years (range) |
|---|---|---|---|---|
| | | Male | Female | |
| Total | 10,147 | 4,921 (48%) | 5,226 (52%) | 53±16 (1-86) |
| Train dataset | 4,936 | 2,571 (52%) | 2,365 (48%) | 43± 19 (1-86) |
| Test dataset | 5,211 | 2,350 (45%) | 2,861 (55%) | 48±17 (1-86) |
| Xiangya Hospital | 8,813 | 4,256 (48%) | 4,557 (52%) | 45±18 (1-86) |
| Changde First People's Hospital | 361 | 193 (53%) | 168 (47%) | 55±13 (6-86) |
| Shanghai Tongji Hospital | 104 | 54 (52%) | 50 (48%) | 54±16 (5-79) |
| University of South China Second Hospital | 110 | 57 (52%) | 53 (48%) | 52±15 (6-80) |
| Shenzhen Second People's Hospital | 52 | 30 (58%) | 22 (42%) | 49±15 (8-75) |
| The Third Xiangya Hospital | 103 | 49 (48%) | 54 (52%) | 52±17 (7-76) |
| Jiangxi Provincial People's Hospital | 232 | 107 (46%) | 125 (54%) | 53±15 (13-82) |
| Chongqing Traditional Chinese medicine Hospital | 42 | 17 (40%) | 25 (60%) | 58±12 (26-78) |
| The First Affiliated Hospital of Nanchang University | 97 | 40 (41%) | 57 (59%) | 46±17 (20-67) |
| Hunan Provincial Children's Hospital | 103 | 56 (54%) | 47 (46%) | 28±17 (1-60) |
| The First Affiliated Hospital of Lanzhou University | 81 | 31 (38%) | 50 (62%) | 52±13 (8-76) |
| The Second Affiliated Hospital of Nanchang University | 49 | 31 (63%) | 18(37%) | 54±12 (13-74) |

**Supplementary Table 2 |** Summary of publicly available brain MRI datasets. The table summarizes a range of datasets, detailing the number of subjects, 2D MRI slices, data modalities (2D/3D), and pathological labels (tumor or healthy). For each dataset, original source information and the specific tasks organized in this study are also provided. Here, "3D Healthy" refers to datasets containing healthy individuals with 3D MRI volumes, while "3D Tumor" refers to datasets containing brain tumor patients with 3D MRI volumes. The PubMed Central dataset includes both normal and brain tumor subjects. For ReMIND dataset, we utilized 108 brain tumor cases, excluding 6 brain disease cases. Please note that, due to policy restrictions (e.g., Radiopaedia), certain portions of the training dataset cannot be shared.

| Data Type | Data Resource | No. of subjects | No. of images | Original Information | Organized Tasks |
|---|---|---|---|---|---|
| 2D web crawling | PubMed Central | 23,849 | 116,296 | Description | 2D MRI slice-text description<br>Tumor classification |
| 2D web crawling | Ctisus | 150 | 1,124 | Description,<br>Diagnosis | 2D MRI slice-text description<br>Tumor classification |
| 2D web crawling | ImageCLEFmedical23 | 484 | 4,631 | Description | 2D MRI slice-text description<br>Tumor classification |
| 2D Kaggle | Br35H | - | 2,137 | Diagnosis<br>Segment | 2D MRI slice-text description<br>Tumor classification |
| 2D Kaggle | Figshare Brain Tumor Dataset | 233 | 3,064 | Diagnosis | Tumor classification |
| 2D Kaggle | Brain Tumor Classification | - | 3,264 | Diagnosis | Tumor classification |
| 2D Kaggle | Brain Tumor MRI images 44 classes | - | 4,479 | Diagnosis | Tumor classification |
| 3D Tumor | LGG-1p19qDeletion | 159 | 32,829 | Diagnosis<br>Segment | 3D MRI-report<br>Tumor classification |
| 3D Tumor | BraTS23 | 3,263 | 2,156,589 | Diagnosis<br>Segment | 3D MRI-report<br>Tumor classification |
| 3D Tumor | ReMIND* | 108 | 60,963 | Diagnosis<br>metadata | 3D MRI-report<br>Tumor classification |
| 3D Tumor | RHUH-GBM | 40 | 122,835 | Diagnosis | Tumor classification |
| 3D Tumor | UPENN-GBM | 630 | 622,112 | Diagnosis<br>Segment | 3D MRI-report<br>Tumor classification |
| 3D Tumor | GLIS-RT | 230 | 83,196 | Diagnosis | Tumor classification |
| 3D Tumor | QIN GBM Treatment Response | 54 | 66,988 | Diagnosis | Tumor classification |
| 3D Tumor | LUMIERE | 91 | 460,155 | Diagnosis | Tumor classification |
| 3D Tumor | OpenNeuro (Diffuse Gliomas) | 42 | 22,412 | Diagnosis | Tumor classification |
| 3D Tumor | Brain-Tumor-Progression | 20 | 46,573 | Diagnosis<br>Segment | 3D MRI-report<br>Tumor classification |
| 3D Tumor | Radiopaedia | 2606 | 376, 374 | Diagnosis<br>Report | 3D MRI-report<br>Tumor classification |
| 3D Tumor | Burdenko | 180 | 460,524 | Diagnosis | Tumor classification |
| 3D Tumor | RIDER | 19 | 115,100 | Diagnosis | Tumor classification |
| 3D Tumor | ACRIN-DSC-MR-Brain | 123 | 1,385,512 | Diagnosis | Tumor classification |
| 3D Tumor | ACRIN-FMISO-Brain | 50 | 508,500 | Diagnosis | Tumor classification |
| 3D Tumor | Meningioma-SEG-CLASS | 96 | 83,556 | Diagnosis<br>metadata | 3D MRI-report<br>Tumor classification |

**Supplementary Table 2 |** Continuation of Table 2. Summary of publicly available brain MRI datasets. The table summarizes a range of datasets, detailing the number of subjects, 2D MRI slices, data modalities (2D/3D), and pathological labels (tumor or healthy). For each dataset, original source information and the specific tasks organized in this study are also provided. Here, "3D Healthy" refers to datasets containing healthy individuals with 3D MRI volumes, while "3D Tumor" refers to datasets containing brain tumor patients with 3D MRI volumes. The PubMed Central dataset includes both normal and brain tumor subjects. For ReMIND dataset, we utilized 108 brain tumor cases, excluding 6 brain disease cases.

| Data Type | Data Resource | No. of subjects | No. of images | Original Information | Organized Tasks |
|---|---|---|---|---|---|
| 3D Tumor | Brain metastase MRI dataset | 75 | 85,400 | Diagnosis Segment | 3D MRI-report Tumor classification |
| 3D Tumor | Brain-TR-GammaKnife | 47 | 8,300 | Diagnosis | Tumor classification |
| 3D Healthy | AOMIC-ID1000 | 928 | 1, 101, 652 | Diagnosis | Tumor or healthy classification |
| 3D Healthy | AOMIC-PIOP1 | 216 | 251, 380 | Diagnosis | Tumor or healthy classification |
| 3D Healthy | IXI | 582 | 398, 774 | Diagnosis | Tumor or healthy classification |
| 3D Healthy | OpenNeuro (Listening Task) | 78 | 136, 694 | Diagnosis | Tumor or healthy classification |
| 3D Healthy | OpenNeuro (Healthy Adults) | 66 | 122, 175 | Diagnosis | Tumor or healthy classification |
| 3D Healthy | OpenNeuro (Visual Audiovisual Speech) | 60 | 105, 160 | Diagnosis | Tumor or healthy classification |
| 3D Healthy | OpenNeuro (Dynamic Passive Threat) | 82 | 126, 299 | Diagnosis | Tumor or healthy classification |
| 3D Healthy | OpenNeuro (UCLA Consortium) | 272 | 285, 910 | Diagnosis | Tumor or healthy classification |
| 3D Healthy | OpenNeuro (NIMH Healthy) | 157 | 461, 281 | Diagnosis | Tumor or healthy classification |
| 3D Healthy | DLBS dataset | 315 | 99, 344 | Diagnosis | Tumor or healthy classification |
| 3D Healthy | QTAB dataset | 422 | 1, 251, 467 | Diagnosis | Tumor or healthy classification |
| Public test | EGD dataset | 774 | - | Diagnosis | Tumor classification |
| Public test | Brain-Mets-Lung | 100 | - | Diagnosis | Tumor classification |
| Public test | Vestibular-Schwannoma-MC2 | 190 | - | Diagnosis | Tumor classification |

**Supplementary Table 3 |** Characteristics of primary prospective dataset, including surgical group and non-operative group for each brain tumor type.

| | Surgical Group | | | | Non-operative group | | | | Entire dataset |
|---|---|---|---|---|---|---|---|---|---|
| | No. of subjects | Sex | | Age in years Mean ± SD (range) | No. of subjects | Sex | | Age in years Mean ± SD (range) | |
| | | Male | Female | | | Male | Female | | |
| Total | 639 | 290 | 359 | 48±17 (2-83) | 74 | 33 | 41 | 41±21 (2-66) | 713 |
| MET | 25 | 16 | 10 | 60±8 (38-72) | 7 | 4 | 3 | 45±11 (34-55) | 32 |
| GCT | 15 | 11 | 6 | 16±9 (7-39) | 3 | 2 | 1 | 8±1 (7-9) | 18 |
| GGN | 119 | 69 | 50 | 44±19 (5-83) | 18 | 8 | 10 | 30±24 (2-60) | 137 |
| MEN | 185 | 44 | 141 | 53±11 (14-78) | 22 | 6 | 16 | 50±11 (29-66) | 207 |
| TSR | 146 | 81 | 68 | 49±16 (4-77) | 17 | 9 | 8 | 54±8 (37-63) | 163 |
| MNM | 40 | 22 | 21 | 38±21 (2-72) | 2 | 1 | 1 | 15±1 (14-16) | 42 |
| CPN | 87 | 36 | 52 | 52±13 (22-76) | 5 | 3 | 2 | 54±15 (42-62) | 92 |
| CPT | 4 | 1 | 3 | 43±15 (19-60) | - | - | - | - | 4 |
| HEM | 10 | 5 | 5 | 64±7 (54-74) | - | - | - | - | 10 |
| EMB | 7 | 5 | 2 | 17±12 (5-35) | - | - | - | - | 7 |
| PIN | 1 | 0 | 1 | 43 | - | - | - | - | 1 |

In this table, we use abbreviations for tumor categories: MET (brain metastases), GCT (germ cell tumors), GGN (gliomas, glioneuronal tumors, and neuronal tumors), MEN (meningioma), TSR (tumors of the sellar region), MNM (mesenchymal, non-meningothelial tumors), CPN (cranial and paraspinal nerve tumors), CPT (choroid plexus tumors), HEM (hematolymphoid tumors), EMB (embryonal tumors), PIN (pineal region tumors), and MEL (melanocytic tumors).

**Supplementary Table 4 |** Characteristics of external prospective dataset, including surgical group and non-operative group for each brain tumor type.

| | Surgical Group | | | | Non-operative group | | | | Entire dataset |
|---|---|---|---|---|---|---|---|---|---|
| | No. of subjects | Sex | | Age in years Mean ± SD (range) | No. of subjects | Sex | | Age in years Mean ± SD (range) | |
| | | Male | Female | | | Male | Female | | |
| Total | 247 | 116 | 131 | 54±15 (5-79) | 49 | 22 | 27 | 55±18 (9-84) | 296 |
| MET | 12 | 10 | 2 | 58±15 (20-72) | 5 | 0 | 5 | 67±12 (56-84) | 17 |
| GCT | 2 | 1 | 1 | 18±1 (17-18) | 2 | 1 | 1 | 13±6 (9-17) | 4 |
| GGN | 54 | 29 | 25 | 51±16 (15-79) | 10 | 6 | 4 | 64±8 (51-75) | 64 |
| MEN | 81 | 31 | 50 | 57±10 (34-77) | 14 | 3 | 11 | 63±8 (50-75) | 95 |
| TSR | 66 | 31 | 35 | 52±16 (11-74) | 12 | 8 | 4 | 49±20 (15-72) | 78 |
| MNM | 6 | 3 | 3 | 34±20 (5-59) | 1 | 1 | 0 | 62±0 (62-62) | 7 |
| CPN | 22 | 8 | 14 | 55±14 (24-78) | 2 | 1 | 1 | 54±10 (47-61) | 24 |
| HEM | 4 | 3 | 1 | 68±6 (63-77) | 2 | 2 | 0 | 35±13 (25-44) | 6 |
| EMB | - | - | - | - | 1 | 0 | 1 | 14±0 (14-14) | 1 |

In this table, we use abbreviations for tumor categories: MET (brain metastases), GCT (germ cell tumors), GGN (gliomas, glioneuronal tumors, and neuronal tumors), MEN (meningioma), TSR (tumors of the sellar region), MNM (mesenchymal, non-meningothelial tumors), CPN (cranial and paraspinal nerve tumors), CPT (choroid plexus tumors), HEM (hematolymphoid tumors), EMB (embryonal tumors), PIN (pineal region tumors), and MEL (melanocytic tumors).

**Supplementary Table 5.** Comprehensive Statistical Comparison using DeLong Test for macro AUCs. **I, II** are the Macro AUC comparison between BrainVLM and other baseline models in the retrospective primary and external dataset. **III** illustrates the comparison between BrainVLM and neuroradiologists of varying expertise levels (Junior, Senior, and Expert) within the 248-case multi-reader study.

| Category | Comparison Group | Baseline AUC | BrainVLM AUC | Δ AUC | P-value |
|---|---|---|---|---|---|
| **I. Model vs. SOTA Baselines (Primary Test Set)** | | | | | |
| | vs. Merlin | 0.62 | 0.85 | 0.23 | **< 0.001** |
| | vs. RadFM | 0.66 | 0.85 | 0.19 | **< 0.001** |
| | vs. Video Swin Transformer | 0.73 | 0.85 | 0.12 | **< 0.001** |
| **II. Model vs. SOTA Baselines (External Test Set)** | | | | | |
| | vs. Merlin | 0.64 | 0.80 | 0.16 | **< 0.001** |
| | vs. RadFM | 0.60 | 0.80 | 0.20 | **< 0.001** |
| | vs. Video Swin Transformer | 0.62 | 0.80 | 0.18 | **< 0.001** |
| **III. Model vs. Human Readers (Primary Test Set)** | | | | | |
| | vs. Junior Radiologist | 0.76 | 0.88 | 0.12 | **< 0.001** |
| | vs. Senior Radiologist | 0.80 | 0.88 | 0.08 | **< 0.001** |
| | vs. Expert Radiologist | 0.89 | 0.88 | -0.01 | 0.28 (NS) |
| | vs. All Readers (Average) | 0.87 | 0.88 | 0.01 | **0.004** |

**Supplementary Table 6 |** Abbreviation of 12 major brain tumors in WHO CNS5.

| Acronym | Original Tumor Type Name |
|---|---|
| GGN | Gliomas, glioneuronal tumors, and neuronal tumors |
| MET | Brain metastases |
| GCT | Germ cell tumors |
| MEN | Meningioma |
| TSR | Tumor of the sellar region |
| MNM | Mesenchymal, non-meningothelial tumors |
| CPN | Cranial and paraspinal nerve tumors |
| CPT | Choroid plexus tumors |
| HEM | Hematolymphoid tumors |
| EMB | Embryonal tumors |
| PIN | Pineal tumors |
| MEL | Melanocytic tumors |

**Supplementary Table 7 |** Overview of additional MRI sequences and patient information utilized in radiological diagnosis in the primary test dataset and external test dataset. Extra MRI sequences refer to imaging techniques beyond standard T1, T1c, T2, and FLAIR. These extra sequences include Diffusion-Weighted Imaging (DWI), Diffusion Tensor Imaging (DTI), Perfusion-Weighted Imaging (PWI), Magnetic Resonance Angiography (MRA), Magnetic Resonance Spectroscopy (MRS), Susceptibility-Weighted Imaging (SWI), Magnetic Resonance Venography (MRV), Blood Oxygen Level Dependent Imaging (BOLD), Computed Tomography (CT), and Apparent Diffusion Coefficient (ADC). Additional patient information includes access to medical history and prior radiology imaging and associated findings, offering further clinical context for diagnostic decision-making.

| Dataset | Total Number | Extra Modality Percentage | Extra Prior Medical History of Pathological diagnosis | Overall Extra Information Percentage |
|---|---|---|---|---|
| Primary test dataset | 3,797 | 39.5% | 19.3% | 58.8% |
| External dataset | 1,309 | 56.1% | 12.7% | 68.8% |

**Supplementary Table 8 |** Summary of MRI scanner manufacturers, models, and scanning parameters used across all medical centers.

| Scanning Protocol | MRI Scanner Manufacturer | | | |
|---|---|---|---|---|
| | GE | Siemens | Philips | UIH |
| Model | Discovery MR750w, Signa HDxt, Signa Excite | Skyra, Prisma, Avanto, Syngo | Ingenia, Ingenuit | uMR 660 |
| Field strength | 1.5 T/ 3.0 T | 1.5 T/ 3.0 T | 3.0 T | 1.5 T |
| Flip angle | 1°, 8°, 12°, 15°, 18°, 19°, 20°, 25°, 30°, 40°, 60°, 70°, 90°, 111°, 112°, 125°, 142°, 155°, 160° | 7°, 8°, 9°, 10°, 15°, 20°, 24°, 25°, 27°, 30°, 40°, 50°, 60°, 70°, 80°, 90°, 120°, 122°, 140°, 145°, 150°, 160°, 180° | 8°, 10°, 15°, 17°, 18°, 20°, 30°, 40°, 45°, 70°, 75°, 80°, 90°, 100° | 10°, 12°, 15°, 20°, 25°, 54°, 60°, 70°, 72°, 90°, 120°, 130°, 140°, 150° |
| Pixel Spacing Range ($mm^2$) | 0.17 × 0.17 - 3.5 × 3.5 | 0.23 × 0.23 - 4.7 × 4.7 | 0.21 × 0.21 - 1.8 × 1.8 | 0.26 × 0.26 - 1.4 × 1.4 |
| Slice Thickness Range (mm) | 0.6 - 278 | 0.47 - 60 | 0.75 - 398 | 0.5-10 |

**Supplementary Table 9 |** The designed special tokens for data differentiation and task-driven prompt design to distinguish between different tasks. A 2D MRI slice is enclosed with ‘<image>’ and ‘</image>’ tags, while 3D MRI volumes use ‘<vid>’ and ‘</vid>’ tags. The <img> and <vi> special tokens represent the 2D image features and 3D volume features, respectively. Additionally, a special token ‘<s>’ is inserted between frames to aid in learning spatial dependencies within the 3D volume.

| Tasks | Visual Tokens | Task-specific Instructions Example | Sample Output |
|---|---|---|---|
| 2D MRI slice level tumor classification | <image><img></image> | Is a tumor present in this MRI slice? If so, what type? | <tumor_i> |
| 2D MRI slice-text description | <image><img></image> | Please make a description for this image. | A lession in left lobe, showing T1 hyperintense |
| 3D MRI tumor classification | <vid1><v1><v1>...</vid1> <vid2><v2><v2>...</vid2> ... | For a 6-year-old male patient, is a tumor present in these MRIs? If so, what type? | <tumor_i> |
| 3D MRI-report generation | <vid1><v1><v1>...</vid1><s> <vid2><v2><v2>...</vid2><s> ... | For a 6-year-old male patient, please generate a radiology report for him. | In the ..., hyperintense... midline shift to ... |
| 3D MRI report generation tumor classification | <vid1><v1><v1>...</vid1><s> <vid2><v2><v2>...</vid2><s> ... | For a 6-year-old male patient, please generate a radiology report and make a diagnosis. | In the ..., hyperintense...; <tumor_i> |
| Uncertainty quantification | <vid1><v1><v1>...</vid1><s> <vid2><v2><v2>...</vid2><s> ... | For a 6-year-old male patient, please generate a radiology report, make a diagnosis, and generate reliable score for your diagnosis. | In the ..., hyperintense... midline shift to ... <tumor_i>, confidence: 80% |

**Supplementary Table 10** |Diagnostic performance comparison between full MRI sequences (T1, T1c, T2, FLAIR) input and excluding one of the four standard sequences, evaluated on 3,571 patients from the primary and external test datasets. Only patients with complete MRI data across all four sequences were included to eliminate potential confounding due to missing sequences. The remaining 1,555 patients in the original primary and external test datasets with missing sequences were excluded from the analysis.

| | $F_1$ score | | | | |
|---|---|---|---|---|---|
| | **Full Sequence** | **Excluded T1** | **Excluded T1c** | **Excluded T2** | **Excluded FLAIR** |
| MET | **0.58** | 0.52 | 0.45 | 0.54 | 0.53 |
| GCT | **0.64** | 0.48 | 0.47 | 0.58 | 0.53 |
| GGN | **0.83** | 0.80 | 0.72 | 0.82 | 0.82 |
| MEN | **0.92** | 0.89 | 0.82 | 0.89 | 0.89 |
| TSR | **0.75** | 0.65 | 0.52 | 0.63 | 0.63 |
| MNM | **0.56** | 0.43 | 0.44 | 0.45 | 0.48 |
| CPN | **0.81** | 0.76 | 0.64 | 0.79 | 0.78 |
| CPT | **0.44** | 0.27 | 0.27 | 0.43 | 0.27 |
| HEM | **0.58** | 0.43 | 0.42 | 0.42 | 0.45 |
| EMB | **0.67** | 0.44 | 0.4 | 0.42 | 0.44 |
| PIN | **0.43** | 0.36 | 0.22 | 0.43 | 0.29 |
| MEL | **0.17** | 0.17 | 0.17 | 0.17 | 0.17 |
| Frequency-weighted F1 | **0.82** | 0.78 | 0.71 | 0.79 | 0.79 |

In this table, we use a series of abbreviations to replace orginal categories: MET for brain metastases, GCT for germ cell tumors, GGN for gliomas, glioneuronal tumors, and neuronal tumors, MEN for meningioma, TSR for tumors of the sellar region, MNM for mesenchymal, non-meningothelial tumors, CPN for cranial and paraspinal nerve tumors, CPT for choroid plexus tumors, HEM for hematolymphoid tumors, EMB for embryonal tumors, PIN for pineal region tumors and MEL for melanocytic tumors.

**Supplementary Table 11 |** Comparison of F1 scores between our BrainVLM model, neuroradiologists, and three state-of-the-art AI models (RadFM, Merlin, and ChatGPT-4o) and traditional deeplearning model(video-swin-transformer (VST)). We evaluated 3,877 patients from the primary test dataset and 1,334 patients from the external test dataset, respectively.

| | $F_1$ score (Primary Testing) | | | | | | $F_1$ score (External Testing) | | | | | |
|---|---|---|---|---|---|---|---|---|---|---|---|---|
| | **Ours** | **Doctor** | **RadFM** | **Merlin** | **GPT-4o** | **VST** | **Ours** | **Doctor** | **RadFM** | **Merlin** | **GPT-4o** | **VST** |
| MET | 0.53 | **0.56** | 0.45 | 0.25 | 0.04 | 0.50 | **0.61** | 0.52 | 0.44 | 0.40 | 0.10 | 0.22 |
| GCT | **0.62** | 0.56 | 0.33 | 0.15 | 0.08 | 0.48 | **0.73** | 0.52 | 0.30 | 0.10 | 0.11 | 0.30 |
| GGN | **0.82** | 0.81 | 0.68 | 0.59 | 0.47 | 0.65 | **0.79** | 0.69 | 0.45 | 0.56 | 0.38 | 0.49 |
| MEN | **0.91** | 0.91 | 0.64 | 0.73 | 0.61 | 0.59 | 0.82 | **0.83** | 0.53 | 0.63 | 0.41 | 0.4 |
| TSR | **0.88** | **0.88** | 0.75 | 0.49 | 0.56 | 0.80 | **0.87** | 0.84 | 0.39 | 0.55 | 0.48 | 0.78 |
| MNM | **0.56** | 0.43 | 0.30 | 0.12 | 0.03 | 0.23 | 0.46 | **0.55** | 0.27 | 0.0 | 0.0 | 0.36 |
| CPN | **0.78** | 0.74 | 0.40 | 0.27 | 0.17 | 0.45 | **0.8** | 0.74 | 0.28 | 0.30 | 0.11 | 0.36 |
| CPT | **0.55** | 0.47 | 0.19 | 0.36 | 0.08 | 0.3 | **0.29** | 0.18 | 0.0 | 0.0 | 0.0 | 0.20 |
| HEM | **0.64** | 0.45 | 0.31 | 0.15 | 0.0 | 0.12 | **0.41** | 0.39 | 0.33 | 0.28 | 0.0 | 0.1 |
| EMB | **0.68** | 0.58 | 0.47 | 0.29 | 0.13 | 0.43 | **0.48** | 0.42 | 0.26 | 0.25 | 0.0 | 0.14 |
| PIN | **0.42** | 0.13 | 0.27 | 0.17 | 0.0 | 0.25 | **0.36** | 0.17 | 0.0 | 0.0 | 0.0 | 0.0 |
| MEL | **0.29** | 0.25 | 0.0 | 0.0 | 0.0 | 0.0 | - | - | 0.0 | 0.0 | 0.0 | 0.0 |
| Weighted $F_1$ | **0.82** | 0.81 | 0.62 | 0.54 | 0.37 | 0.59 | **0.76** | 0.71 | 0.43 | 0.52 | 0.30 | 0.42 |

In this table, we use a series of abbreviations to replace original categories: MET for brain metastases, GCT for germ cell tumors, GGN for gliomas, glioneuronal tumors, and neuronal tumors, MEN for meningioma, TSR for tumors of the sellar region, MNM for mesenchymal, non-meningothelial tumors, CPN for cranial and paraspinal nerve tumors, CPT for choroid plexus tumors, HEM for hematolymphoid tumors, EMB for embryonal tumors, PIN for pineal region tumors and MEL for melanocytic tumors.

**Supplementary Table 12 |** Multi-reader consensus performance on the bottom 10% lowest confidence cases identified by BrainVLM (N=385). This table compares the diagnostic accuracy of BrainVLM against two junior (J1, J2) and two expert (E1, E2) neuroradiologists on this challenging subset.

| Tumor Type | Patient Number | Model Accuracy | Junior Radiologists | | Expert Radiologists | |
|---|---|---|---|---|---|---|
| | | | J1 Acc. | J2 Acc. | E1 Acc. | E2 Acc. |
| MET | 44 | **0.55** | 0.27 | 0.18 | 0.36 | **0.55** |
| GCT | 20 | 0.60 | 0.40 | 0.40 | 0.60 | **0.70** |
| GGN | 72 | 0.70 | **0.80** | 0.44 | **0.83** | 0.67 |
| MEN | 88 | 0.74 | 0.70 | 0.63 | 0.90 | **0.92** |
| TSR | 48 | 0.83 | 0.67 | **0.90** | **0.90** | 0.83 |
| MNM | 32 | 0.38 | 0.33 | 0.13 | **0.63** | 0.40 |
| CPN | 56 | 0.57 | 0.20 | 0.29 | **0.69** | 0.57 |
| CPT | 3 | **0.33** | 0.0 | 0.0 | 0.0 | **0.33** |
| HEM | 3 | **0.33** | **0.33** | 0.0 | **0.33** | 0.0 |
| EMB | 16 | **0.63** | 0.25 | 0.13 | 0.5 | 0.25 |
| PIN | 4 | **0.50** | 0.0 | 0.0 | 0.25 | 0.0 |
| MEL | 1 | 0.0 | 0.0 | 0.0 | 0.0 | 0.0 |
| Total | 385 | 0.65 | 0.51 | 0.44 | **0.72** | 0.66 |

In this table, we use a series of abbreviations to replace original categories: MET for brain metastases, GCT for germ cell tumors, GGN for gliomas, glioneuronal tumors, and neuronal tumors, MEN for meningioma, TSR for tumors of the sellar region, MNM for mesenchymal, non-meningothelial tumors, CPN for cranial and paraspinal nerve tumors, CPT for choroid plexus tumors, HEM for hematolymphoid tumors, EMB for embryonal tumors, PIN for pineal region tumors and MEL for melanocytic tumors.

# Supplementary Figures

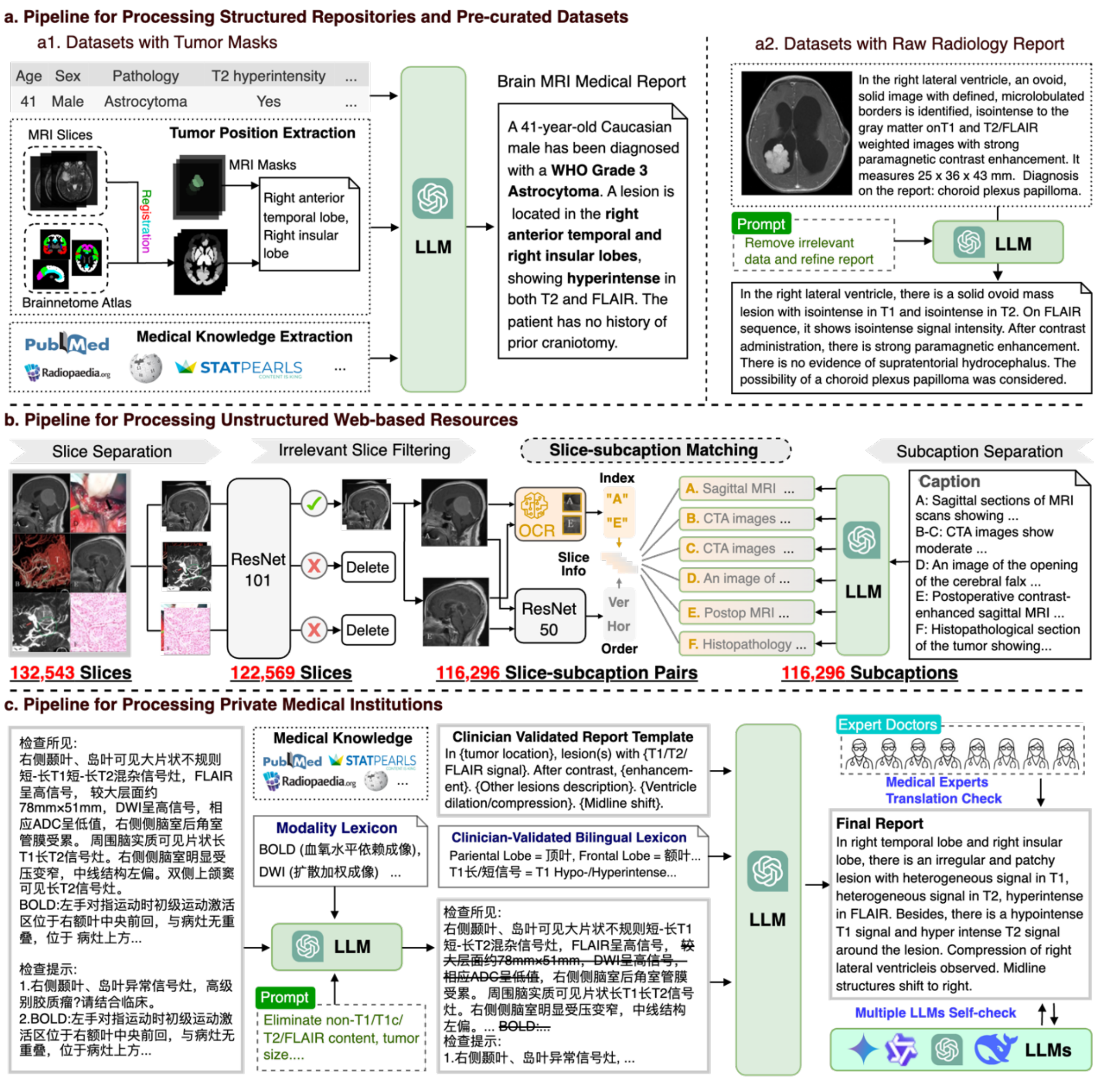

**Supplementary Figure 1 | Preprocessing pipeline for BrainTumor48K. a. The preprocessing pipeline for structured repositories and precurated datasets. a1:** For pre-curated datasets containing patient metadata and tumor masks, multi-sequence MRIs and their corresponding segmentation masks were co-registered to a standard neuroanatomical atlas to localize tumors. All available metadata—including imaging modality, sex, diagnostic labels, and extracted tumor descriptions—were utilized to generate comprehensive medical reports using ChatGPT-4o, prompted with RAG techniques. **a2:** For datasets comprising multi-parametric MRI scans and accompanying radiology reports (Radiopaedia), irrelevant information was removed from the original reports using prompt-based filtering with ChatGPT-4o. **b. Preprocessing workflow for unstructured web-based resources, such as PubMed.** Using PubMed Central as an example, we illustrate the preprocessing strategy for extracting MRI slice–subcaption pairs. First, brain tumor MRI figures were retrieved through keyword search, and irrelevant images were filtered using a ResNet-101 classifier. Second, compound images were decomposed into individual slices with a Faster R-CNN detector. Third, subcaptions were associated with image slices using Google OCR and ChatGPT-4o. The numbers of MRI slices and resulting slice–subcaption pairs are also summarized below. **c. The preprocessing pipeline for institutional medical data.** Chinese radiology reports were trimmed to retain modality-specific information, reformatted to the standard template, translated into English, and quality-checked by clinicians and multiple large-language models.

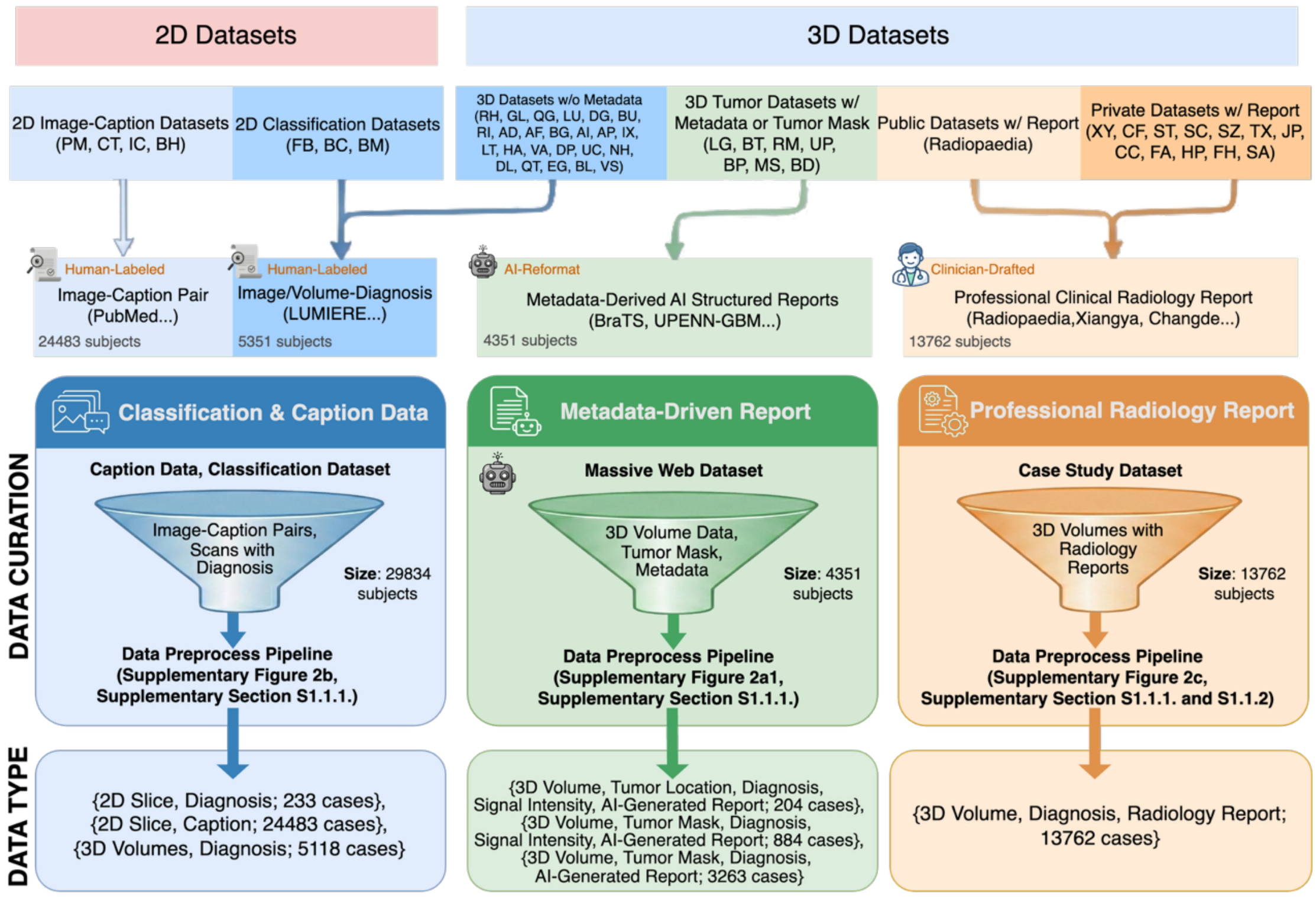

**Supplementary Figure 2 | Overview of the BrainTumor48K dataset curation pipeline and model training utilization.** The dataset integrates diverse 2D and 3D sources into three categories tailored for specific training stages: **(1) Classification & Caption Data (Blue, left):** 29,834 subjects from public 2D image-caption and classification datasets. This large-scale, heterogeneous data is primarily utilized in Stage 1 (2D slice-text representation learning) and Stage 2 (Hybrid 2D-3D training). **(2) Metadata-Driven Reports (Green, middle):** 4,351 subjects from 3D public datasets featuring AI-structured reports. Providing detailed spatial and signal grounding, this category is utilized in Stage 2 and Stage 3 (3D volumetric instruction tuning). **(3) Professional Radiology Reports (Orange, right):** 13,762 subjects with clinician-drafted reports from private and high-quality public sources. This strictly audited, high-fidelity data is also used in Stage 2 and Stage 3. **Dataset Abbreviations:** PM (PubMed Central), CT (Ctisus), IC (ImageCLEFmedical23), BH (Br35H), FB (Figshare Brain Tumor Dataset), BC (Brain Tumor Classification), BM (Brain Tumor MRI images 44 classes), LG (LGG-1p19qDeletion), BT (BraTS23), RM (ReMIND), RH (RHUH-GBM), UP (UPENN-GBM), GL (GLIS-RT), QG (QIN GBM Treatment Response), LU (LUMIERE), DG (OpenNeuro Diffuse Gliomas), BP (Brain-Tumor-Progression), BU (Burdenko), RI (RIDER), AD (ACRIN-DSC-MR-Brain), AF (ACRIN-FMISO-Brain), MS (Meningioma-SEG-CLASS), BD (Brain metastase MRI dataset), BG (Brain-TR-GammaKnife), AI (AOMIC-ID1000), AP (AOMIC-PIOP1), IX (IXI), LT (OpenNeuro Listening Task), HA (OpenNeuro Healthy Adults), VA (OpenNeuro Visual Audiovisual Speech), DP (OpenNeuro Dynamic Passive Threat), UC (OpenNeuro UCLA Consortium), NH (OpenNeuro NIMH Healthy), DL (DLBS dataset), QT (QTAB dataset), EG (EGD dataset), BL (Brain-Mets-Lung), VS (Vestibular-Schwannoma-MC2), XY (Xiangya Hospital), CF (Changde First People's Hospital), ST (Shanghai Tongji Hospital), SC (University of South China Second Hospital), SZ (Shenzhen Second People's Hospital), TX (The Third Xiangya Hospital), JP (Jiangxi Provincial People's Hospital), CC (Chongqing Traditional Chinese Medicine Hospital), FA (The First Affiliated Hospital of Nanchang University), HP (Hunan Provincial Children's Hospital), FH (The First Affiliated Hospital of Lanzhou University), SA (The Second Affiliated Hospital of Nanchang University).

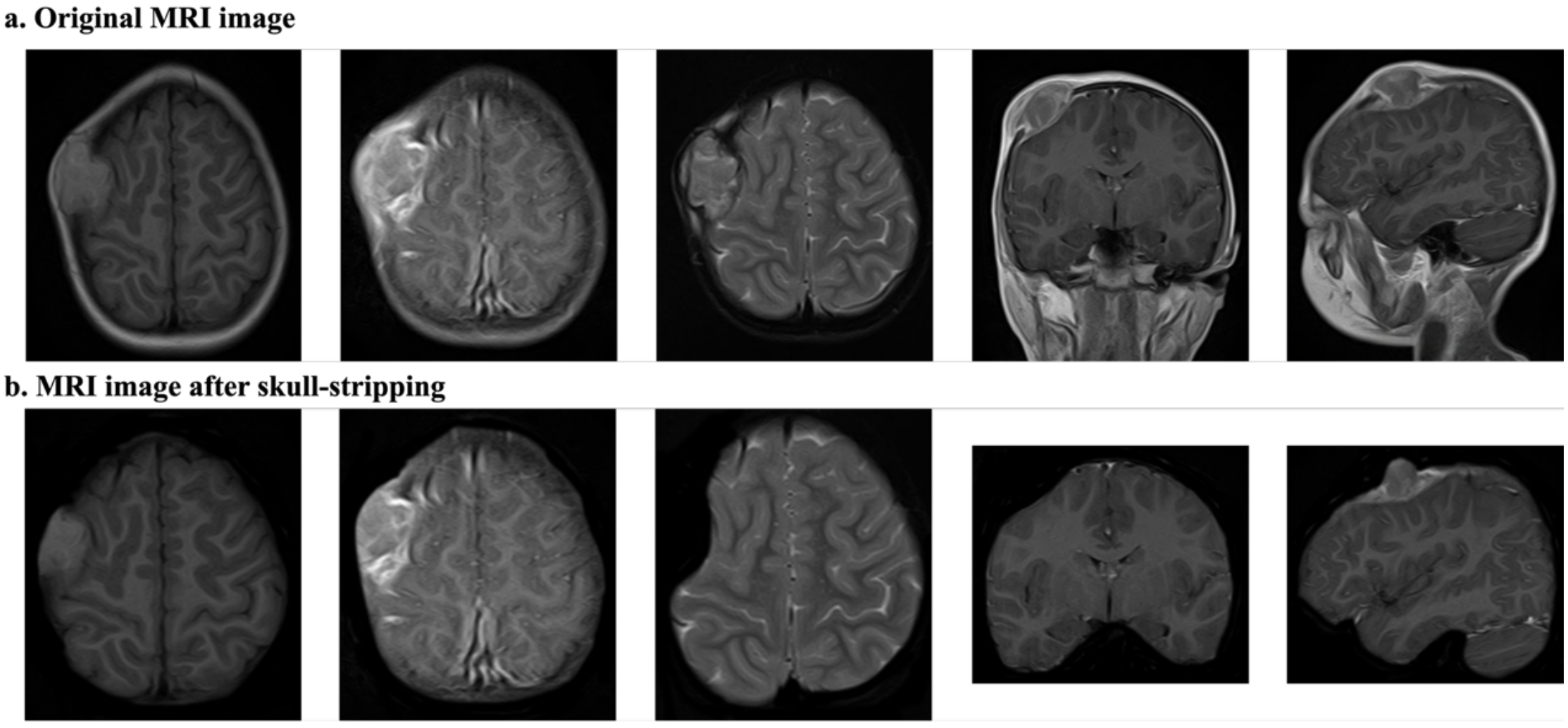

**Supplementary Figure 3 | Comparison of unprocessed and skull-stripped MRI images.** Representative examples of MRI scans before and after skull-stripping. While conventional skull-stripping may aid certain tumor analyses, it is not universally appropriate. Crucially, lesions located in regions like the pia mater or invading the cranial vault can be inadvertently removed during skull stripping, leading to the loss of critical anatomical and pathological information – a risk our approach deliberately avoids.

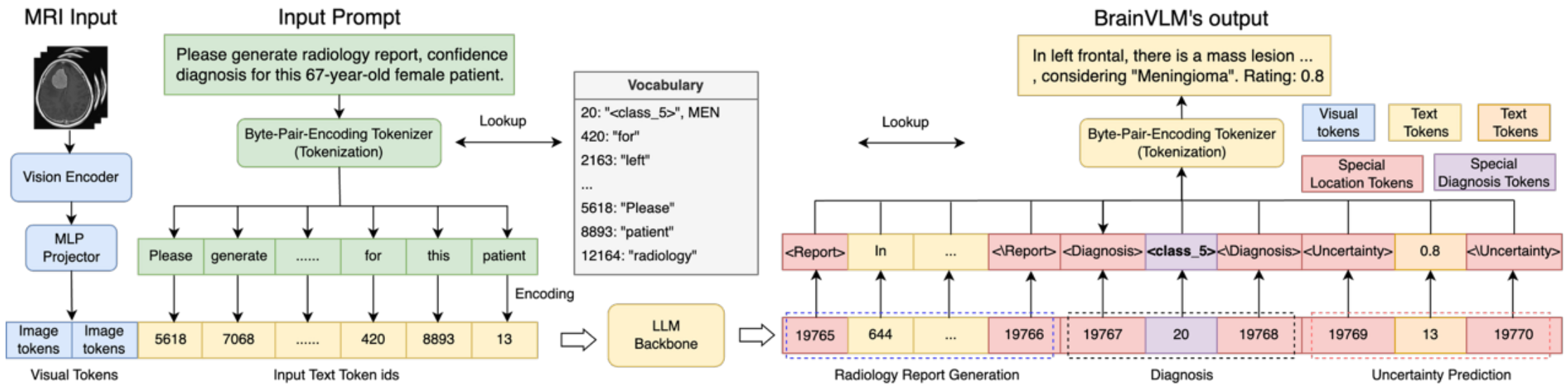

**Supplementary Figure 4 |** Data flow diagram of BrainVLM. This diagram provides a step-by-step visualization of how BrainVLM processes multi-modal input data and generates structured outputs. The model processes multi-modal inputs: multi-parametric MRI scans (processed by a vision encoder and projected into visual tokens) and textual data (patient metadata and instruction prompts, tokenized via BPE). These visual and textual token sequences are concatenated and fed into the LLM decoder (fine-tuned with LoRA). During autoregressive generation, three special start tokens (<Report>, <Diagnosis>, <Uncertainty>) guide the sequential output. First, standard BPE tokens are generated for the radiology report and decoded into free text. Next, a specialized classification token (e.g., <class_5>) is produced and mapped to a specific tumor grade or subtype (e.g., MEN) via a lookup table. Finally, numerical BPE tokens (e.g., "0", ".", "9") are generated and decoded into a confidence score (e.g., 0.9), which was discretized into six levels during training.

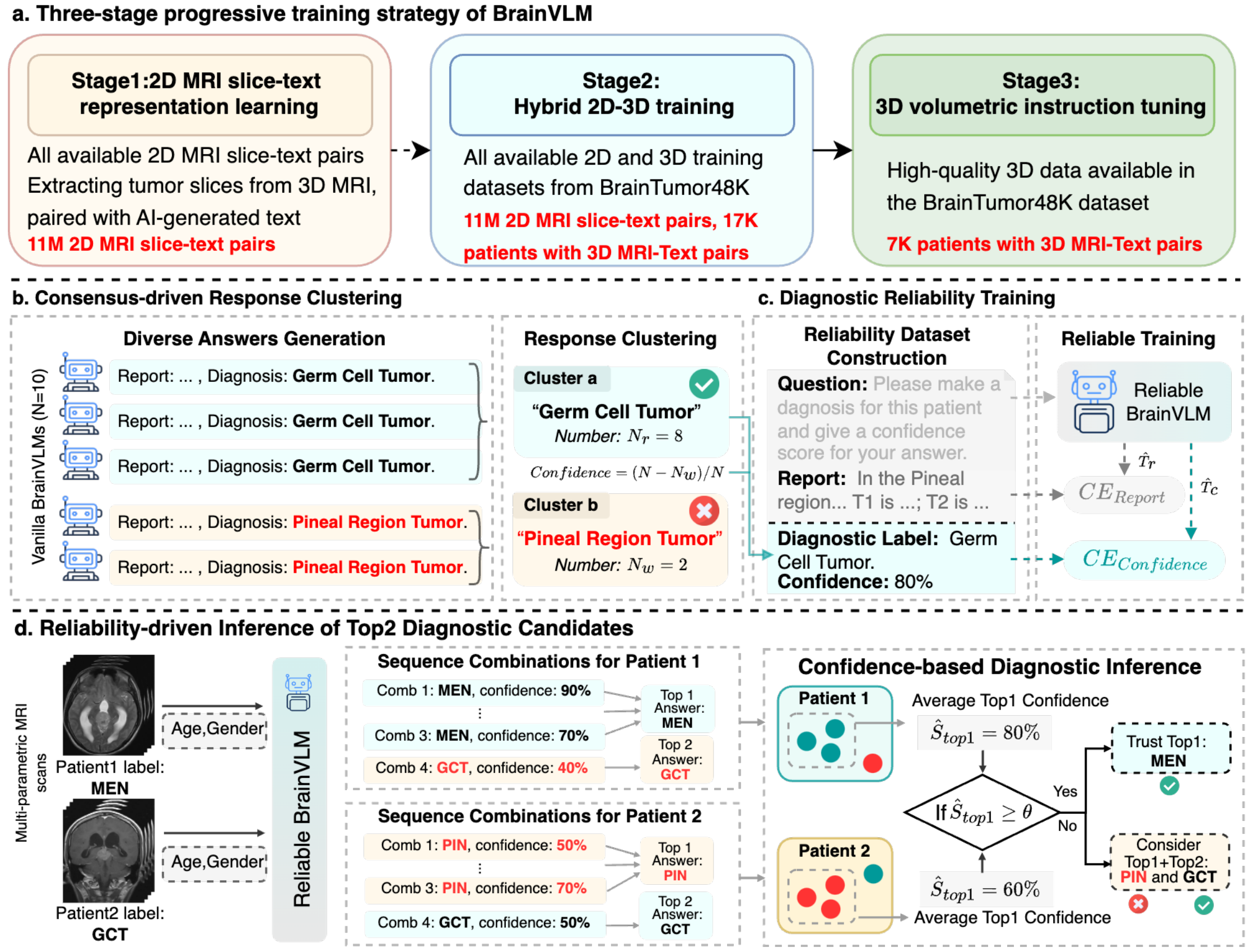

Supplementary Figure 5 | **Three-stage progressive training strategy and uncertainty quantification implementation of BrainVLM, including the confidence-triggered Top 2 supplementary diagnosis inference pipeline. a. Three-stage progressive training strategy of BrainVLM.** BrainVLM was developed using a three-stage progressive training approach: (1) 2D MRI slice-text representation learning using all curated 2D MRI slice-text pairs; (2) volumetric context integration via hybrid 2D-3D training with all available 2D and 3D training datasets; and (3) comprehensive 3D volumetric instruction tuning leveraging the high-quality 3D data in BrainTumor48K. **b,c,** Uncertainty quantification implementation pipeline. The uncertainty quantification of our BrainVLM consists of two steps: consensus-driven reliability dataset construction and confidenceaware reliable model finetuning. **b. Consensus-driven reliability dataset construction.** A reliability dataset is constructed by evaluating N vanilla BrainVLM models on the training dataset and clustering their diagnostic answers to derive confidence scores. **c. Confidence-aware reliable BrainVLM finetuning.** The consensus-derived reliability knowledge in the reliability dataset is distilled into reliable BrainVLM through confidence-aware reliable fine-tuning. **d. Confidence-triggered Top 2 supplementary diagnosis inference.** During deployment, each prediction made by the reliable BrainVLM is accompanied by a confidence score. If the confidence score of the Top 1 prediction falls below a predefined threshold (set at 75% in this study), the Top 2 prediction is also reported as a supplementary diagnosis to enhance clinical robustness. In this figure, we use abbreviations for tumor categories: MET (brain metastases), GCT (germ cell tumors), GGN (gliomas, glioneuronal tumors, and neuronal tumors), MEN (meningioma), TSR (tumors of the sellar region), MNM (mesenchymal, non-meningothelial tumors), CPN (cranial and paraspinal nerve tumors), CPT (choroid plexus tumors), HEM (hematolymphoid tumors), EMB (embryonal tumors), PIN (pineal region tumors), and MEL (melanocytic tumors).

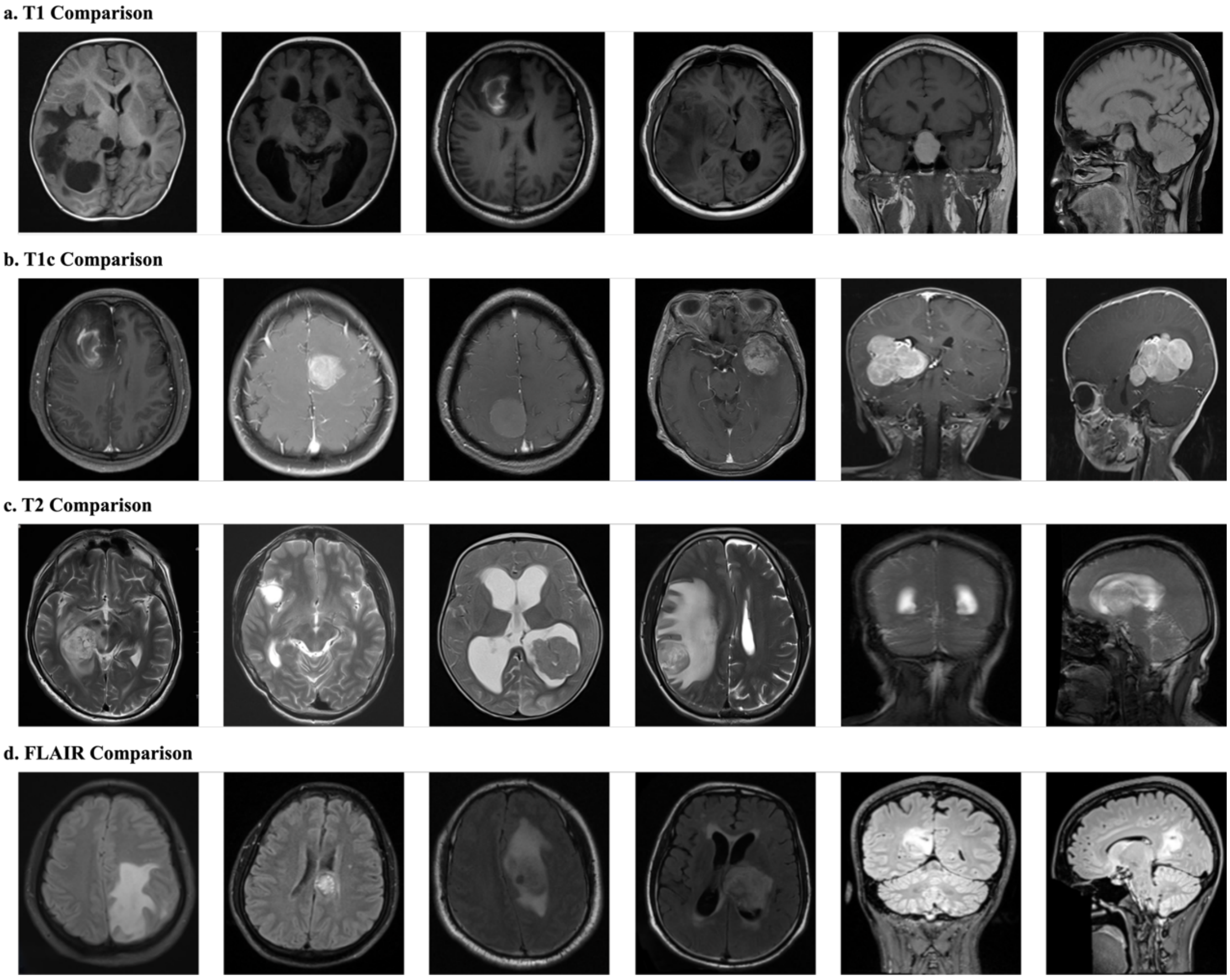

**Supplementary Figure 6 | Visualization of MRI images acquired using different imaging parameters across MRI sequences.** Shown are T1-weighted (T1), contrast-enhanced T1-weighted (T1c), T2-weighted (T2), and T2-Flair (FLAIR) sequences. For each sequence, four axial images obtained using different imaging parameters and MRI devices are presented. Additionally, one representative sagittal and one coronal view acquired under a specific parameter setting are included for each sequence.

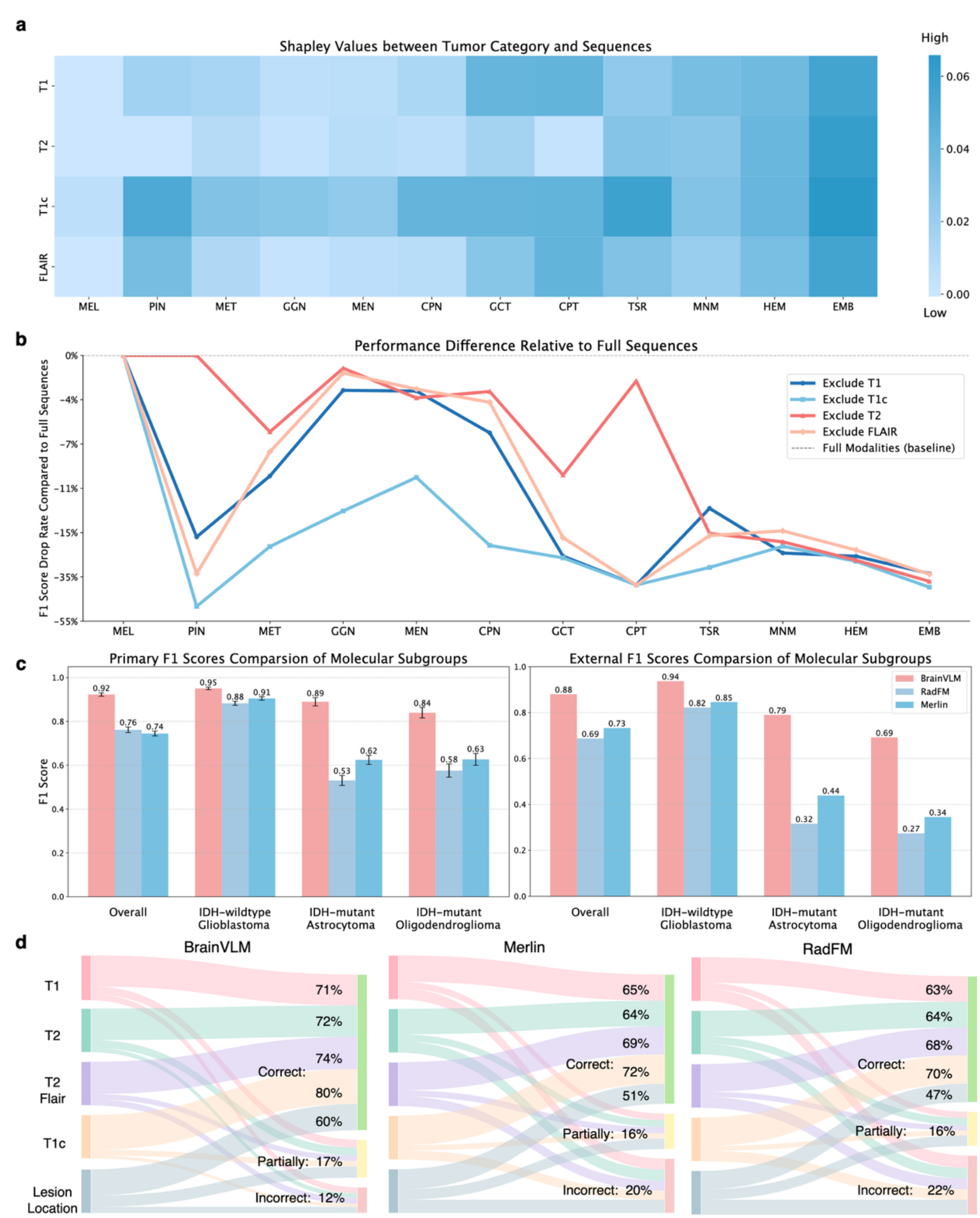

**Supplementary Figure 7 | Impact of individual MRI sequences on BrainVLM diagnostic accuracy for brain tumor classification and report result.** Diagnostic performance was evaluated in 3,571 patients by comparing the full-sequence model (T1, T1c, T2, T2-Flair) to models in which each sequence was individually omitted. Only patients with complete MRI data across all four sequences were included to eliminate potential confounding due to missing sequences. The remaining 1,555 patients in the original primary and external test datasets with missing sequences were excluded from the analysis. For each tumor type, the performance drop rate (measured in F1 score) resulting from the exclusion of a specific modality is shown (**b**). Both the direct performance drop rate and Shapley value analysis (**a**) highlight distinct sequence dependencies among tumor classes: T1c removal resulted in the greatest performance decline across all tumor types, underscoring its essential role. EMB classification was uniformly sensitive to the absence of any sequence (35% mean drop), while GGN and MEN predictions remained stable when T2 or T2-Flair or T1 was missing. In contrast, CPT predictions were more dependent on T1, T1c and T2-Flair, but showed resilience to the exclusion of T2. **c,** Molecular subgroups F1 scores comparison between our BrainVLM with Merlin and RadFM. **d.** Comparative analysis of key description accuracy among BrainVLM, Merlin, and RadFM, validated by Qwen-2.5-72B on the overall retrospective dataset.

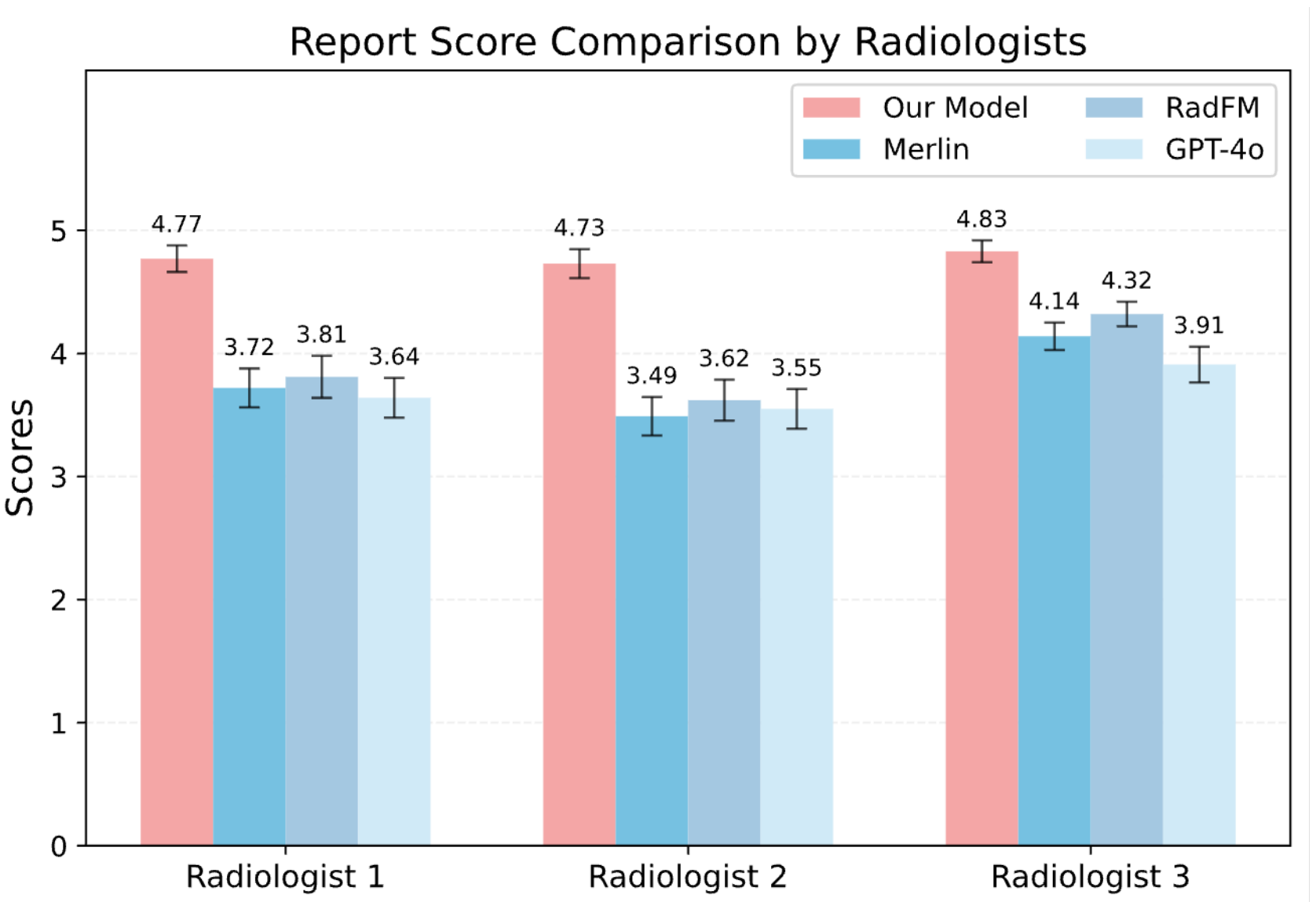

**Supplementary Figure 8 |** Comparison of report generation quality by three expert neuroradiologists. Reports generated by different models were scored on a 0-5 scale based on their consistency with ground truth clinical reports.

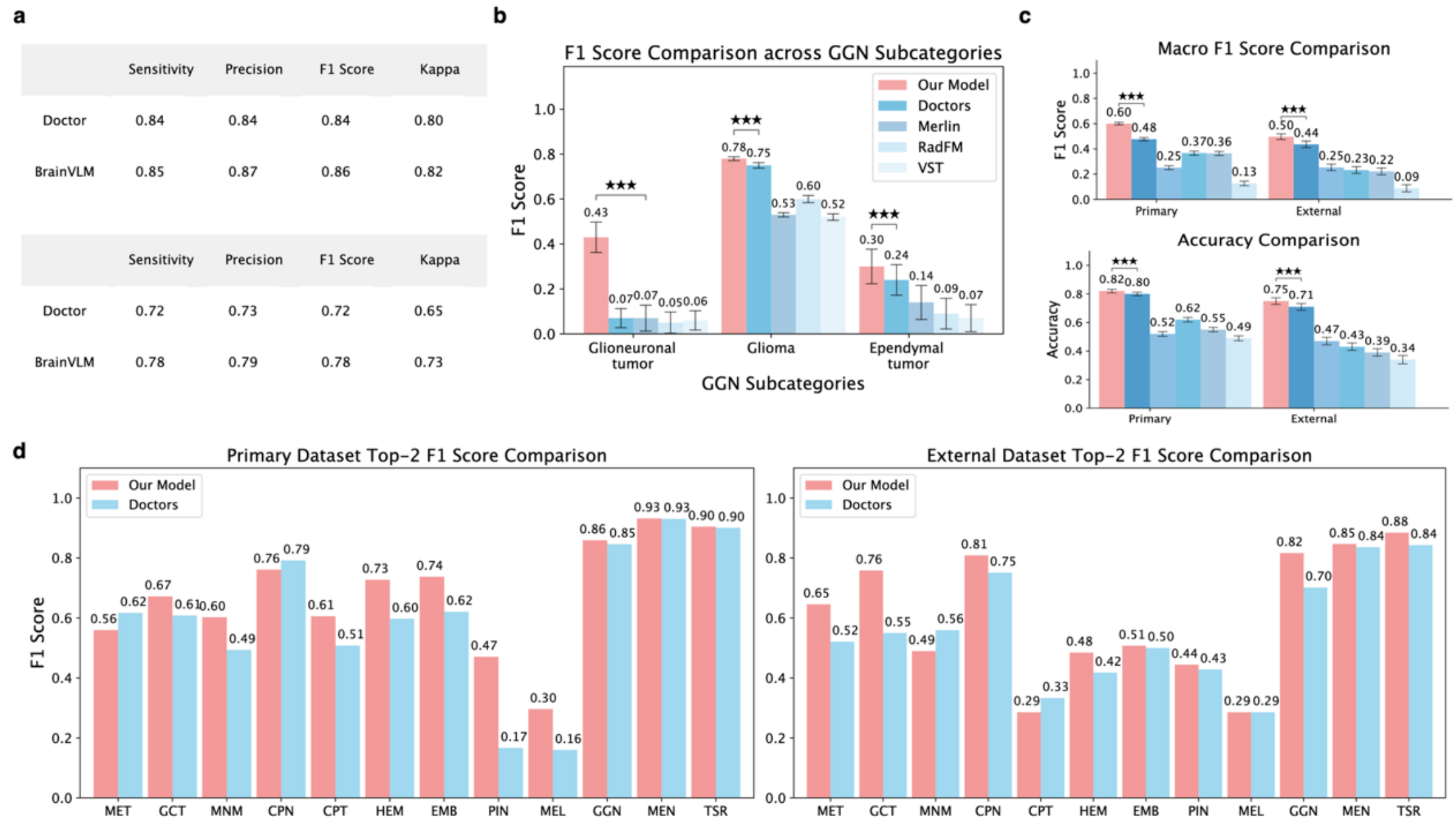

| | Sensitivity | Precision | F1 Score | Kappa |
|---|---|---|---|---|
| Doctor | 0.84 | 0.84 | 0.84 | 0.80 |
| BrainVLM | 0.85 | 0.87 | 0.86 | 0.82 |

| | Sensitivity | Precision | F1 Score | Kappa |
|---|---|---|---|---|
| Doctor | 0.72 | 0.73 | 0.72 | 0.65 |
| BrainVLM | 0.78 | 0.79 | 0.78 | 0.73 |

**Supplementary Figure 9 | Top 2 diagnosis performance of BrainVLM and analysis in AI-incorrect scenarios for multi-reader study. a,** Top 2 diagnosis metrics including F1 score, sensitivity, precision, and Cohen's kappa comparison between our BrainVLM and neuroradiologists. **b,** Diagnostic performance (F1 score) across specific GGN subcategories (glioneuronal tumors, gliomas, and ependymal tumors), comparing BrainVLM with clinicians and baseline models. **c,** Comparison of Macro F1 scores (top) and weighted accuracy (bottom) between BrainVLM, clinicians, and baseline models (Merlin, RadFM) on the primary and external test datasets.**d,** F1 score comparison across 12 tumor types, evaluating the Top 2 predictions of our model and neuroradiologists on both primary and external datasets. Abbreviations: MET = brain metastases, GCT = germ cell tumors, GGN = gliomas, glioneuronal tumors, and neuronal tumors, MEN = meningioma, TSR = tumors of the sellar region, MNM = mesenchymal, non-meningothelial tumors, CPN = cranial and paraspinal nerve tumors, CPT = choroid plexus tumors, HEM = hematolymphoid tumors, EMB = embryonal tumors, PIN = pineal region tumors and MEL = melanocytic tumors.

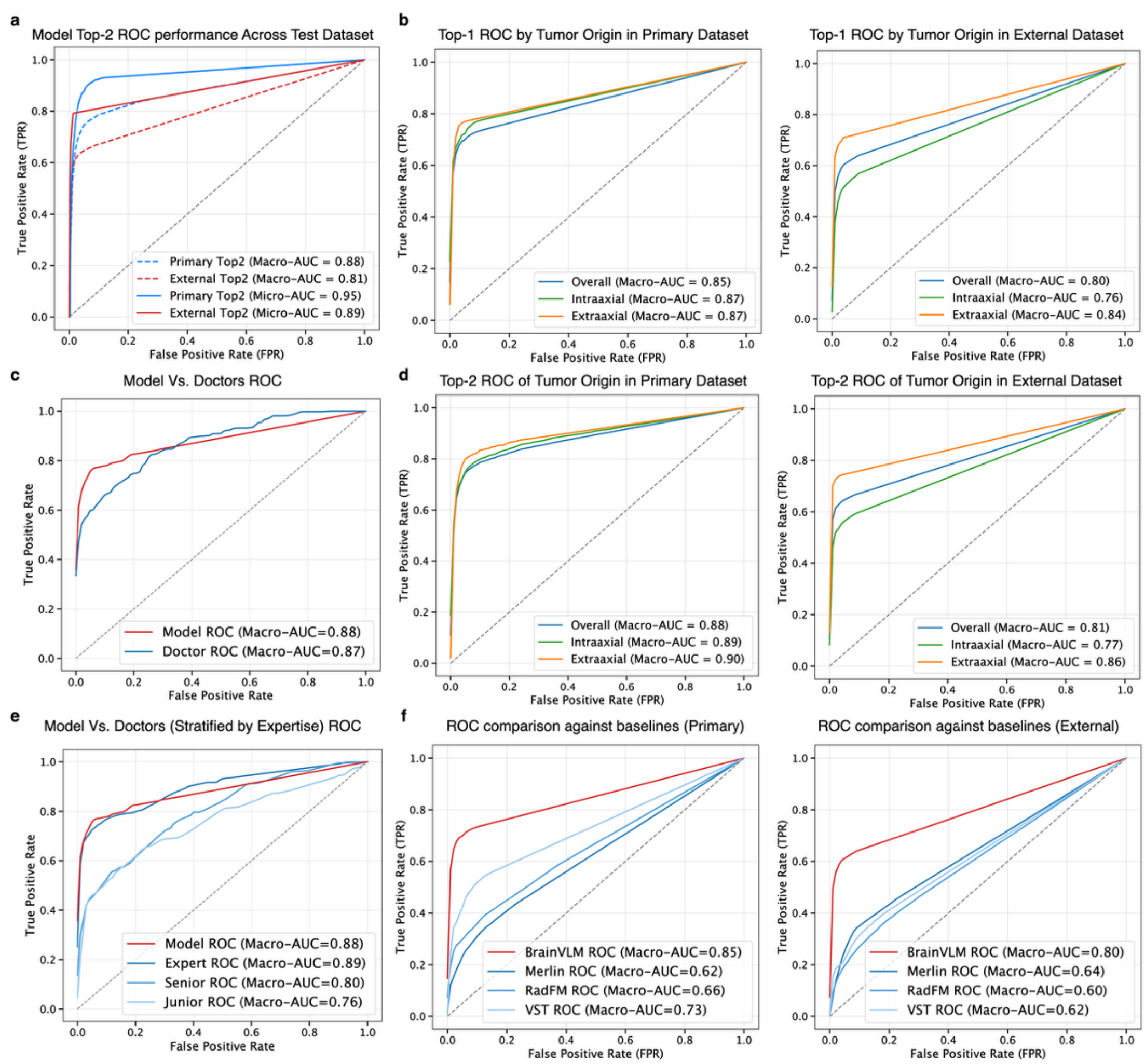

**Supplementary Figure 10 | Comparison of ROC curves among BrainVLM, Neuroradiologists and other baseline models**. **a,** ROC curves for Top-2 predictions on the retrospective primary and external datasets, with macro-averaged AUCs of 0.88/0.81 and micro-averaged 0.95/0.89, respectively. **b,** Macro Top1 ROC across intraaxial and extraaxial tumor type. **c,** Micro-averaged ROC curves comparing BrainVLM against human readers in an expanded multi-reader study. **d,** Macro Top2 ROC across intraaxial and extraaxial tumor type. **e,** Performance was evaluated on 248 retrospective brain tumor cases interpreted by 12 neuroradiologists, who were stratified by their expertise levels into expert, senior, and junior groups. **f,** Macro-ROC comparison against baseline models (Merlin, RadFM, and VST) on the primary (left) and external (right) datasets.

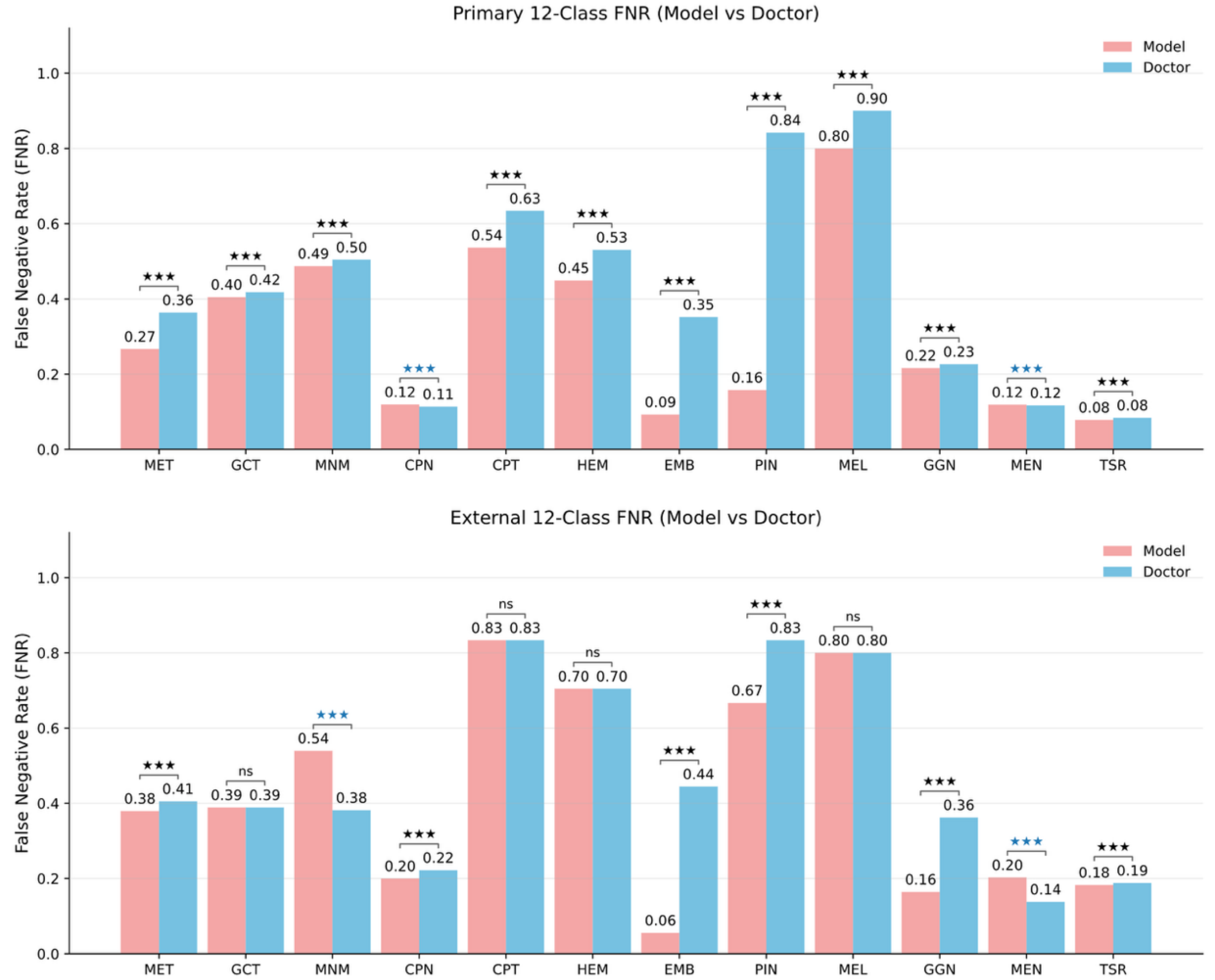

**Supplementary Figure 11 | Comparison of the false negative rates for 12 brain tumor types between BrainVLM and board-certified neuroradiologists in the primary (top) and external (bottom) retrospective datasets**. Statistical significance was determined using a one-sided Wilcoxon test. Significance levels are denoted by asterisks (★ $p<0.05$, ★★ $p<0.01$, ★★★ $p<0.001$; ns, not significant). The color of the significance markers indicates the direction of the one-sided hypothesis test: black markers denote tests evaluating if the model's FNR is significantly lower than the doctors' (Model < Doctor), while blue markers denote tests evaluating if the doctors' FNR is significantly lower than the model's (Doctor < Model).

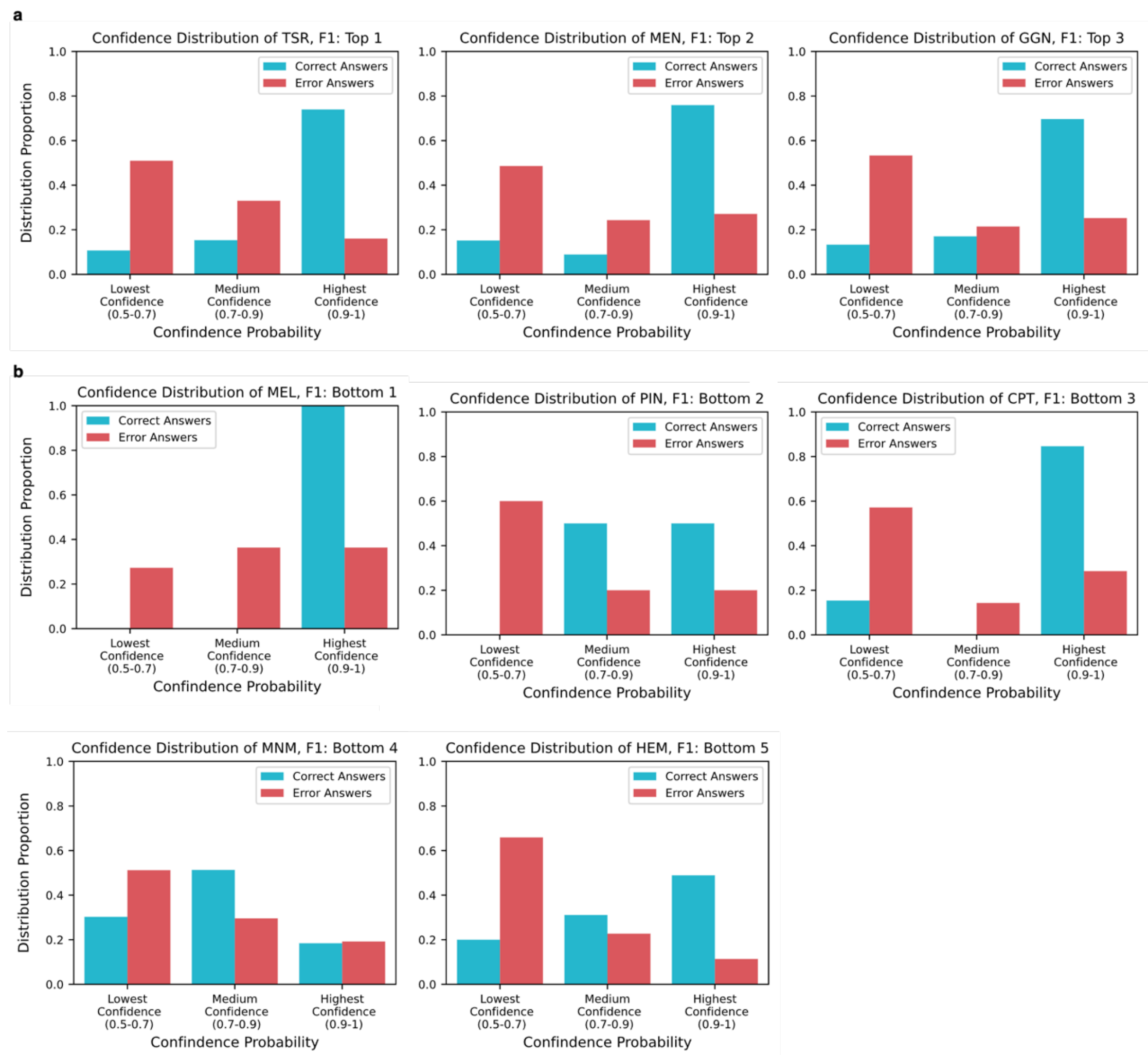

**Supplementary Figure 12 | a.** Confidence distribution of top 3 classes including TSR, GGN and MEN. **b.** Confidence distribution of bottom 5 classes based on F1 score including PIN, MEL, CPT, MNM and MET. In this table, we use abbreviations for tumor categories: MET (brain metastases), GCT (germ cell tumors), GGN (gliomas, glioneuronal tumors, and neuronal tumors), MEN (meningioma), TSR (tumors of the sellar region), MNM (mesenchymal, non-meningothelial tumors), CPN (cranial and paraspinal nerve tumors), CPT (choroid plexus tumors), HEM (hematolymphoid tumors), EMB (embryonal tumors), PIN (pineal region tumors), and MEL (melanocytic tumors).

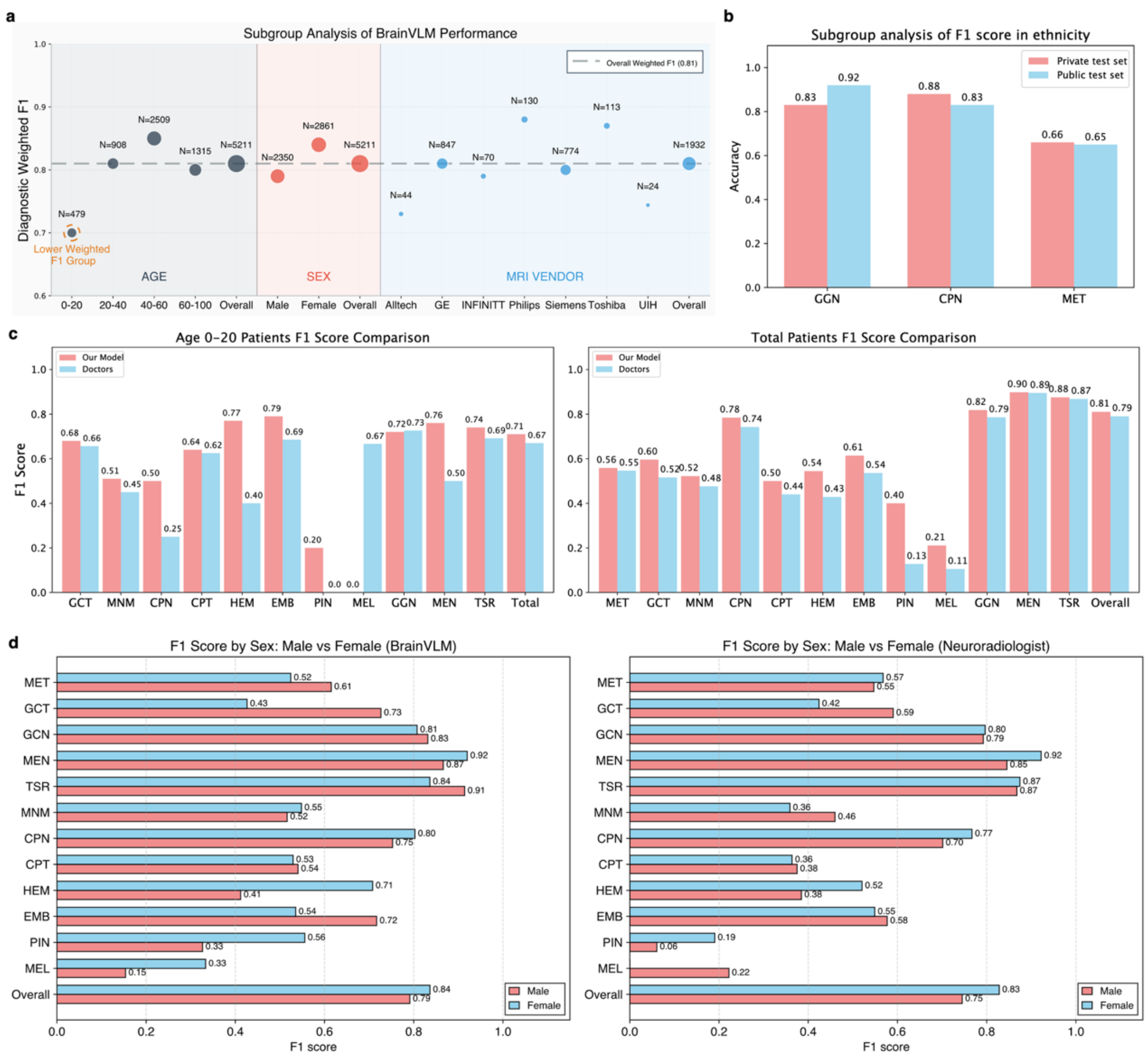

**Supplementary Figure 13 | Subgroup analysis across age, sex, and MRI vendor. a.** BrainVLM *F1* score across different age, sex, and MRI vendors (GE, Philips, Siemens, etc.). **b.** F1 comparison between different ethnicity source: Newly public dataset (EGD dataset, Brain-Mets-Lung and Vestibular-Schwannoma-MC2). **c.** F1-score comparison between BrainVLM (red) and doctors (blue) for pediatric (0–20 years) and total cohorts **d.** F1 comparison between BrainVLM and Radiologist in male and female. In this table, we use abbreviations for tumor categories: MET (brain metastases), GCT (germ cell tumors), GGN (gliomas, glioneuronal tumors, and neuronal tumors), MEN (meningioma), TSR (tumors of the sellar region), MNM (mesenchymal, non-meningothelial tumors), CPN (cranial and paraspinal nerve tumors), CPT (choroid plexus tumors), HEM (hematolymphoid tumors), EMB (embryonal tumors), PIN (pineal region tumors), and MEL (melanocytic tumors).

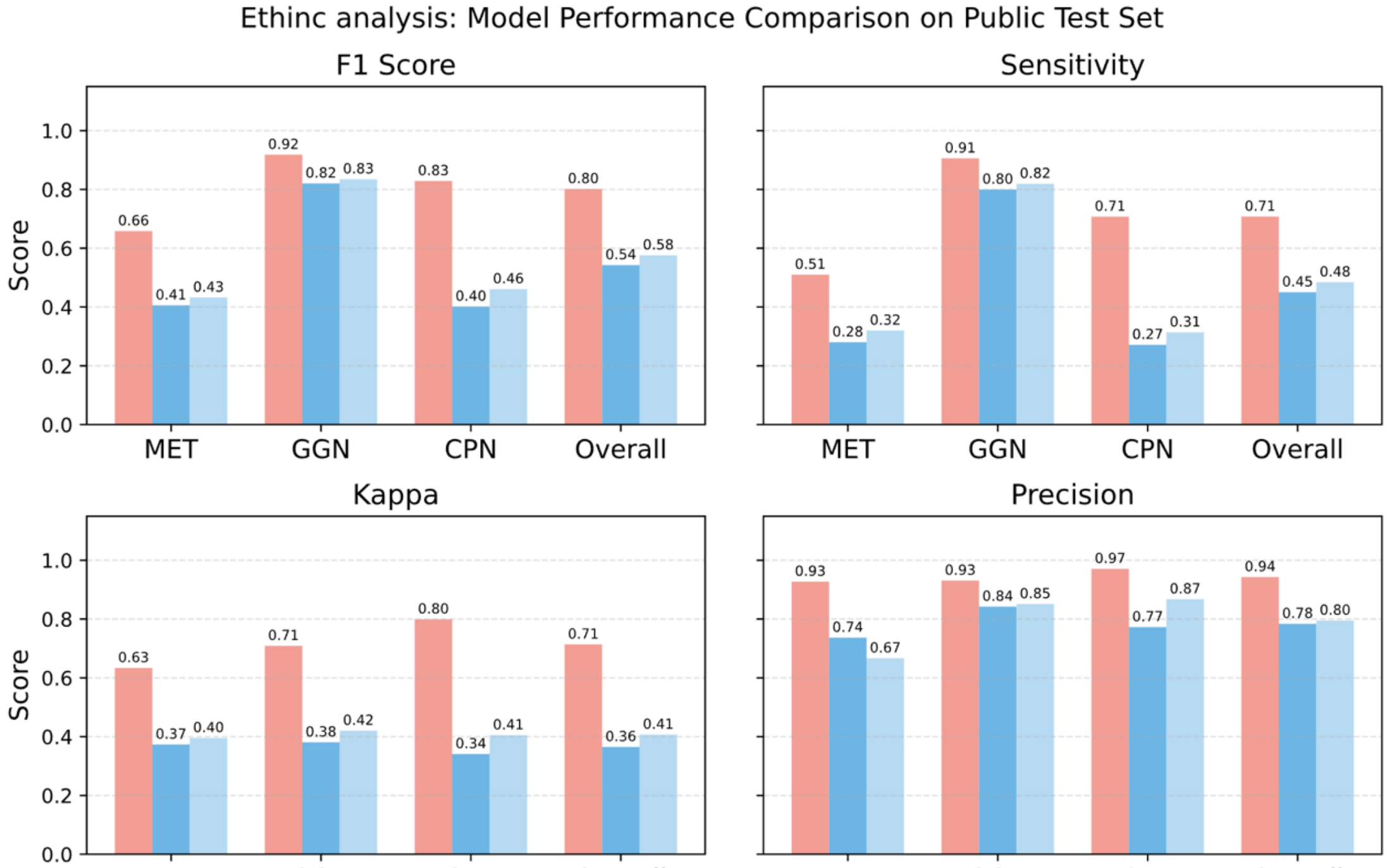

**Supplementary Figure 14 | Performance comparison between BrainVLM, Merlin and RadFM in the public ethnic dataset.** GGN represents the EGD dataset, CPN represents the Vestibular-Schwannoma-MC2 dataset and MET represents the Brain-Mets-Lung dataset.

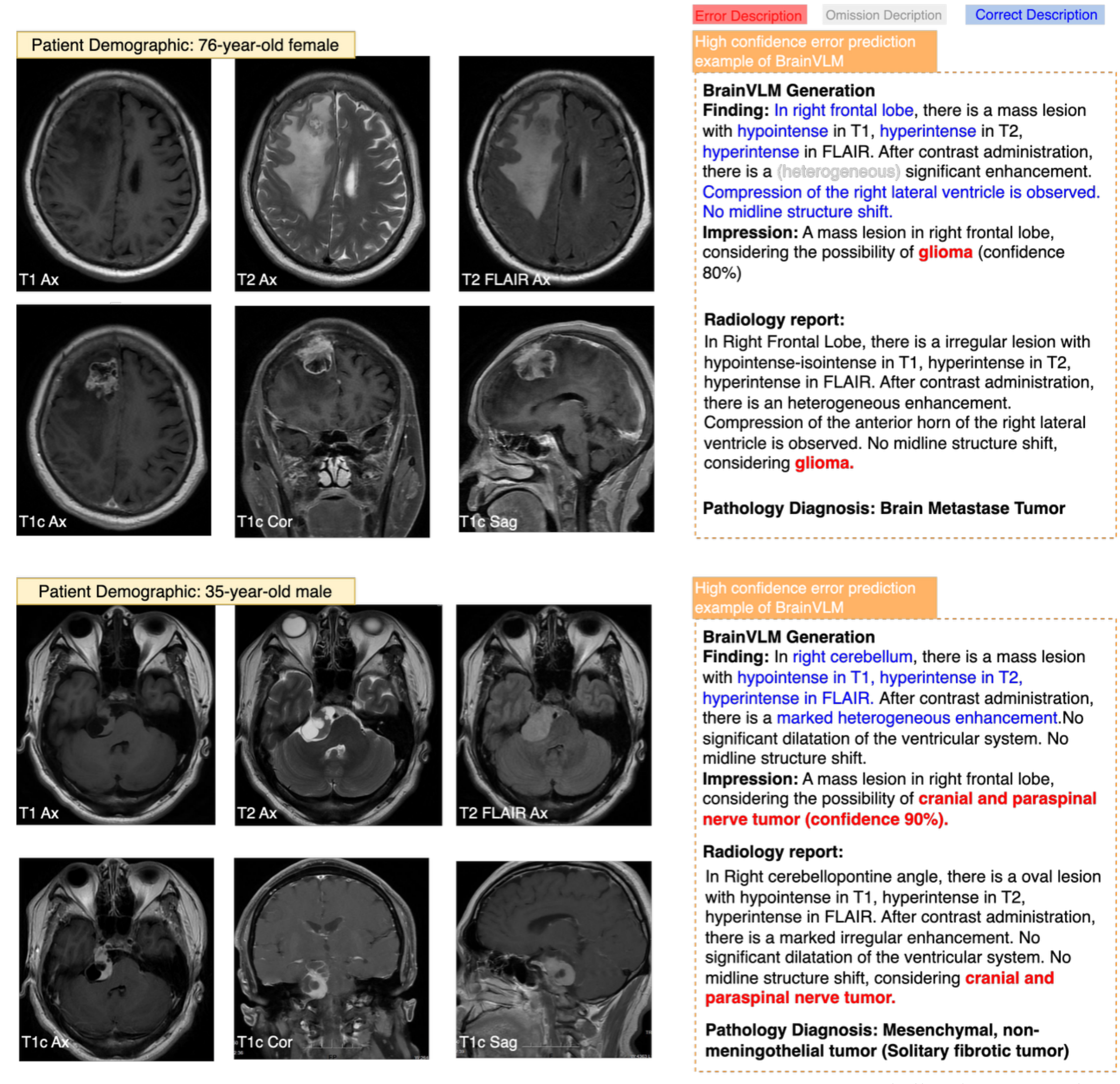

**Supplementary Figure 15 | High-confidence diagnostic errors by BrainVLM.** Two challenging cases where BrainVLM generated incorrect diagnoses with high confidence. The generated reports are compared with multimodal MRI, the neuroradiologist's report, and the pathology gold standard. Text colors indicate accurate features (blue), omissions (grey), and diagnostic errors (red). Notably, BrainVLM's misdiagnoses aligned with the human radiologists' errors, highlighting shared clinical reasoning vulnerabilities in highly deceptive cases.

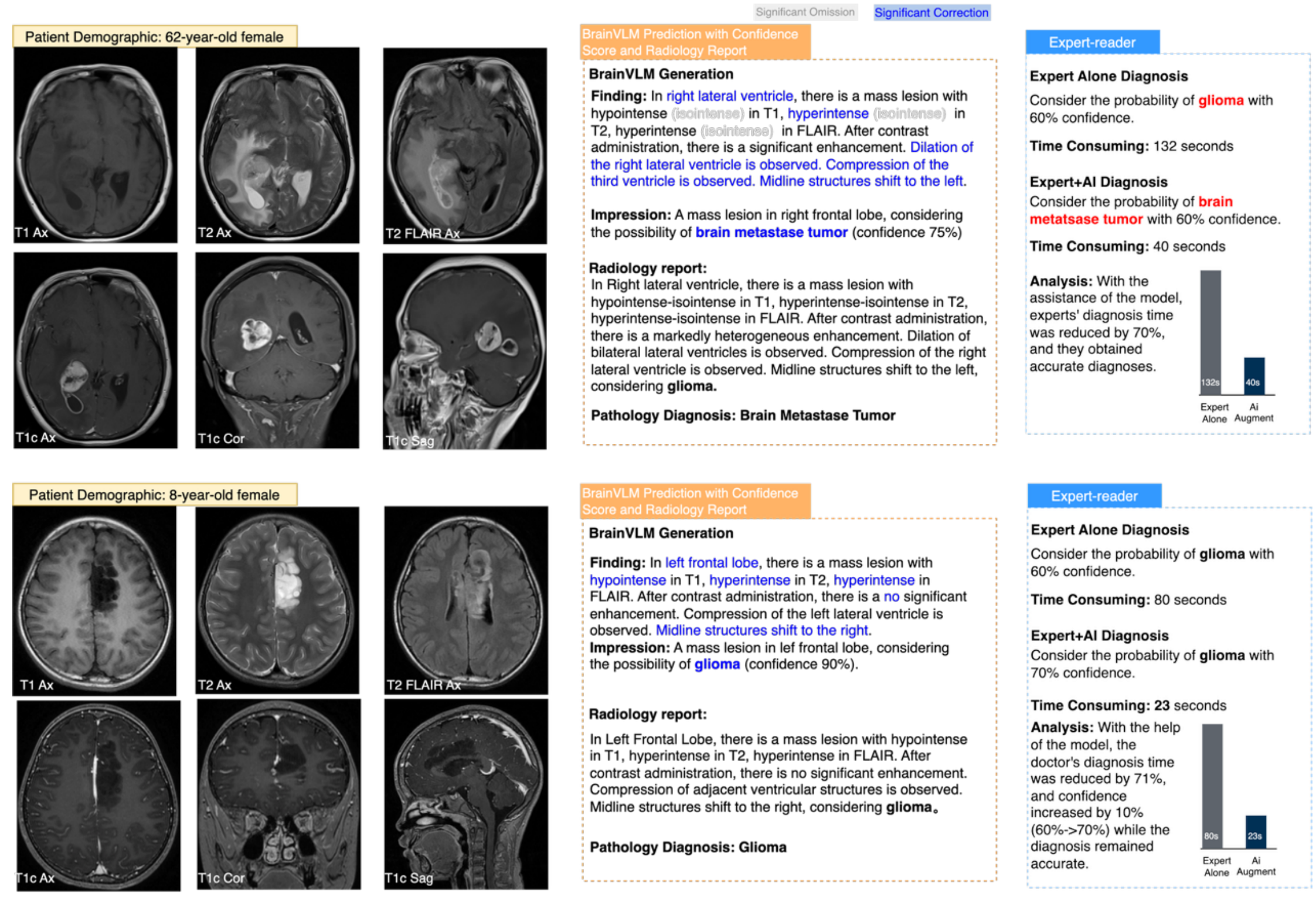

**Supplementary Figure 16 |** Case 1: A brain metastasis patient where BrainVLM-assisted expert diagnosis reduced reading time by 70% (from 132s to 40s) and corrected the initial misdiagnosis. Case 2: A glioma patient with BrainVLM assistance reduced reading time by 71% while increasing diagnostic confidence from 60% to 70%.

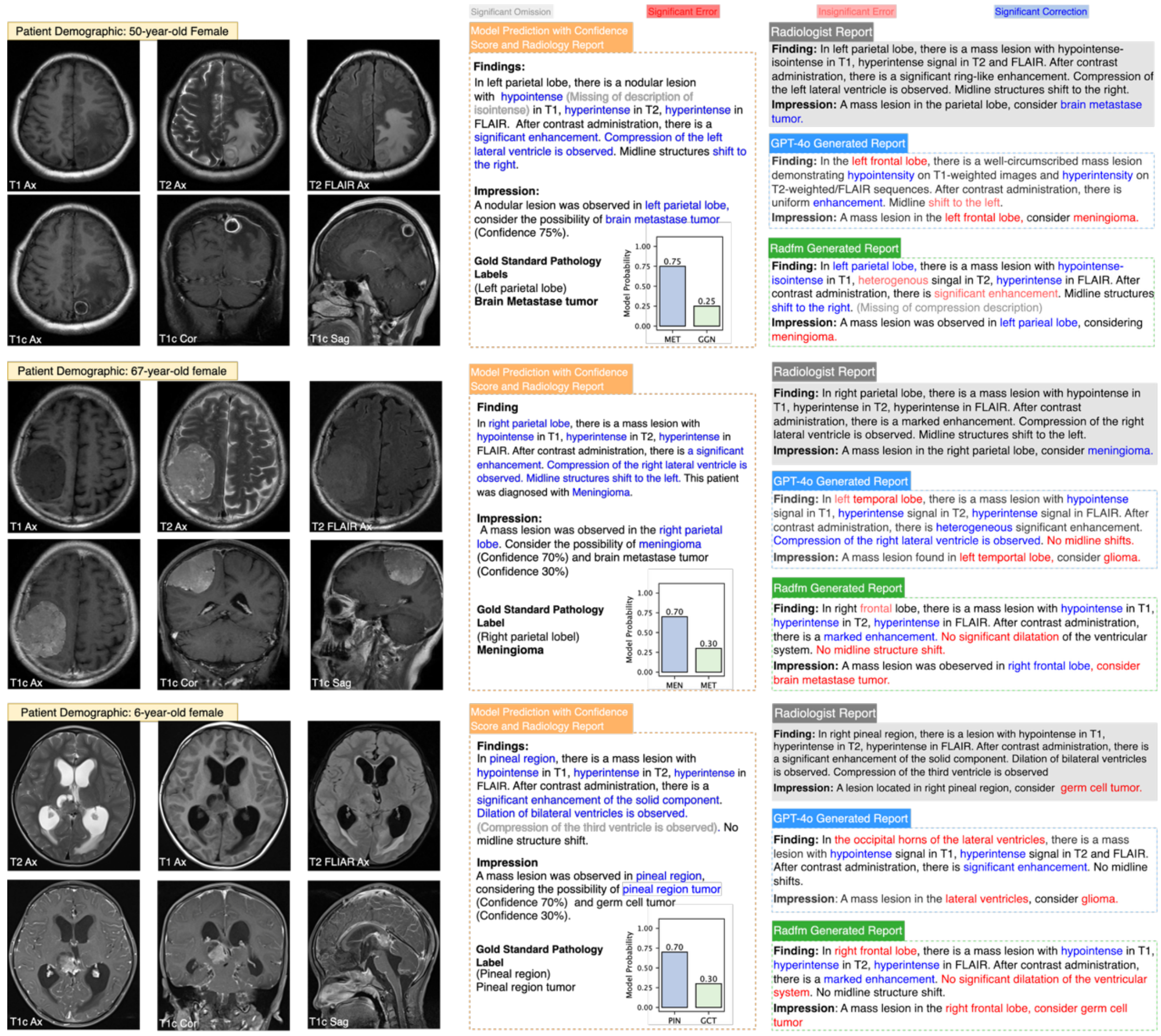

**Supplementary Figure 17 | Examples of reports and diagnoses generated by our BrainVLM, compared with those from doctors (expert-curated radiology report) and other state-of-the-art AI models (RadFM and ChatGPT-4o).** For each diagnosis generated by BrainVLM, a corresponding reliability score is also provided. We use a series of abbreviations to replace original categories: MET for brain metastases, GCT for germ cell tumors, GGN for gliomas, glioneuronal tumors, and neuronal tumors, MEN for meningioma, TSR for tumors of the sellar region, MNM for mesenchymal, non-meningothelial tumors, CPN for cranial and paraspinal nerve tumors, CPT for choroid plexus tumors, HEM for hematolymphoid tumors, EMB for embryonal tumors, PIN for pineal region tumors and MEL for melanocytic tumors.

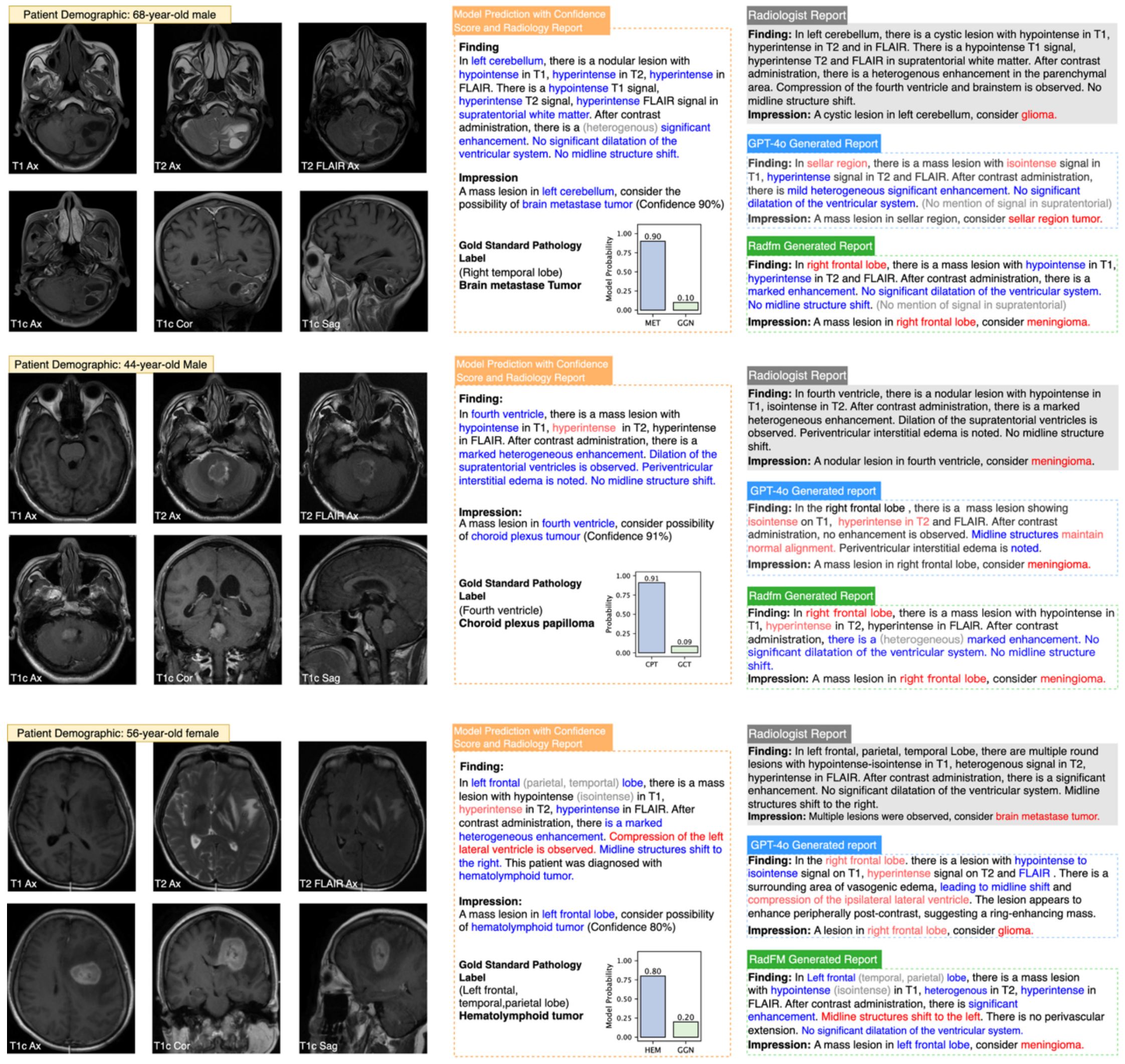

**Supplementary Figure 18 | Examples of reports and diagnoses generated by our BrainVLM, compared with those from doctors (expertcurated radiology report) and other state-of-the-art AI models (RadFM and ChatGPT-4o).** For each diagnosis generated by BrainVLM, a corresponding reliability score is also provided. We use a series of abbreviations to replace original categories: MET for brain metastases, GCT for germ cell tumors, GGN for gliomas, glioneuronal tumors, and neuronal tumors, MEN for meningioma, TSR for tumors of the sellar region, MNM for mesenchymal, non-meningothelial tumors, CPN for cranial and paraspinal nerve tumors, CPT for choroid plexus tumors, HEM for hematolymphoid tumors, EMB for embryonal tumors, PIN for pineal region tumors and MEL for melanocytic tumors.